\documentclass[journal]{IEEEtran}
\usepackage{amsmath,amsfonts}
\usepackage{algorithmic}
\usepackage{algorithm}
\usepackage{array}
\usepackage[caption=false,font=footnotesize]{subfig}
\usepackage{textcomp}
\usepackage{stfloats}
\usepackage{url}
\usepackage{verbatim}
\usepackage{graphicx}

\usepackage{tikz}
\usetikzlibrary{external}
\usetikzlibrary{shadows}
\usetikzlibrary{patterns}
\usetikzlibrary{mindmap,trees}
\usetikzlibrary{calc,3d}
\usetikzlibrary{patterns,hobby}
\usepackage{etoolbox} 
\usepackage{listofitems} 
\usepackage{adjustbox}
\usepackage{pgfplots}
\usepackage{forest}
\usepackage{pifont}
\usepackage{enumitem}

\usepackage[style=ieee,citestyle=numeric-comp,maxcitenames=2,mincitenames=1]{biblatex}
\usepackage[acronym]{glossaries}
\makeglossaries
\newacronym{ml}{ML}{Machine Learning}
\newacronym{ai}{AI}{Artificial Intelligence}
\newacronym{dl}{DL}{Deep Learning}
\newacronym{gnn}{GNN}{Graph Neural Networks}
\newacronym{gcn}{GCN}{Graph Convolutional Networks}

\newacronym{xai}{xAI}{explainable AI}
\newacronym{iml}{IML}{interpretable Machine Learning}
\newacronym{cv}{CV}{Computer Vision}
\newacronym{sota}{SOTA}{state-of-the-art}
\newacronym{bp}{BP}{Backpropagation}
\newacronym{ood}{OOD}{Out-Of-Distribution}
\newacronym{ssl}{SSL}{Self-Supervised Learning}
\newacronym{moref}{MoRef}{Most Relevant First}
\newacronym{iou}{IoU}{Intersection over Union}

\newacronym{eo}{EO}{Earth Observation}
\newacronym{rs}{RS}{Remote Sensing}
\newacronym{sst}{SST}{Sea Surface Temperature}
\newacronym{sar}{SAR}{Synthetic Aperture Radar}
\newacronym{dem}{DEM}{Digital Elevation Model}
\newacronym{modis}{MODIS}{Moderate Resolution Imaging Spectroradiometer}
\newacronym{RGB}{RGB}{Red-Green-Blue}
\newacronym{LiDAR}{LiDAR}{Light Detection and Ranging}
\newacronym{uav}{UAV}{Unmanned Aerial Vehicles}
\newacronym{dsm}{DSM}{Digital Surface Model}
\newacronym{ndvi}{NDVI}{Normalized Difference Vegetation Index}
\newacronym{fapar}{FAPAR}{Fraction of Absorbed Photosynthetically Active Radiation}
\newacronym{wlai}{WLAI}{Water-Limited Leaf Area Index}
\newacronym{polsar}{PolSAR}{Polarimetric SAR}
\newacronym{asc}{ASC}{attribute scattering center}
\newacronym{cap}{CAP}{Common Agricultural Policy}

\newacronym{lr}{LR}{Linear Regression}
\newacronym{rf}{RF}{Random Forest}
\newacronym{svm}{SVM}{Support Vector Machine}
\newacronym{xgboost}{XGBoost}{eXtreme Gradient Boosting}
\newacronym{gb}{GB}{Gradient Boosting}
\newacronym{ebm}{EBM}{Explainable Boosting Machines}
\newacronym{gp}{GP}{Gaussian Process}
\newacronym{knn}{kNN}{k-Nearest-Neighbor}

\newacronym{nn}{NN}{Neural Network}
\newacronym{mlp}{MLP}{Multilayer Perceptron}
\newacronym{dnn}{DNN}{Deep Neural Network}
\newacronym{rnn}{RNN}{Recurrent Neural Network}
\newacronym{cnn}{CNN}{Convolutional Neural Network}
\newacronym{fnn}{FNN}{Feed Forward Neural Network}
\newacronym{lstm}{LSTM}{Long Short-Term Memory neural network}
\newacronym{gru}{GRU}{Gated Recurrent Unit}
\newacronym{vt}{ViT}{Vision Transformer}
\newacronym{gan}{GAN}{Generative Adversarial Network}

\newacronym{snn}{SNN}{Superposable Neural Network}

\newacronym{mdi}{MDI}{Mean Decrease in Impurity}

\newacronym{cam}{CAM}{Class Activation Mapping}
\newacronym{gradcam}{Grad-CAM}{Gradient-weighted Class Activation Mapping}
\newacronym{guidedgradcam}{GuidedGrad-CAM}{Gradient-weighted Class Activation Mapping}
\newacronym{lda}{LDA}{Latent Dirichlet Allocation}
\newacronym{pdp}{PDP}{Partial Dependence Plot}
\newacronym{ale}{ALE}{Accumulated Local Effects}
\newacronym{shap}{SHAP}{SHapley Additive exPlanations}
\newacronym{dtd}{DTD}{Deep Taylor Decomposition}
\newacronym{lime}{LIME}{Local Interpretable Model-agnostic Explanation}
\newacronym{lrp}{LRP}{Layer-wise Relevance Propagation}
\newacronym{glm}{GLM}{Generalized Linear Model}
\newacronym{gam}{GAM}{Generalized Additive Model}
\newacronym{smGrad}{SmoothGrad}{Smooth Gradient}
\newacronym{ig}{IG}{Integrated Gradients}
\newacronym{eg}{EG}{Expected Gradients}
\newacronym{I*G}{I*G}{Input*Gradient}
\newacronym{umap}{UMAP}{Uniform Manifold Approximation and Projection}
\newacronym{pca}{PCA}{Principal Component Analysis}
\newacronym{tsne}{t-SNE}{t-distributed Stochastic Neighbor Embedding}
\newacronym{fls}{FLS}{Fuzzy Logic System}
\newacronym{owa}{OWA}{Ordered Weighted Averaging}
\newacronym{anfis}{ANFIS}{Adaptive Neuro Fuzzy Inference System}
\newacronym{drb}{DRB}{Deep Rule-Based}
\newacronym{tcav}{TCAV}{Testing with Concept Activation Vectors}
\newacronym{deeplift}{DeepLIFT}{Deep Learning Important FeaTures}
\newacronym{nlp}{NLP}{Natural Language Processing}
\newacronym{ram}{RAM}{Regression Activation Mapping}
\newacronym{kl}{KL}{Kullback-Leiber}
\newacronym{tda}{TDA}{topological data analysis}
\newacronym{pfi}{PFI}{Permutation Feature Importance}

\newacronym{aCNN}{aCNN}{CNN with attention}
\newacronym{aLSTM}{aLSTM}{LSTM with attention}
\newacronym{GFFS}{GFFS}{Grouped Forward Feature Selection}
\newacronym{vae}{VAE}{Variational Autoencoder}
\newacronym{SMLP}{SMLP}{MLP with a presingle-connection layer}

\setacronymstyle{long-short}

\usepackage{longtable}

\usepackage{listings}

\usepackage[hidelinks]{hyperref}
\hypersetup{
    colorlinks = true,
    citecolor = {cyan},
    linkcolor =.,
    menucolor=.,
}

\newcommand{\highlight}[2]{%
  \hyperref[#1]{\fcolorbox{cyan}{white!10}{#2}}%
}

\begin{document}

\title{Opening the Black-Box: A Systematic Review on Explainable AI in Remote Sensing}

\author{Adrian Höhl$^{*}$, Ivica Obadic$^{*}$, Miguel-\'Angel Fern\'andez-Torres, Hiba Najjar, \\ Dario Oliveira, Zeynep Akata, Andreas Dengel and Xiao Xiang Zhu, ~\IEEEmembership{Fellow,~IEEE.}\\
\thanks{
$^{*}$Adrian Höhl and Ivica Obadic share the first authorship. Corresponding authors: Adrian Höhl, Ivica Obadic, and Xiao Xiang Zhu.}
\thanks{Adrian Höhl is with the Chair of Data Science in Earth Observation, Technical University of Munich (TUM), 80333 Munich, Germany (email: \href{mailto:adrian.hoehl@tum.de}{adrian.hoehl@tum.de})}
\thanks{Ivica Obadic is with the Chair of Data Science in Earth Observation, Technical University of Munich (TUM) and the Munich Center for Machine Learning, 80333 Munich, Germany (email: \href{mailto:}{ivica.obadic@tum.de})}
\thanks{Miguel-\'Angel Fern\'andez-Torres is with the Image Processing Laboratory (IPL), Universitat de València (UV), 46980 Paterna (València), Spain (email: \href{mailto:miguel.a.fernandez@uv.es}{miguel.a.fernandez@uv.es})}
\thanks{Hiba Najjar is with the University of Kaiserslautern-Landau, Germany; German Research Center for Artificial Intelligence (DFKI), Kaiserslautern, Germany (email: \href{mailto:hiba.najjar@dfki.de}{hiba.najjar@dfki.de})}
\thanks{Dario Oliveira is with the School of Applied Mathematics,
Getulio Vargas Foundation, Brazil. (email: \href{mailto:darioaugusto@gmail.com}{darioaugusto@gmail.com})}
\thanks{Zeynep Akata is with the Institute for Explainable Machine Learning at Helmholtz Munich and with the Chair of Interpretable and Reliable Machine Learning, Technical University of Munich, 80333, Germany. (email: \href{mailto:zeynepakata@helmholtz-munich.de}{zeynep.akata@helmholtz-munich.de})}
\thanks{Andreas Dengel is with the University of Kaiserslautern-Landau, Germany; German Research Center for Artificial Intelligence (DFKI), Kaiserslautern, Germany (email: \href{mailto:andreas.dengel@dfki.de}{andreas.dengel@dfki.de})}
\thanks{Xiao Xiang Zhu is with the Chair of Data Science in Earth Observation, Technical University of Munich (TUM) and with the Munich Center for Machine Learning, 80333 Munich, Germany (email: \href{mailto:xiaoxiang.zhu@tum.com}{xiaoxiang.zhu@tum.com})}
}




\maketitle

\begin{abstract}
In recent years, black-box machine learning approaches have become a dominant modeling paradigm for knowledge extraction in remote sensing.
Despite the potential benefits of uncovering the inner workings of these models with explainable AI, a comprehensive overview summarizing the explainable AI methods used and their objectives, findings, and challenges in remote sensing applications is still missing.
In this paper, we address this gap by performing a systematic review to identify the key trends in the field and shed light on novel explainable AI approaches and emerging directions that tackle specific remote sensing challenges. We also reveal the common patterns of explanation interpretation, discuss the extracted scientific insights, and reflect on the approaches used for the evaluation of explainable AI methods. As such, our review provides a complete summary of the state-of-the-art of explainable AI in remote sensing. Further, we give a detailed outlook on the challenges and promising research directions, representing a basis for novel methodological development and a useful starting point for new researchers in the field.
\end{abstract}


\glsresetall
\section{Introduction}
\label{sec:introduction}

\begin{figure}[tbp]
  \centering
  \includegraphics[trim={0.25cm 0.6cm 0.25cm 0.25cm},clip,width=0.48\textwidth]{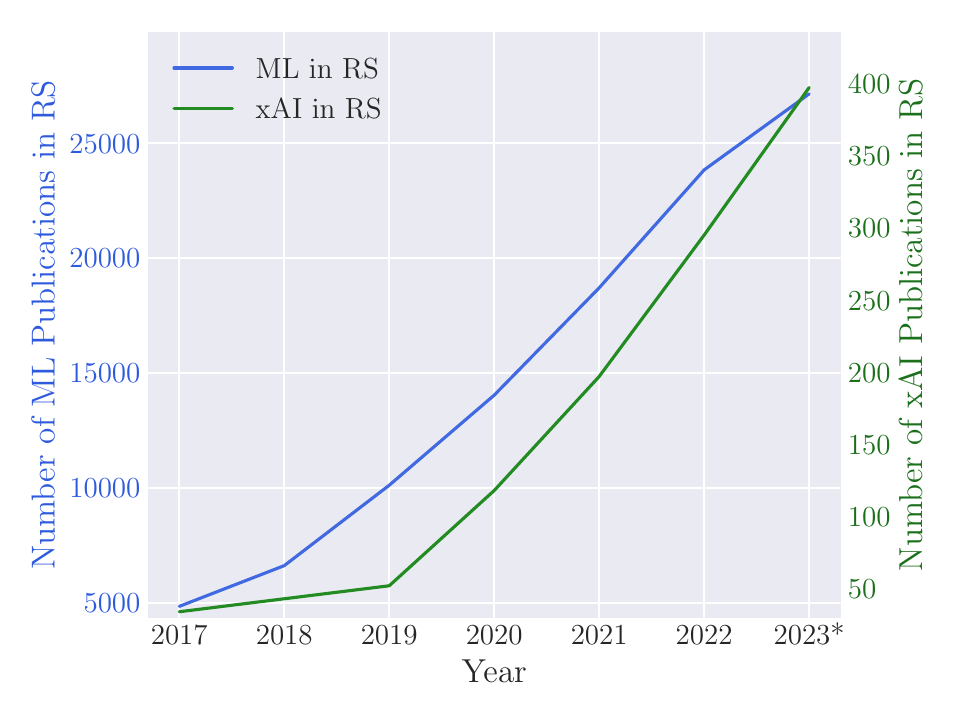}
  \caption{The number of publications of \gls{ml} in \gls{rs} (blue curve) and \gls{xai} in \gls{rs} (green curve), obtained by using the search query described in the Appendix \ref{chap:search_query}, differ by a factor of $\approx70$. (*Calculated amount of publications given the first ten months of the year and assuming a linear trend in 2023.)}
  \label{fig:nr_pubs}
\end{figure}

\IEEEPARstart{M}achine Learning (\gls{ml}) methods have shown outstanding performance in numerous \acrfull{eo} tasks \cite{Camps-Valls2021, Zhu2017}, but mostly they are complex and lack the interpretability and explanation of their decisions.
In \gls{eo} applications, understanding the model's functioning and visualizing the interpretations for analysis is crucial \cite{reichsteinDeepLearning2019}, as it allows practitioners to gain scientific insights, discover biases, assess trustworthiness and fairness for policy decisions, and to debug and improve a model.
The European Union adopted an \gls{ai} Act to ensure that the methods developed and used in Europe align with fundamental rights and values such as safety, privacy, transparency, explicability, as well as social and environmental wellbeing \cite{ruschemeier2023ai}. It is anticipated that other governments worldwide will implement similar regulations \cite{terkonda2023artificial}.
Many applications in \gls{eo} could potentially violate these values when data and \gls{ai} are employed for analysis and decision-making. Nevertheless, \acrfull{xai} can contribute to aligning these practices with rights and laws.
Hence, \gls{xai} emerges as a promising research direction to tackle the above-mentioned scientific and regulatory challenges with observational data \cite{tuiaCollectiveAgenda2021}.

Despite these potential benefits, currently, there is a gap between the usage of \gls{ml} methods in \gls{rs} and the works that aim to reveal the workings of these models. This gap is illustrated in Figure \ref{fig:nr_pubs}, where the blue curve shows that the number of \gls{ml} papers in \gls{rs} have drastically increased in the past few years. \IEEEpubidadjcol Although the works dealing with \gls{xai} in \gls{rs}, shown by the green curve, have also increased rapidly, there is still a gap by a factor of $\approx70$ compared to the number of papers of \gls{ml} in \gls{rs}. This increasing number of \gls{xai} in \gls{rs} papers motivates us to summarize the existing work in the field and provide an overview to \gls{rs} practitioners about the recent developments, which might lead to narrowing this gap and making \gls{xai} approaches more common in the field of \gls{rs}.

\gls{xai} methods are typically designed to work on natural images. However, \gls{rs} images have different properties than natural images \cite{rolfMissionCritical2024}.
First, images are captured from above. This perspective comes with unique scales, resolutions, and shadows. For instance, a \gls{rs} image can cover whole landscapes, thousands of square kilometers, while natural images can only cover a tiny fraction of it \cite{goodchildScaleGIS2011}.
Second, \gls{rs} captures images in other electromagnetic spectra, apart from the usual \gls{RGB} channels. From hyperspectral over \gls{sar} to \gls{LiDAR}, \gls{rs} does cover a wide range of reflectance data.
Third, usual \gls{RGB} cameras are primarily passive, while \gls{rs} can be active, which changes properties like the radar shadows, foreshortening, layover, elevation displacement, and speckle effects \cite{lillesand2015remote}.
Besides the different image properties, the tasks differ as well. \Gls{cv} primarily addresses dynamic scenarios where the objects dynamically move and cover each other. In contrast, the observed processes in \gls{rs} happen on different spectral and spatiotemporal scales, long-term and short-term, and the systems modeled are very complex and diverse. Often, the observed systems are not fully understood and can only be observed indirectly and not completely, for example, weather events, hydrology, hazards, ecosystems, and urban dynamics.
Although there are numerous reviews for \gls{xai} in the literature \cite{rasExplainableDeep2022, minh2022explainable, speith2022review}, they typically do not reflect on the works specific for \gls{rs} nor reveal how the existing \gls{xai} approaches tackle the above challenges related to remote sensing data. Therefore, a review of \gls{xai} tailored to the field of \gls{rs} is necessary to reveal the key trends, common objectives, challenges, and latest developments.

While current reviews of \gls{xai} in \gls{eo} are focused on social, regulatory, and stakeholder perspectives \cite{gevaertExplainableAI2022,leluschkoGoalsStakeholder2023}, specific subtopics in \gls{rs} \cite{hallReviewExplainableAI2022}, or do not provide a broad literature database \cite{xingChallengesIntegrating2023, hallReviewExplainableAI2022, roscherExplainIt2020}, this paper targets the applications and approaches from \gls{xai} in \gls{rs} and follows a systematic approach to gather a comprehensive literature database.
To this end, we conduct a systematic literature search for \gls{xai} in \gls{rs} in three commonly used literature databases in \gls{rs}, namely IEEE, Scopus, and Springer (the research method is thoroughly outlined in Appendix \ref{sec:method}). In parallel, we propose a categorization of \gls{xai} methods, which provides a detailed overview and understanding of the \gls{xai} taxonomy and techniques. We identify the relevant papers from the literature database and rely on the proposed categorization of the \gls{xai} methods to summarize the typical usages, objectives, and new approaches in the field. 
Furthermore, we discuss the alignment of the usage of \gls{xai} in \gls{rs} with standard practices in \gls{xai} and the evaluation of \gls{xai} in \gls{rs}. As such, this review aims to assist users in the field of \gls{eo} for the usage of \gls{xai}. Finally, we identified the challenges and limitations of \gls{xai} for addressing the unique properties of \gls{rs} data, the interpretability of \gls{dl} models along with the lack of labels in \gls{rs}, as well as the combination of \gls{xai} with related fields such as uncertainty or physics. These challenges extend the initial insights presented in \cite{hoehl_obadic_cvpr_2024} and are more extensively explored here.

In particular, we attempt to answer the following research questions: (RQ1) Which explainable \gls{ai} approaches have been used and which methods have been developed in the literature for \gls{eo} tasks?
(RQ2) How are \gls{xai} explanations analyzed, interpreted, and evaluated?
(RQ3) What are the objectives and findings of using xAI in RS?
(RQ4) How do the utilized \gls{xai} approaches in \gls{rs} align with the recommended practices in the field of \gls{xai}?
(RQ5) What are the limitations, challenges, and new developments of \gls{xai} in \gls{rs}?

Therefore, this review provides the following main contributions:
\begin{itemize}
    \item First, we present an overview and categorization of current \gls{xai} methods (Section \ref{sec:background}).
    \item Second, we summarize the \gls{sota} \gls{xai} approaches in \gls{rs} through the analysis of a comprehensive literature database (Section \ref{sec:results} - RQ1 and RQ2).
    \item Third, we identify objectives and practices for evaluating \gls{xai} methods in \gls{rs} (Section \ref{sec:results} - RQ3).
    \item Finally, we discuss challenges, limitations, and future directions for \gls{xai} in \gls{rs} (Section \ref{sec:discussion} - RQ4 and RQ5).
\end{itemize}

\section{Related Work}
\label{sec:related_work}
\begin{figure*}
 \centering
 \includegraphics[width=\textwidth]{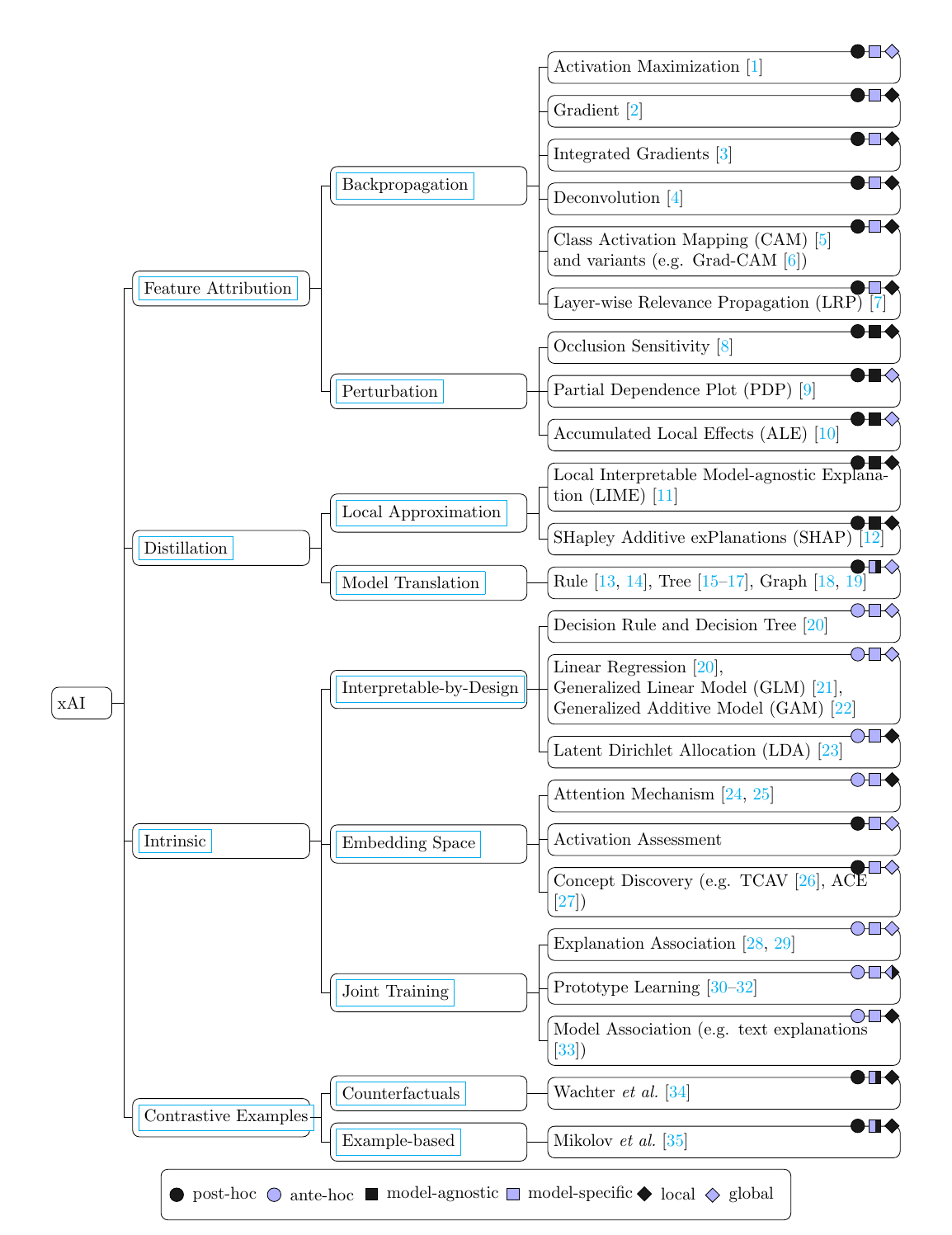}
  \caption{Categorization of \gls{xai} methods based on \citeauthor{rasExplainableDeep2022} \cite{rasExplainableDeep2022}}
  \label{fig:xai_fam}
\end{figure*}
\noindent Numerous resources for either the field of \gls{xai} \cite{Roscher2020, rasExplainableDeep2022, Molnar2019} or \gls{ml} for \gls{eo} applications \cite{Camps-Valls2021, Karpatne2019} are available in the literature.
In contrast, to the best of our knowledge, there are only two reviews \cite{gevaertExplainableAI2022, roscherExplainIt2020} in the overlapping area of these two fields. \citeauthor{gevaertExplainableAI2022} aims to summarize the existing works of \gls{xai} in \gls{eo} and addresses the \gls{xai} usage from a regulatory and societal perspective, discussing the requirements and type of \gls{xai} that is needed in \gls{eo} from policy, regulation, and politics \cite{gevaertExplainableAI2022}.
The work of \citeauthor{roscherExplainIt2020} categorizes the identified works in \gls{xai} according to the general challenges in the bio- and geosciences \cite{roscherExplainIt2020}. Their categorization of \gls{xai} properties and their emphasis on considering expert knowledge constitute two highlights of this review. Furthermore, the presented challenges are still faced by researchers today. However, neither of these reviews uses a broad literature database necessary to provide a comprehensive overview of current \gls{xai} approaches in \gls{eo}.
There also exist \gls{xai} reviews for specific \gls{eo} tasks or perspectives \cite{leluschkoGoalsStakeholder2023, hallReviewExplainableAI2022, xingChallengesIntegrating2023, mamalakis_explainable_2022}. 
In detail, \citeauthor{leluschkoGoalsStakeholder2023} conduct a review on the stakeholders and goals within human-centered \gls{xai} applied in \gls{rs} \cite{leluschkoGoalsStakeholder2023}. Their findings indicate an underrepresentation of non-developer stakeholders in this area. \citeauthor{hallReviewExplainableAI2022} \cite{hallReviewExplainableAI2022} review \gls{dl} methods and investigate to which degree the methods can explain human wealth or poverty from satellite imagery. Next, \citeauthor{xingChallengesIntegrating2023} \cite{xingChallengesIntegrating2023} focus on \gls{xai} in conjunction with \glspl{dnn} that incorporate geographic structures and knowledge. They discuss three challenges when applying \gls{xai} to geo-referenced data: challenges from \gls{xai}, geospatial \gls{ai}, and geosocial applications. Based on a short use case on land use classification and relying on \gls{shap} explanations, they show that the geometry, topology, scale, and localization are of great importance.
Finally, a group at Colorado State University published a survey of their work using \gls{xai} for climate and weather forecasting \cite{mamalakis_explainable_2022}. While all of these studies shed light on the \gls{xai} practices on specific \gls{eo} applications, they do not comprehensively summarize the work done in the broad research field of \gls{xai} in \gls{eo}.

To overcome these shortcomings, we approach \gls{xai} in \gls{eo} systematically and provide an extensive literature database, resulting in a comprehensive summary of the current literature. This work is characterized by our interest in the application and usage of \gls{xai} techniques on \gls{rs} data, while others focus on the application of geological features, natural sciences, or social implications.
Not only do we present an overview of the current challenges in the field, but we also highlight the state-of-the-art methods for tackling these challenges. We believe this could provide valuable insight into current limitations faced in the field.
Compared to the existing literature, we look at the topic from a technical perspective without reflecting on regulatory or ethical implications originating from integrating \gls{xai} into the field of \gls{eo}. Because of the orthogonal approaches in physics-aware ML, uncertainty quantification, or causal inference, this review excludes works from related and overlapping domains. Instead, we refer the reader to overviews in these fields: \cite{karniadakis_physics-informed_2021}, \cite{gawlikowski_survey_2022}, and \cite{perez-suayCausalInference2019, rungeInferringCausationTime2019}, respectively.
\section{Explainable AI Methods in Machine Learning}
\label{sec:background}

\noindent This chapter provides a general overview of \gls{xai}. We first present the taxonomy used to describe common distinctions between explanation methods. Subsequently, we introduce our categorization of \gls{xai} methods. Furthermore, we describe in detail the commonly used methods in the field of \gls{rs} in Appendix \ref{ssec:prom_xai_methods}. Finally, we give an overview of the different metrics proposed in the literature for evaluating these methods and present the main objectives of using \gls{xai}. There exist several terms for \acrlong{xai}, such as \gls{iml} and interpretable \acrshort{ai}. These terms often refer to the same concept: the explanation or interpretation of \gls{ai} models \cite{adadiPeekingBlackBox2018} and we will use them interchangeably.

In the literature, three common distinctions exist when categorizing \gls{xai} methods: \emph{ante-hoc} vs. \emph{post-hoc}, \emph{model-agnostic} vs. \emph{model-specific}, and \emph{local} vs. \emph{global} \cite{speith2022review,holzinger2022explainable,samek2021explaining,Roscher2020}. 
The \emph{ante-hoc} and \emph{post-hoc} taxonomy refer to the stage where the explanation is generated. A \gls{xai} method that provides interpretations within or simultaneously with the training process is called ante-hoc. In contrast, a post-hoc method explains the model after the training phase using a separate algorithm.
Further, \emph{model-agnostic} methods have the ability to generate explanations for any model. The method does not access the model's internal state or parameters, and the explanations are created by analyzing the changes in the model's output when modifying its inputs. Opposed to them, \emph{model-specific} methods are exclusively designed for specific architectures and typically have access to the model's inner workings.
Finally, \emph{local} methods explain individual instances and the model's behavior at a particular sample. In contrast, \emph{global} methods explain the model's behavior on the entire dataset.
In practice, local explanations can be leveraged to achieve global explanations. Through aggregation over a set of input instances chosen to represent the dataset, local explanations can provide insights into the general behavior of the inspected model. The aggregation mechanism should be carefully defined since straightforward aggregation on some \gls{xai} methods might lead to erroneous results. Some researchers further investigated the question of finding meaningful aggregation rules of local explanations \cite{lundbergLocalExplanationsGlobal2020, setzuGLocalXLocalGlobal2021}.

\subsection{Categorization of Explainable AI Methods}
\label{ssec:xai_tree}
\begin{figure*}
    \centering
    \includegraphics[width=\textwidth]{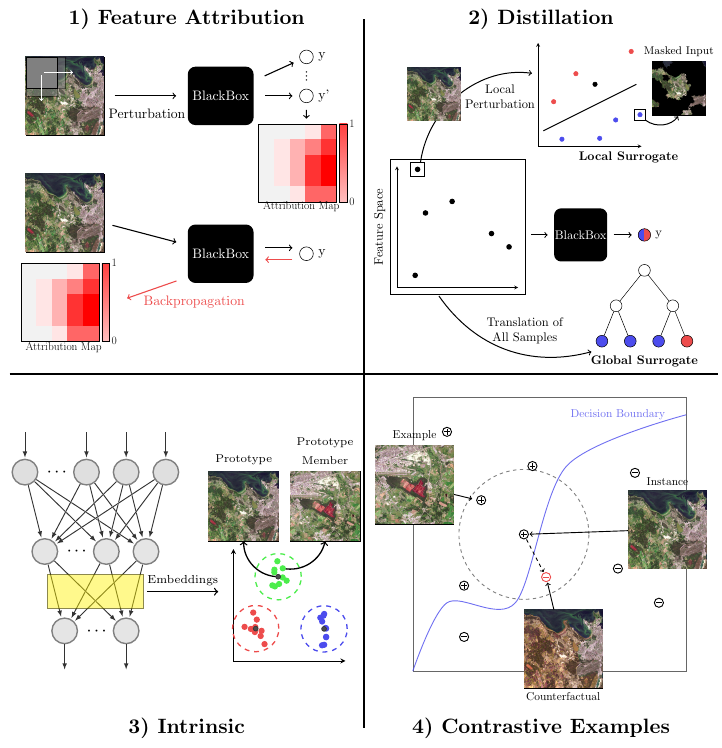}
  \caption{Visualization of representative methods for the four main \gls{xai} categories, given models that receive images as input. (1) Feature attribution: The graph on the top exemplifies perturbation strategies, representing the well-known \emph{Occlusion} method. This method computes the sensitivity of the outputs $y$ to perturbations when sliding a mask over the input image, which generates an attribution map. The second graph illustrates backpropagation methods, which compute gradient-based attribution maps. Attribution maps are normalized to $[0,1]$ for visualization purposes. (2) Distillation: Within this category, we differentiate between local approximation methods (top) and model translation methods (bottom). Local approximation methods provide explanations for each sample by training a local surrogate model (a linear model is shown as an example on the top plot), given perturbed samples. Usually, these methods mask superpixel regions for image-type inputs. Model translation methods rely on global surrogate models that attempt to reproduce the black-box decisions, given all samples. (3) Intrinsic: A joint training method is illustrated in this example. The embedding space of the model is organized into different groups, and the centroids of these clusters constitute prototypes, which can be visually compared with other instances (prototype members). (4) Contrastive examples: Here, we visualize the decision boundary of a model. The graph identifies a counterfactual, which is a sample generated by changing the model decision, and an example-based instance, typically a similar data point from the training set. (Image contains modified Copernicus Sentinel 2 data (2020), processed by ESA.)}
  \label{fig:main_xai_cat}
\end{figure*}

\noindent In this section, we introduce a categorization of the most important \gls{xai} methods in the literature that is presented in Figure \ref{fig:xai_fam}.
We build on the categorical foundation of \cite{rasExplainableDeep2022} and further adjust these categories to capture a larger set of \gls{xai} methods. The categories are structured in a tree-like design. The tree has two internal layers, which describe a hierarchy of primary and secondary categories, and a leaf layer, which indicates an individual or group of specific methods. Additionally, these methods are labeled according to the three universal categories described above. In the following, a high-level overview of the categories of \gls{xai} methods is explained following the structure of Figure \ref{fig:xai_fam}.
Our four primary categories are \emph{feature attribution}, \emph{distillation}, \emph{intrinsic explanations}, and \emph{contrastive examples} and are visualized in Figure \ref{fig:main_xai_cat}. \emph{Feature attribution} methods highlight the input features that significantly influence the output. Alternatively, \emph{distillation} builds a new interpretable model from the behavior of the complex model. \emph{Intrinsic} methods focus on making the model itself or its components inherently interpretable. Lastly, \emph{contrastive examples} concentrate on showing simulated or real examples and allow an explanation by comparing them. 

For the sake of completeness, it should be noted that \emph{feature selection} methods are not considered in our categorization and are listed separately in the results. Even though they are related to feature interpretation, they serve different purposes. Feature selection can be defined as a strategy to reduce the dimensionality of input space to improve the model performance and reduce its computational cost \cite{liuFeatureSelection2010}. While feature selection can constitute the first step in a \gls{ml} pipeline, feature interpretation is usually the last step and typically involves more advanced techniques than only looking at the predictive performance. Therefore, feature selection and \gls{xai} can be complementary. On the one hand, feature selection reduces the input space that needs to be interpreted. On the other hand, \gls{xai} can provide more qualitative insights for selection, such as uncovering a bias introduced by a feature.

\vspace{0.25cm}
\subsubsection{Feature Attribution} \label{sec:feature_attribution}
Feature attribution methods rely on the trained \gls{ml} model to estimate the importance of the input features. Depending on whether the explanations are generated by inspecting the model internals or by analyzing the changes in the model's output after modifying the input features, the methods in this category are further split into backpropagation and perturbation methods. The output of these methods is commonly a saliency plot which determines the contribution of the input features to the model prediction. In the case of imagery inputs, the output of these methods can be visualized as a heatmap, also called a saliency map, which highlights regions relevant to the model prediction.

\paragraph{Backpropagation}
\label{sec:backpropagation}
These methods leverage the inherent structure of \glspl{dnn} to estimate the relevant features by propagating the output of the network layers to the input features. The majority of these methods compute gradients for this purpose.
The \emph{Deconvolution} method, also known as Deconvolutional \gls{nn} \cite{zeilerAdaptiveDeconvolutional2011}, is designed to reverse the convolutional operations from \glspl{cnn}. Reconstructing the input space from the feature maps of the \gls{cnn} allows visualizing which information was learned and how the input is transformed across different network layers.
Similarly, \emph{\acrfull{lrp}} \cite{lrp_bach2015pixel} calculates relevance scores for individual input features through layerwise backpropagating the neuron's activations from the output, utilizing specialized propagation rules. The scores indicate the significance of the connection between input and output. Various adaptations have been proposed which apply different propagation rules based on the design of the networks \cite{kohlbrenner2020towards, montavon2019layer, Arras2019a}.
In contrast, the \emph{Gradient} or Saliency method \cite{saliency_simonyan2014deep} uses the partial derivative with respect to the input to create the attribution maps.
Rather than computing the gradient once, \emph{Integrated Gradients} \cite{sundararajanAxiomaticAttribution2017} calculates the integral of gradients with respect to the input features along an interpolation or path defined between a baseline input and the instance to be explained.
\emph{\gls{cam}} \cite{zhou2016learning} visually explains \glspl{cnn} through attribution heatmaps by introducing a global pooling layer right before the top fully connected one. Using the weights of the latter layer for a particular class, the heatmap is generated by computing the weighted average of the activation maps in the last convolutional layer before being upsampled to match the size of the input tensor for explainability purposes. One extension of \gls{cam} is \gls{gradcam} \cite{selvaraju2017grad}, which replaces these weights by the gradient of the output with respect to the last convolutional layer, thus removing the original requirement of a final global pooling layer.

\paragraph{Perturbation}
\label{sec:perturbation}
The perturbation methods assess feature importance by measuring the sensitivity of model predictions to changes in the input features.
These methods are distinguished by how the features are perturbed. Among others, perturbations include blurring, averaging, shuffling, or adding noise. For example, the \emph{Occlusion} method \cite{occ_zeiler2014visualizing} tries to remove a feature by occluding the features with a neutral value. \emph{\gls{pfi}} \cite{Fisher2019} permutes the features along their dimension, destroying the original relationship between input and output values.
The \emph{\gls{pdp}} \cite{friedman2001greedy} method is designed to show the average influence of a single input feature on the decision while marginalizing the remaining features, which are fixed. Therefore, it assumes feature independence. A similar approach called \emph{\gls{ale}} \cite{apley2020visualizing} can handle correlated features by averaging over the conditional distribution.

\vspace{0.25cm}
\subsubsection{Model Distillation}
\label{sec:distillation}
Model distillation methods approximate the predictive behavior of a complex model by training a simpler surrogate model that is usually interpretable-by-design. By replicating the predictions of the complex model, the surrogate model offers hypotheses about the relevant features and the correlations learned by the complex model without providing further insights into its internal decision mechanism. Distillation approaches are categorized into a) local approximation methods, which train the surrogate model in a small neighborhood around an individual local example, and b) model translation methods, which replicate the behavior of the complex model over the entire dataset.

\paragraph{Local Approximation} 
\label{sec:local_approximation}
Approaches in this category focus on explaining individual predictions of the complex model by inspecting a small neighborhood around the instances to be explained. In contrast to the backpropagation and perturbation methods, which operate on the raw input features, the local approximation approaches transform the input features into a simplified representation space, such as superpixels for imagery inputs. A prominent approach in this category is \emph{\gls{lime}} \cite{ribeiroWhyShouldTrust2016}. It creates a new dataset in the neighborhood of the target instance by perturbing its simplified representation. Next, an interpretable surrogate model is trained to approximate the predictions of the complex model on this newly created dataset. Hence, the explanation for the complex model is distilled to the interpretation of the surrogate model. 
A similar strategy is employed by the \emph{\gls{shap}} framework \cite{lundbergUnifiedApproach2017}. Concretely, \citeauthor{lundbergUnifiedApproach2017} \cite{lundbergUnifiedApproach2017} introduce Kernel SHAP, which utilizes the \gls{lime}'s framework under specific constraints to obtain the feature importance by approximating their Shapley values, a method grounded in game theory for estimating the player's contribution in cooperative games \cite{shapley:book1952}. 
While Kernel SHAP is a model-agnostic approach, Shapley values can also be approximated with model-specific approaches such as Deep SHAP \cite{lundbergUnifiedApproach2017} for neural networks and Tree SHAP \cite{lundberg2018consistent} which enables fast approximation of these values for tree-based models.
     
\paragraph{Model Translation}
\label{sec:model_translation}
These methods approximate the model's decisions on the entire dataset with a simple global surrogate model. Typically, the interpretable-by-design methods summarized in the next section are used as surrogate models, such as rule-based \cite{augastaReverseEngineering2012, zhouMedicalDiagnosis2003}, tree-based \cite{frosst2017distilling, liuImprovingInterpretability2018} or graph-based \cite{zhang2018interpreting, alvarez-melisCausalFramework2017}.

\vspace{0.25cm}
\subsubsection{Intrinsic}
\label{sec:intrinsic}
Intrinsically interpretable \gls{ml} models provide an explanation by themselves based on their structure, components, parameters, or outputs. Alternatively, a human-interpretable explanation can be obtained by visualizing them.

\paragraph{Interpretable-by-Design} \label{sec:interpretable_design} These methods are interpretable by humans because of their simplicity in design, architecture, and decision process.
\emph{Decision Rules} are hierarchical IF-THEN statements, assessing conditions and determining a decision. Fuzzy rules \cite{bouchon-meunierLearningFuzzy1999} are designed to address uncertainty and imprecision, frequently encountered in nature. While classical precise rules struggle to represent this uncertainty, fuzzy rules incorporate it, e.g., by using partial membership of classes (fuzzy sets) \cite{zadeh1965fuzzy}. Due to their proximity to natural language, they are interpretable by humans.
The \emph{Decision Tree} \cite{breiman1984classification} greedily learns decision rules. Their internal structure is a binary tree, where each internal node is a condition and each leaf is a decision.

\emph{\acrfullpl{gam}} \cite{hastie1992generalized} are statistical modeling techniques that approximate the response variable as a sum of smooth, non-linear functions transforming the input features.
On the other hand, \emph{\acrfullpl{glm}} \cite{annette2018introduction} considers a linear relationship defined through a specific distribution. Each function becomes a coefficient and the response variable is now computed as a weighted sum of features, which allows representing the mean of various exponential-family distributions. A simple example of this model type is \emph{\gls{lr}}, which assumes a Gaussian data distribution where the response variable is the identity. In this context, the model explanation can be obtained by examining the coefficients.

A well-known approach for generative probabilistic topic modeling is \emph{\acrfull{lda}} \cite{blei2003latent}. A dataset is assumed to be organized in corpora or collections, and each collection contains discrete units, such as documents comprised of words. A distribution of these units characterizes both the collections and the topics. By analyzing the proportions of the units in the collections, \acrshort{lda} estimates the underlying topics. The decisions of the \gls{lda} model and the identified topics can be interpreted by examining the predicted proportions for each collection.
    
\paragraph{Embedding Space}
\label{sec:embedding_space}
These approaches process the activations in the latent space of \glspl{dnn} to interpret its workings. 
The \emph{attention mechanism} \cite{bahdanau2014neural, vaswani2017attention} creates high-level feature representations by using the attention weights to model the dependencies between the different elements in the input. Hence, visualizing the attention weights is a common procedure to assess the relevant features for the model decisions.
\emph{Activation assessment} analyzes the activations in the latent space of a \gls{nn} based on projection techniques. Commonly used are dimensionality reduction approaches \cite{lipton2018mythos} (e.g., \gls{tsne} \cite{van2008visualizing} and \gls{umap} \cite{mcinnes2018umap})  or neuron receptive fields in \glspl{cnn} \cite{brendelApproximatingCNNs2019}.
\emph{Concept-based explanations} summarize the activations in the latent space in terms of interpretable, high-level concepts. The concepts represent visual patterns in computer vision, and they are either user-defined in \gls{tcav} \cite{kim2018interpretability} or learned in an unsupervised manner \cite{ghorbani2019towards}. These approaches provide global explanations by quantifying the concept relevance per class. 
    
\paragraph{Joint Training} \label{sec:joint_training} This category provides ante-hoc explanations by introducing an additional learning task to the model. This task is jointly optimized with the original learning objective and is used as an explanation. The methods in this family typically differ based on the explanation representation and how the additional task is integrated with the original model. 
    
\emph{Explanation association} methods impose the inference of a black-box model to rely upon human-interpretable concepts. A prominent approach in this category are the concept bottleneck models that represent the concepts as neurons in the latent space of the model. They first introduce an additional task of predicting the interpretable concepts in an intermediate layer of a \gls{dl} model. Then, the model predictions are derived based on these concepts.
During training, a regularization term is added to the loss function that enforces alignment of the latent space according to the interpretable concepts \cite{koh2020concept}. To estimate the concept importance directly, \citeauthor{marcos2019semantically} predict the final output by linearly combining the concept activation maps \cite{marcos2019semantically}.

\emph{Prototype learning} approaches aim to identify a set of representative examples (prototypes) from the dataset and provide an interpretable decision mechanism by decomposing the model predictions based on the instance's similarity with the learned prototypes \cite{bien2011prototype}. Thus, visualizing the prototypes enables global model interpretability, while the similarity with an input instance offers local model explanations.
One popular approach for prototype learning on image classification tasks is the ProtoPNet architecture introduced by \citeauthor{chen2019looks} \cite{chen2019looks}. The prototypes represent image parts and are encoded as convolutional filters in a prototype layer of the proposed network. Their weights are optimized with the supervised learning loss of the network and additional constraints that ensure both the clustering of the prototypes according to their class and the separability from the other classes. One extension of this approach is the Neural Prototype Trees \cite{nauta2021neural}, which organizes the prototypes as nodes in a binary decision tree. Each node computes the similarity of the corresponding prototype with an instance. These similarities are used to route the instance towards the leaves of the tree containing the class predictions.

In contrast to the previous approaches, which encode the explanation within the model that performs inference for the original learning task, the \emph{model association} methods introduce an external model that generates explanations. These approaches are often utilized to provide textual explanations for \gls{cv} tasks. An example of such an approach is presented by \citeauthor{kim2018textual}, who derive text explanations for self-driving cars based on jointly training a vehicle controller and a textual explanation generator \cite{kim2018textual}. The vehicle controller is a \gls{cnn} model that recognizes the car's movements with spatial attention maps. Next, the explanation generator, which is a \gls{lstm} model, processes the context vectors and the spatial attention maps from the controller to produce the text explanation.

\vspace{0.25cm}
\subsubsection{Contrastive Examples} \label{sec:contrastive_examples}
Methods within this category provide alternative examples to an input instance and allow obtaining an explanation by comparing them. Usually, examples that are close to each other in the input space yet lead to a different outcome than the original input instance are shown.
\paragraph{Counterfactuals} \label{sec:counterfactuals}
This explanation type aims to discover the smallest change required for an instance to achieve a predefined prediction. Essentially, they answer the question, "Why does it yield output X rather than output Y?" which is very close to human reasoning \cite{millerContrastiveExplanation2021}. They have a close proximity to adversarial examples, although their objectives differ significantly. Adversarial examples usually want to achieve a confident prediction with a minimal perturbed instance, whose change should remain imperceptible for humans. Conversely, counterfactuals aim to provide a diverse set of examples and should allow representing the decision boundary of the model.
\citeauthor{wachterCounterfactualExplanations2017} \cite{wachterCounterfactualExplanations2017} introduce an optimization problem whose primary objective is to find a counterfactual that is as close as possible to the original input. As such, the distance function between the counterfactual and the original input should be minimized. 
\paragraph{Example-based Explanations} \label{sec:example_based}
Unlike counterfactuals, which can generate artificial instances, example-based explanations usually present existing "historical" training instances and showcase similar instances to the input under consideration \cite{lipton2018mythos}. The user can connect, correlate, and reason based on the analogies. The explanatory approach aligns with case-based interpretable-by-design model explanations, e.g., \glspl{knn}.
For example, \citeauthor{mikolovDistributedRepresentations2013} \cite{mikolovDistributedRepresentations2013} train a skip-gram model and evaluate the model using the nearest neighbors determined by the distances in the embedded space. Furthermore, they illustrate that the acquired word representations have a linear relationship, allowing for the computation of analogies through vector addition.

\subsection{Evaluation of Explainable AI Methods}
\label{sec:background_xai_methods_evaluation}
\noindent Evaluating explanation quality and its trustworthiness is an essential methodological challenge in \gls{xai}, which has received considerable attention in recent years. The existing evaluation strategies can be categorized into (1) functional approaches based on quantitative metrics and (2) user studies \cite{doshi2017towards}.

\vspace{0.25cm}
\subsubsection{Functional Evaluation Metrics}
\label{sec:func_eval}
Quantitatively evaluating explanations poses a challenge because of the lack of ground truth explanations against which the generated explanations can be compared. The numerous functional evaluation metrics typically assess different aspects of the explanation quality by quantitatively describing to which extent it satisfies a certain set of desired properties. \citeauthor{MeikeFromAnecdotal2023} propose to categorize the 12 essential explanation quality properties into three main groups and suggest to evaluate as many as possible of these properties for a comprehensive quantitative explanation assessment \cite{MeikeFromAnecdotal2023}. The first group of properties evaluates the explanation \emph{content}, describing how correct, complete, consistent, or discriminative the explanations are, among other properties. Conversely, properties in the second group, such as \emph{complexity}, which refers to the sparsity of an explanation, or \emph{composition} that can measure whether the explanation can localize the ground-truth region of interest, assess the \emph{presentation} of the explanations. Last but not least, the third group of properties focuses on the \emph{relevance} of explanations for the user and their alignment with the domain knowledge.

\emph{Faithfulness} and \emph{robustness}, included in the first group of properties, are two of the most common evaluation metrics in the literature. The \emph{faithfulness} property (also called correctness) asserts how close an explanation method approximates the actual model workings. Various functional metrics are proposed to evaluate explanation faithfulness. For example, metrics based on randomization tests are introduced in \cite{adebayoSanityChecks2018} to measure the explanation sensitivity to randomization in model weights and label permutation. The results reveal that most of the evaluated backpropagation methods do not pass these tests. Another common approach to evaluate explanation faithfulness is based on the perturbation of the input features. For instance, \cite{ijcai2020p417} measures the changes in the model output after perturbation of the supposedly important features, as estimated by the explanation method. Perturbation-based metrics are also used to evaluate \emph{robustness} (also referred to as explanation sensitivity), which inspect the impact of small changes in the input features on the resulting explanation \cite{hedstromQuantusExplainable2023,yeh2019fidelity}. Explanations with low sensitivity are preferred as this indicates that the explanation is robust to minor variations in the input. However, it is worth noting that perturbation-based metrics might result in examples with different distribution than the instances used for model training, which questions whether the drop in model performance can be attributed to the distribution shift or to the perturbation of the important features. To address this issue, \citeauthor{hooker2019benchmark} \cite{hooker2019benchmark} propose model retraining on a modified dataset where a fraction of the most important features identified by the \gls{xai} method is perturbed. Further, \citeauthor{rong2022consistent} \cite{rong2022consistent} presented an improved evaluation strategy that avoids the need for model retraining based on information theory.

\vspace{0.25cm}
\subsubsection{User Studies} Conversely, experiments in which humans evaluate the quality of explanations can also be conducted. User studies in \cite{doshi2017towards} are further categorized into an application-grounded evaluation and human-grounded evaluation, depending on the evaluation task, the type of participants, and the considered explanation quality criteria. On the one hand, \emph{application-grounded} evaluation studies typically involve domain experts who evaluate the explanation in the context of the learning task. For instance, \citeauthor{suresh2022intuitively} \cite{suresh2022intuitively} measures the physicians' agreement with various explanation methods for the problem of classifying electrocardiogram heartbeats. 
On the other hand, \emph{human-grounded} evaluations are usually conducted by participants who are typically non-domain experts who evaluate more general notions of explanation quality. For example, \citeauthor{alqaraawi2020evaluating} \cite{alqaraawi2020evaluating} evaluate whether explanations help lay users recruited via an online crowdsourcing platform to understand the decisions of a \gls{cnn} model for image classification. For a detailed survey on the user studies conducted for \gls{xai} methods evaluation, we refer the reader to \cite{rong2022towards}.

\subsection{Explainable AI Objectives}
\label{chap:xai_obj}
\noindent In this study, we group the objectives of utilizing \gls{xai} in \gls{rs} according to the following four reasons, as defined in \cite{adadiPeekingBlackBox2018}: (1) explain to justify, (2) explain to control, (3) explain to discover, and (4) explain to improve. \emph{Explain to justify} is motivated by the need to explain individual outcomes, which ensures that the \gls{ml} systems comply with legislations, such as enabling users the "right to explanations". Furthermore, the explanations can enable a detailed understanding of the workings of the \gls{ml} model. Hence, the \emph{explain to control} objective is relevant for assessing model trustworthiness and can help to identify potential errors, biases, and flaws of the \gls{ml} model. These insights can be used to \emph{discover} scientific knowledge and new insights about the underlying process that is modeled with the \gls{ml} system or to further \emph{improve} the existing model. The improvement techniques based on the \gls{xai} insights are classified by \citeauthor{weber2023beyond} into augmenting the (1) input data, (2) intermediate features, (3) loss function, (4) gradient, or (5) \gls{ml} model \cite{weber2023beyond}. For the sake of completeness, we consider adapting existing \gls{xai} methods as another improvement strategy.
 
\section{Explainable AI in Remote Sensing}
\label{sec:results}
\noindent Here we summarize the research on \gls{xai} in \gls{rs} and answer the research questions RQ1, RQ2, and RQ3. First, we bring attention to the common practices and highlight new approaches (RQ1). Next, we summarize the methodologies to understand and evaluate the model explanations (RQ2). Further, we outline the common research questions across the different \gls{rs} tasks that practitioners aim to answer with \gls{xai} (RQ3). 

\subsection{RQ1: Usage/Applications of Explainable AI in Remote Sensing}
\label{sec:xai_patterns_in_rs}
\noindent Table \ref{tab:summary} contains the full list of publications included in our study. The papers are arranged by considering groups for the different \gls{eo} tasks (see  Appendix \ref{sec:glossary}) and follow the \gls{xai} categorization shown in Figure \ref{fig:xai_fam}. The table lists the used \gls{xai} methods, objectives, and evaluation types for each paper. Also, Appendix \ref{sec:3_eo_tasks} describes in detail the methods employed in the three most common \gls{eo} tasks: landcover mapping, agricultural monitoring, and natural hazard monitoring.

\begin{figure*}[htbp]
  \centering
  \includegraphics[trim=.2cm 0cm 0cm .2cm,clip,width=\textwidth]{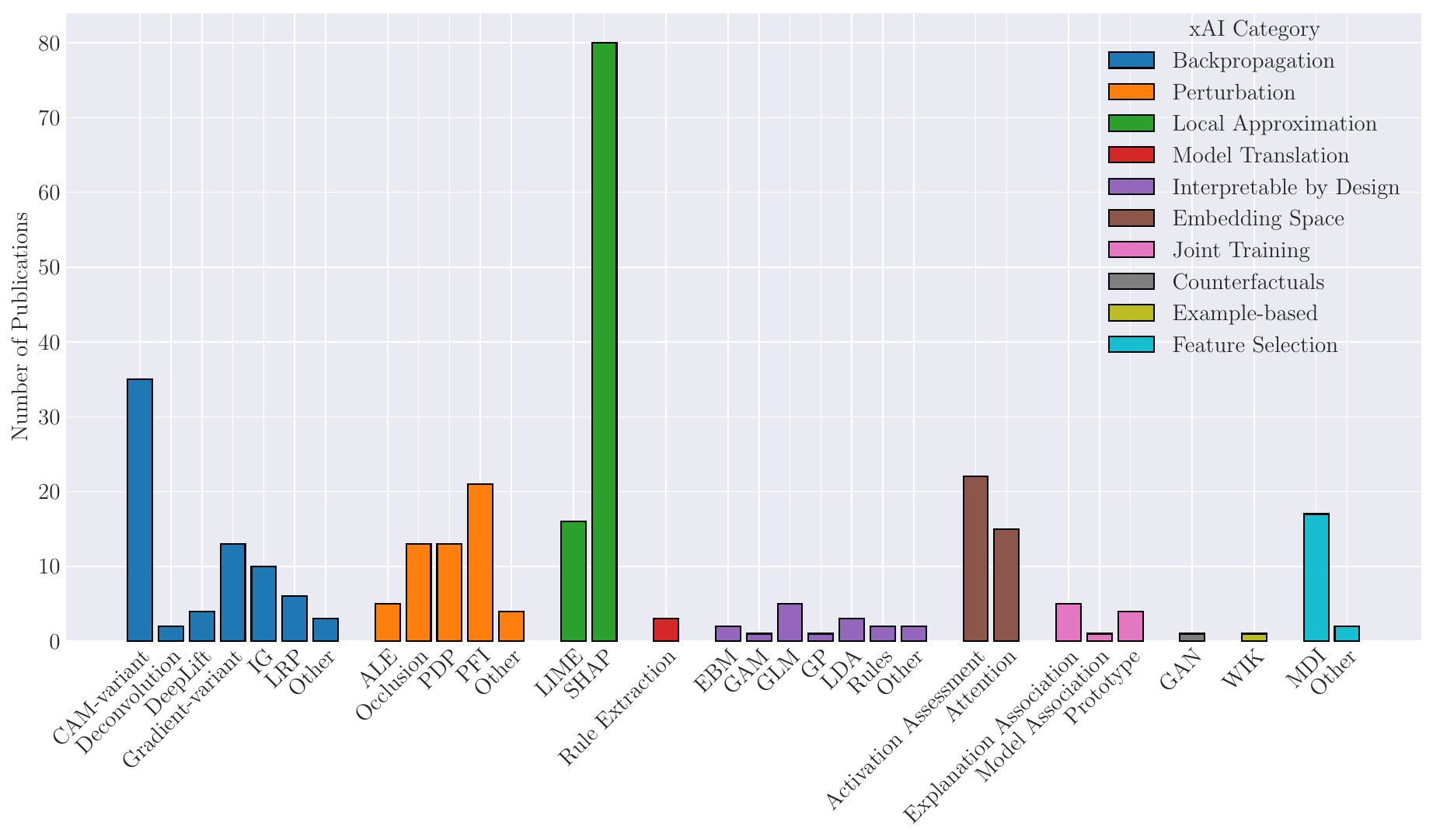}
  \caption{Number of publications per different \gls{xai} methods, grouped according to our categorization in Figure \ref{fig:xai_fam}. Most \gls{rs} studies rely on local approximation approaches, particularly the \gls{shap} method. Second most often backpropagation approaches are used, among which \gls{cam} methods receive the highest focus. Perturbation and embedding space techniques are also prominently used, while the other \gls{xai} categories occur less frequently.}
  \label{fig:xai_in_xaif}
\end{figure*}

\begin{figure}[htbp]
  \centering
  \includegraphics[trim={2cm 1.75cm 2cm 0cm},clip, width=0.5\textwidth]{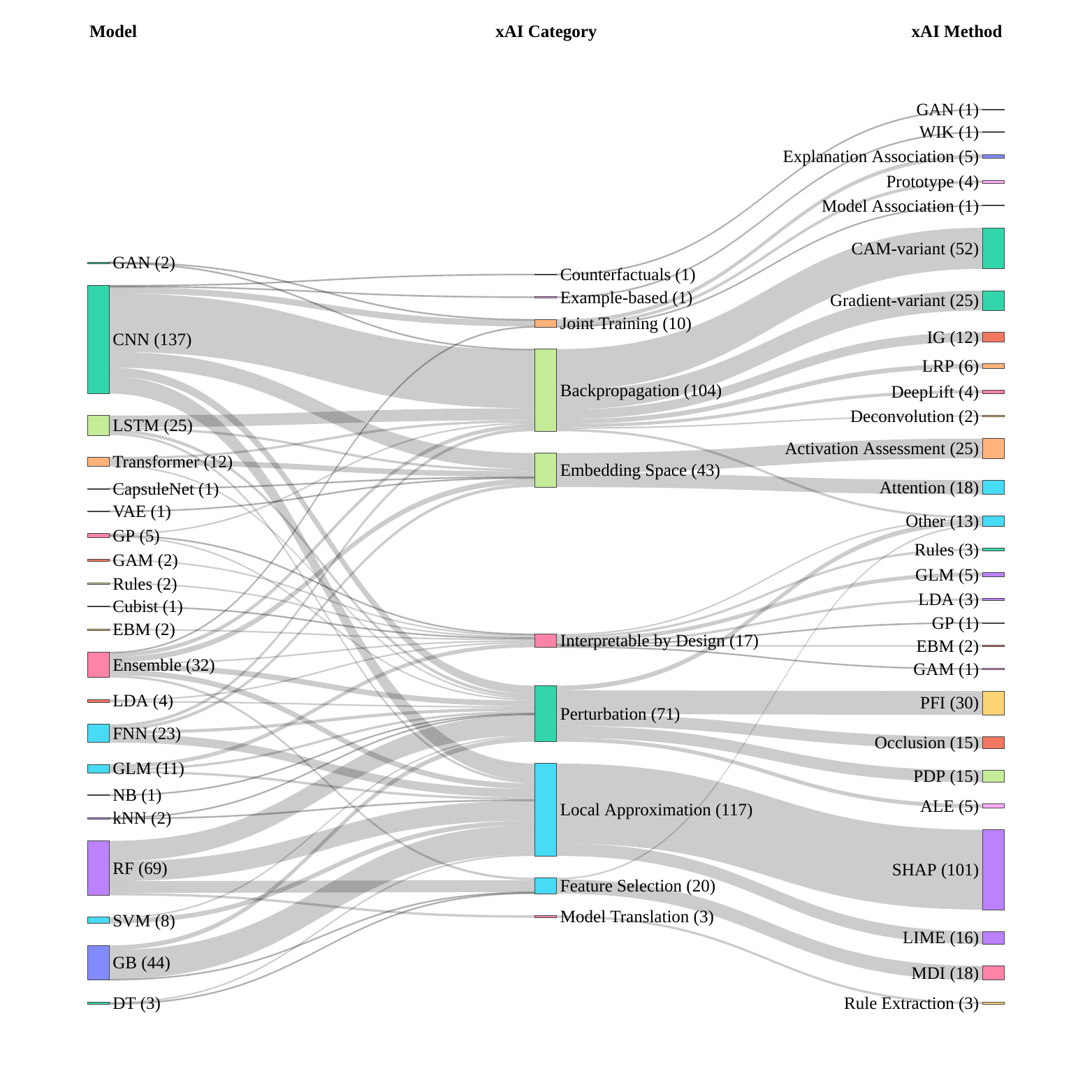}
  \caption{The number of times \gls{xai} categories, methods, and models are mentioned in the identified literature.
  Typically, local approximation and perturbation methods are applied to non \gls{dl} models such as \gls{rf} and \gls{gb}. However, they also find usage in explaining \glspl{cnn}.
  While \gls{cnn} and \gls{lstm} models mostly rely on backpropagation methods, the transformer models are almost exclusively explained with embedding space techniques.}
  \label{fig:xai_models_categories_methods}
\end{figure}

\subsubsection{Application of Explainable AI Methods}
\label{sec:results_application_existing_methods_in_rs}
Figure \ref{fig:xai_in_xaif} shows the number of papers using a \gls{xai} method grouped by the categories introduced, while Figure \ref{fig:xai_models_categories_methods} illustrates all the combinations of model and methods in the literature. Local approximation methods are the most frequently used in over 94 publications. Through their popularity, they are used on the most diverse set of \gls{eo} tasks and models. As shown in Figure \ref{fig:xai_models_categories_methods}, in most cases, they interpret tree-based models (i.e., \gls{rf} and tree ensembles), followed by \glspl{cnn} and \glspl{mlp}. Also, with over 65\%, most of these publications rely solely on local approximation methods without evaluating other methods.
Backpropagation methods follow with a small gap of 72 papers, mainly leveraging \gls{cam} variants. While many \gls{cnn} architectures are interpreted, most time series models, like the \gls{lstm}, are also located here.
In contrast, most papers using transformers are among the 35 publications that leverage embedding space interpretation techniques, which was to be expected since attention is already the centerpiece of the architecture.
Followed by a large gap to the \gls{shap} and \gls{cam} variant methods, 56 publications use various perturbation methods. The number of publications is fairly well proportioned between the methods, but \gls{pfi} is the most widely used, followed by Occlusion and \gls{pdp}.
Further, 15 publications leverage a diverse set of interpretable-by-design models. Although \glspl{glm}, particularly \gls{lr}, are the most common, newer models like the \gls{ebm} are gaining recognition.
Only a few papers employ joint training, like prototypes or explanation associations. Even less popular are model translations, counterfactuals, and example-based explanations.
Furthermore, as shown in \cite{hoehl_obadic_cvpr_2024}, local approximation and perturbation methods follow an increasing trend, and it can be expected that their proportion of publications will further increase while other categories, like backpropagation, stagnate.
Next, the \gls{mdi} or Gini importance is often used in feature selection for global importance measurements and can be easily obtained for tree-based methods \cite{sotomayorSupervisedMachine2023,chenEnhancingLand2023,fisherUncertaintyawareInterpretable2022,jayasingheCausesTea2023,liAdvancingSatellite2021,matinEarthquakeinducedBuildingdamage2021,milaEstimatingDaily2023,mudeleModelingTemporal2020,newmanExplainableMachine2021,orynbaikyzyCropType2020,rubiPerformanceComparison2023,wuEstimateNearsurface2023,xuInterpretingMultitemporal2021,yanSpatialTemporalInterpretable2021,chenEstimationAtmospheric2022,chenEstimationNearsurface2023,chenExploringHighresolution2023}. This shows that \gls{xai} and feature selection have fluent boundaries, as the Gini index is used for feature selection, i.e., when deciding about the depth in the tree with the purity of a split, but also allows for the interpretation of the decision process of the model.
Finally, Figures \ref{fig:xai_in_xaif} and \ref{fig:xai_models_categories_methods} show that interpretable-by-design models are leveraged in diverse contexts. For instance, linear regression supports more complex methods and evaluates the linear trends \cite{nayakInterpretableMachine2022, zhuInvestigatingImpacts2022}. The application of \gls{glm} is observed for small sample sizes \cite{lambertVegetationChange2020} and \glspl{gam} are used for larger datasets \cite{zhouRelativeImportance2021, celikExplainableArtificial2023, maxwellExplainableBoosting2021, fangNewApproach2023}. \citeauthor{zhouRelativeImportance2021} reduce the feature space with \gls{pca} in \cite{zhouRelativeImportance2021}, while advanced versions based on \glspl{ebm} are introduced in \cite{celikExplainableArtificial2023, maxwellExplainableBoosting2021}. \glspl{ebm} use pairwise feature interactions within the \gls{gam} and consider gradient boosting to train each feature function consecutively. \citeauthor{fangNewApproach2023} follows this idea and uses \glspl{fnn} as function approximations \cite{fangNewApproach2023}.
Further, a decision tree incorporating linear regression models at the terminal leaves is employed in \cite{chengImprovedUnderstanding2022} to gain scientific insights into the partitioning of precipitation into evapotranspiration and runoff. In another example, \citeauthor{karmakarFeaturefreeExplainable2021} apply \gls{lda} to \gls{sar} images. Their bag-of-words approach uses superpixels as words to do landcover mapping \cite{karmakarFeaturefreeExplainable2021,karmakarFrameworkInteractive2022}.
Last but not least, \citeauthor{martinez-ferrerCropYield2021} analyze the weights of the \gls{gp} to find anomalous samples \cite{martinez-ferrerCropYield2021}.

Besides the above-described common practices of applying \gls{xai} in \gls{rs}, we also identified distinct modeling approaches, such as concept bottleneck models, fuzzy logic-based models, and integration of \gls{xai} into the training pipeline that are applied to specific \gls{eo} tasks.

In \gls{eo} studies, concept bottleneck models are used to associate model predictions of socioeconomic indicators with human-understandable concepts. In addition to the work of \citeauthor{leveringRelationLandscape2021} \cite{leveringRelationLandscape2021} (also pointed in  \cite{gevaertExplainableAI2022}) that uses landcover classes as interpretable concepts to explain landscape aesthetics, a similar approach is also used in two recent studies \cite{ leveringPredictingLiveability2023, scepanovicJaneJacobs2021}, 
Concretely, in \cite{leveringPredictingLiveability2023}, the same authors propose a semantic bottleneck model for estimating the living quality from aerial images with the interpretable concepts capturing population statistics, building quality, physical environment, safety, and access to amenities. The other work of \citeauthor{scepanovicJaneJacobs2021} estimates the vitality of Italian cities by relying on vitality proxies such as land use, building characteristics, and activity density \cite{scepanovicJaneJacobs2021}.

Fuzzy logic-based models are another approach that is mainly used to evaluate the trustworthiness of \gls{ml} models. An \gls{owa} fusion function is presented in \cite{stroppianaFullyAutomatic2021} for burned area mapping, which allows controlling if the fusion results are affected by more false positives than false negatives, and vice versa. At the same time, it foresees if there are only a few highly or many low relevant factors when providing a particular output. The outputs of \glspl{fls} for tree monitoring \cite{leon-garzaBigBangbig2020} can also be easily validated. Last but not least, measure-, integral- and data-centric indices based on the Choquet integral (an aggregation function defined with respect to the fuzzy measure) are introduced in \cite{murrayExplainableAI2018, islamEnablingExplainable2020, murrayInformationFusion2text2020, andersonFuzzyChoquet2018} intending to develop more understandable ensembles for landcover mapping.

Lastly, some works explore integrating \gls{xai} into the training pipeline \cite{liDeepNetworks2018,xiongInterpretableFusion2022}. \citeauthor{xiongInterpretableFusion2022} \cite{xiongInterpretableFusion2022} leverage \gls{gradcam} to create masks that occlude features the network has emphasized, encouraging the network to exploit other features \cite{xiongInterpretableFusion2022}.
The output of three \glspl{nn} is merged through attention in another study \cite{chenAttributionDeep2023}. There, the outputs and the attributions of the \gls{deeplift} backpropagation method are the key values of the attention layer.
In contrast, \citeauthor{liDeepNetworks2018} \cite{liDeepNetworks2018} directly classify objects on top of the \gls{cam} attribution map. Hence, their weakly supervised method does not need bounding boxes, making the image labels sufficient.

\subsubsection{Adapted Explainable AI Approaches}
\label{sec:results_adapted_approaches}

\begin{figure}[!t]
    \centering
    \includegraphics[trim=0.1cm .8cm 0cm 0cm,clip,width=0.5\textwidth]{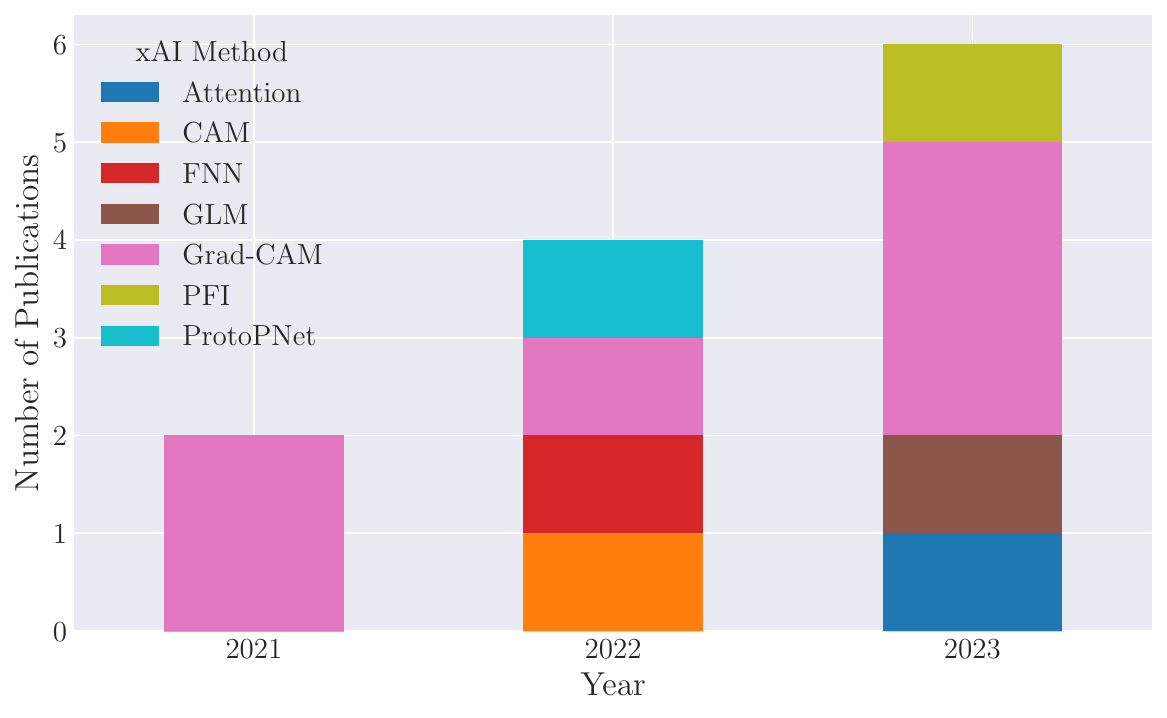}
    \caption{Number of papers adapting existing \gls{xai} methods, grouped per year. The existing \gls{xai} methods are increasingly adapted to address the \gls{rs} challenges with the highest attention given to \gls{cam} techniques.}
    \label{fig:xai_derived}
\end{figure}

As described in Section \ref{sec:introduction},  the popular \gls{xai} methods are designed initially to work on natural images, which significantly differ from remote sensing acquisitions. This raises the question of whether the utilized \gls{xai} methods fit remote sensing data well. In this respect, we identified several works that propose new approaches considering remote sensing data properties to produce better explainability insights. Figure \ref{fig:xai_derived} illustrates that recently, there has been an increase in such approaches, with 50\% of the publications being published in the last year, and no novel method was identified before 2021. These approaches typically adapt the existing \gls{xai} methods with a particular focus on the \gls{cam} and \gls{gradcam} methods or propose new \gls{dl} architectures.
For example, \citeauthor{fengSelfMatchingCAM2021} exploit that the target objects in \gls{sar} images occupy only a small portion of the image to propose a new \gls{cam} method which, instead of upsampling the feature map of the convolutional layer to the input image, downsamples the input image to the feature map of the last convolutional layer \cite{fengSelfMatchingCAM2021}. This operation results in saliency maps that localize precisely the targets in SAR images compared to the \gls{gradcam} method.
Additionally, a \gls{cam} method able to produce much more fine-grained saliency maps than the prior \gls{cam} methods is introduced by \citeauthor{guoVisualExplanations2023} in \cite{guoVisualExplanations2023}. Similar to Layer-CAM \cite{jiangLayerCAMExploring2021}, they use shallow layers to get more fine-grained results but also rely on scores, following the idea of Score-CAM \cite{wangScoreCAMScoreWeighted2020}, which are not as noisy as gradients. 
Another attempt to improve current \gls{cam} methods was proposed by \citeauthor{marvasti-zadehCrownCAMInterpretable2023}, who utilize the attribution maps from all network layers and decrease their number through only retaining maps which minimize the information loss according to the \gls{kl}-divergence \cite{marvasti-zadehCrownCAMInterpretable2023}. Additionally, local attribution maps are generated by masking the image and weighting the maps by the corresponding bounding box and their prediction confidence. These local maps must be smoothed with a Gaussian kernel to avoid sharp boundaries in the resulting \gls{cam}. This novel approach, called Crown-CAM, is evaluated on a localization metric and outperforms (augmented) Score-CAM and Eigen-CAM on a tree crown localization task. 
\gls{cam} variants for hyperspectral images are developed in \cite{deluciaExplainableAI2022}. The saliency map is now a 3D volume instead of a 2D image, and each voxel attributes the different channels in depth, which provides pixel-wise and spectral-cumulative attributions.
Other \gls{gradcam} adaptions proposed are a median pooling \cite{songMedianpoolingGradCAM2021} and a pixel-wise \cite{geInterpretableDeep2023} variant.

When it comes to new \gls{dl} approaches, a model prototype approach for \gls{rs} is proposed in \cite{barnesThisLooks2022}, where the ProtoPNet architecture \cite{chen2019looks} is adapted to also consider the location of the features. The network is iteratively trained in 3 stages. Firstly, the encoder and prototype layers are trained to produce the prototypes. Secondly, the prototypes are replaced by the nearest prototype of the corresponding class. Lastly, only the output layer weights are trained to produce the final prediction. In contrast to ProtoPNet, the prototype similarity is scaled with a location value learned by the network. This acknowledges the location of the prototypes in the image and makes them location-aware.
Another approach is presented in \cite{huangBetterVisual2022}, where a reconstruction objective is added to the loss function to enable the \gls{gradcam}++ method to more accurately localize multiple target objects within an aerial image scene.

Finally, we identified one approach that addressed model agnostic methods. Specifically, \citeauthor{brenningSpatialMachinelearning2023} adapts the model-agnostic \gls{pfi} method to incorporate spatial distances \cite{brenningSpatialMachinelearning2023}. Analogous to the \gls{pfi} method, the features are permuted, and the mean decrease in predictive accuracy is assessed. Notably, features are permuted across various predefined distances, revealing the spatial importance or sensitivity of the model.

\subsection{RQ2: Interpretation and Evaluation of Explanations in Remote Sensing}
\subsubsection{Understanding and Validating Explanations}
As mentioned in Section \ref{sec:introduction}, remotely sensed data usually depicts complex relationships, which can hinder the intuitive understanding of the semantics of the explanations.
Therefore, an obstacle when applying \gls{xai} in \gls{rs} is explanation interpretation, as the relevant features often do not have a straightforward interpretation. We identify that this challenge is frequently tackled by transforming the raw features into interpretable features used for model training \cite{andresiniSILVIAEXplainable2023, mateo-sanchisInterpretableLongShort2023} or by associating domain knowledge with the explanation at the post-hoc stage \cite{obadicExploringSelfAttention2022, xuInterpretingMultitemporal2021}.

\paragraph{Creation of Interpretable Features}
Ensuring human-understandable features is essential for comprehending input-output relationships or gaining knowledge of the \gls{ml} model. Our study identified that these features are typically derived with spectral indices or dimensionality reduction techniques. For optical imagery, many works utilize standard spectral indices such as \gls{ndvi} to create interpretable features. Further, the problem of creating interpretable features from \gls{sar} images is tackled by \citeauthor{geInterpretableDeep2023} in \cite{geInterpretableDeep2023} by transforming the \gls{sar} pixels into human-interpretable factors using the U-Net architecture. In detail, they derive three interpretable variables from the VH polarization backscatter coefficients and the VV polarization interferometric coherence of the Sentinel-1 images, providing insights into the temporal variance and minimum. While the temporal variance changes between different crops and landforms over time, the temporal minimum is specific for flooded rice fields due to their proximity to water. This facilitates the understanding of the attribution of the applied \gls{gradcam} method.
Dimensionality reduction techniques can also help derive an interpretable representation from a complex and correlated feature space. This is demonstrated by 
\citeauthor{brenningInterpretingMachinelearning2023} who employs structured \gls{pca} for feature space reduction and \gls{rf} for classification \cite{brenningInterpretingMachinelearning2023}.
His \gls{xai} analysis reveals that the behavior of the features can be identified in the principal components and allows the extraction of the relationship between the main feature groups.

\paragraph{Interpreting Explanations with Domain Knowledge}
Domain knowledge is often utilized to reveal the semantics of the relevant features from raw inputs. It is usually derived from already established indicators or based on expert knowledge. 

For instance, indicators such as \gls{ndvi} describing vegetation phenology are commonly used in agriculture monitoring. They are typically related to the relevant time steps identified with a \gls{xai} approach to reveal the critical phenological events for crop disambiguation \cite{xuInterpretingMultitemporal2021, obadicExploringSelfAttention2022, orynbaikyzyCropType2020} or yield prediction \cite{huberExtremeGradient2022}. Further, for landslide susceptibility,  \citeauthor{zhangInsightsGeospatial2023} relate the spatial heterogeneity of the \gls{shap} values to the natural characteristics and human activities for the following factors: lithology, slope, elevation, rainfall, and \gls{ndvi} \cite{zhangInsightsGeospatial2023}. They argue that the differences in factor contributions can be attributed to local and regional characteristics such as topography, geology, or vegetation. Another indicator is the land cover classes used by \citeauthor{abitbolInterpretableSocioeconomic2020} to identify the relationship between urban topology and the average household income \cite{abitbolInterpretableSocioeconomic2020}. They relate the \gls{gradcam} attributions of the image pixels to their landcover classes to identify that commercial/residential units are characterized by low income, while natural areas describe higher income.

In certain works, expert knowledge is used to validate or interpret the findings.
For example, the most important drivers of landslides noted by the field investigation reports are compared to the features sorted by explanation magnitudes in \cite{chenAttributionDeep2023}. \gls{lda} is leveraged in  \cite{karmakarFeaturefreeExplainable2021} for unsupervised sea ice classification, and closely related classes are identified by \gls{kl} divergence. The \gls{lda} derived topics and probabilities, together with the interclass distances and segmented images, enable experts to assess the physical relationship between these classes. For example, water bodies, melted snow, and water currents have a similar topic distribution and a substantial physical similarity: liquid water.
In \cite{liExplainableDimensionality2023}, expert knowledge is employed to interpret and guide the process of finding and validating dwelling styles and their evolution within ethnic communities. The building footprints are classified with \gls{xgboost}, and \gls{xai} is applied by leveraging \gls{shap} to determine the importance. Then, the experts infer the semantic meaning. The results reveal the emergence of mixed ethical styles inheriting from the three traditional styles, which can be correlated to migration records.

\subsubsection{Evaluation of Explainable AI Methods}
\label{sec:results_eval_xai}
\begin{figure}[!t]
  \centering
  \includegraphics[trim=.1cm .35cm 0cm 0cm,clip,width=0.5\textwidth]{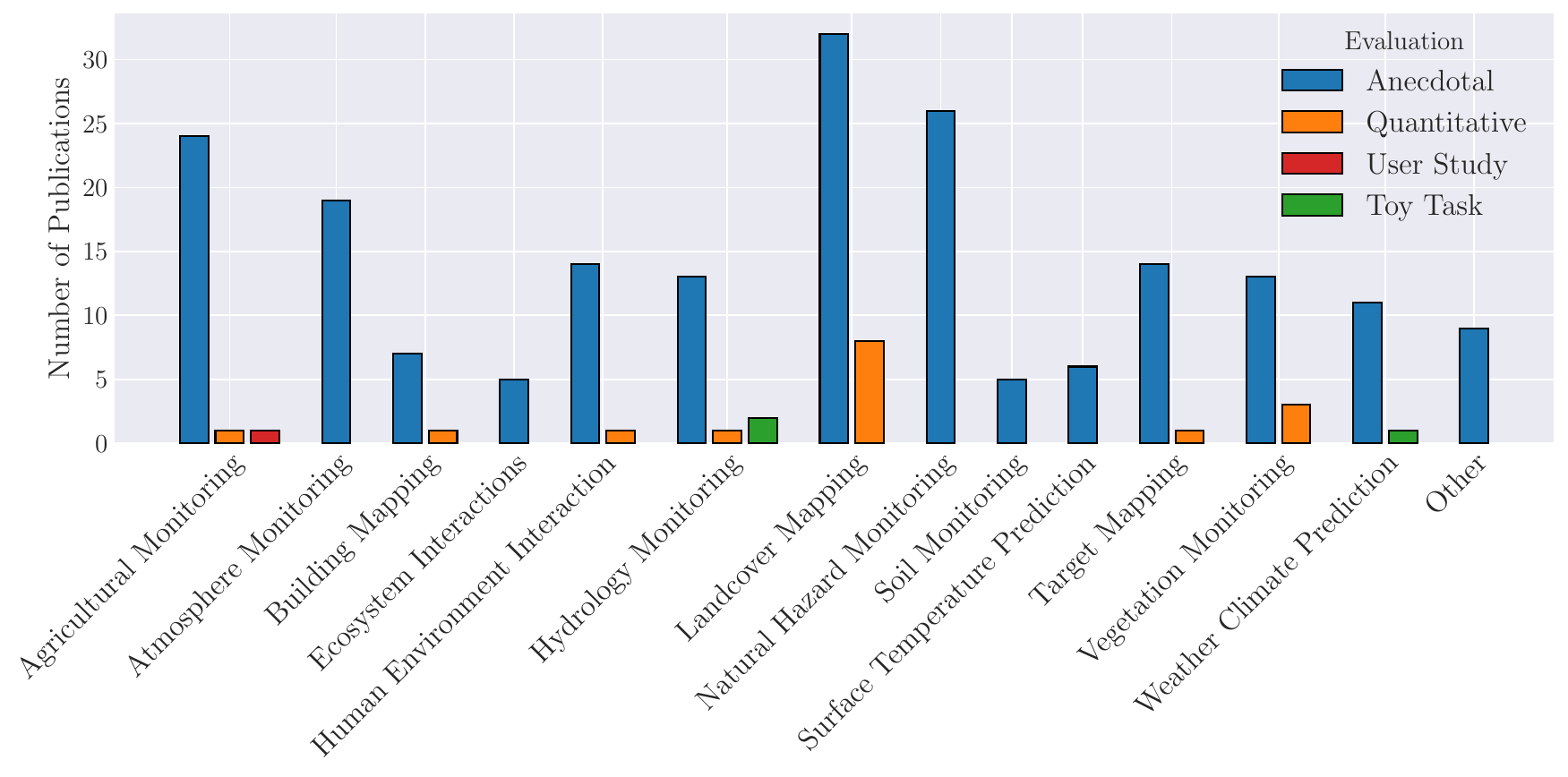}
  \caption{The number of times the evaluation types are considered in the different \gls{eo} tasks. While anecdotal evaluation is prominent across the \gls{eo} tasks, quantitative evaluation is rare, and most often, it is conducted for landcover mapping problems. A single work relies on user studies for agriculture, and three studies apply their methods to toy tasks.}
  \label{fig:evaluation}
\end{figure}
As indicated in Section \ref{sec:background_xai_methods_evaluation}, \gls{xai} evaluation poses an open challenge. In remote sensing, most of the literature relies on anecdotal evidence, often involving the visualization of arbitrarily selected or cherry-picked examples. Figure \ref{fig:evaluation} illustrates that
quantitative evaluation is mostly conducted in well-established \gls{eo} tasks, such as landcover mapping and vegetation monitoring. Further, a single user study is conducted for agriculture monitoring. 
In addition to these types of evaluation, some authors \cite{barnesThisLooks2022, blixGaussianProcess2017, zhangFeatureImportance2023} evaluate their methods on straightforward toy tasks, where humans can easily identify the sought relationships. 

\citeauthor{paudelInterpretabilityDeep2023} exclusively conduct a user survey to assess (1) the importance of the \gls{dl} features by experts and (2) judge the importance by the \gls{xai} method \cite{paudelInterpretabilityDeep2023}. Five crop modeling experts assigned importance scores to the features. Subsequently, these scores were compared to the feature importance estimated with \gls{shap}. Afterward, the experts categorized the model explanations into four categories (strong) agree, (strong) disagree, and should justify their decision. Overall, it is demonstrated that experts can understand the model explanations, and the explanations enable the experts to get insights into the models. However, the task remains challenging and has the potential for misconceptions about the model behavior.

When it comes to the quantitative evaluation, we identified 15 studies testing the \gls{xai} methods on \gls{rs} problems with functional metrics. These metrics mainly asses the explanation quality properties described in Section \ref{sec:background_xai_methods_evaluation}. Particularly, the backpropagation methods are most frequently evaluated with a specific focus on the \gls{cam} approaches. 
In detail, \citeauthor{kakogeorgiouEvaluatingExplainable2021} \cite{kakogeorgiouEvaluatingExplainable2021} evaluate eight backpropagtion methods, Occlusion, and \gls{lime} on landcover mapping tasks. Utilized metrics are max-sensitivity, file size, computation time, and \gls{moref}. Max-sensitivity measures the reliability (maximum change in explanation) when the input is slightly perturbed, the file size is used as a proxy for explanation sparsity, and \gls{moref} measures how fast the classification accuracy declines when removing the most relevant explanations. The results indicate no obvious choice for this task. While Occlusion, \gls{gradcam}, and \gls{lime} were the most reliable according to the max-sensitivity metric, they lack high-resolution explanations.
Further, the studies \cite{deluciaExplainableAI2022, songMedianpoolingGradCAM2021, fengSelfMatchingCAM2021} evaluate various \gls{cam} methods on class sensitivity and by measuring drop/increase in confidence when occluding parts of the image. In a different study, \cite{kucklickTacklingAccuracyinterpretability2023} demonstrates the low faithfulness of the \gls{gradcam} explanations based on similar metrics. The authors also conduct model and data randomization tests to find that \gls{gradcam} is sensitive to changes in network weights and label randomization.
Other studies also evaluate the localization ability of \gls{cam} methods by turning the attributions into segmentation masks and comparing the IoU or classification accuracy \cite{suWhichCAM2022,guoVisualExplanations2023,marvasti-zadehCrownCAMInterpretable2023}. 
Additionally, \cite{kimFederatedOnboardground2022} compare attention networks and \gls{cam} variants on the metrics max-sensitivity and average \% drop/increase in confidence.
Regarding other \gls{xai} approaches, the attention weights are evaluated in \cite{obadicExploringSelfAttention2022} by inspecting drops in the accuracy for crop mapping when the transformer model is trained on a subset of dates with the highest attention values. The results verify that attention weights select the key dates for crop discrimination, as training the model with only the top 15 attended dates is sufficient to approximate the accuracy of the model trained on the complete dataset.
Further, \citeauthor{dantasCounterfactualExplanations2023} \cite{dantasCounterfactualExplanations2023} employ distinct metrics for their counterfactual generation. These metrics aim to ensure a certain quality of the counterfactuals. They used proximity (evaluating closeness to the original input instance, measured by $l_2$ distance), compactness (ensuring a small number of perturbations across time steps), stability (measuring the consistency for comparable input samples) and plausibility (measuring adherence to the same data distribution). Finally, despite the high usage of the local approximation methods in \gls{rs}, we did not find any works that quantitatively evaluate their explanations.

\subsection{RQ3: Explainable AI Objectives and Findings in Remote Sensing}
\begin{figure}[!t]
  \centering
  \includegraphics[trim=.1cm .2cm 0cm 0cm,clip,width=0.5\textwidth]{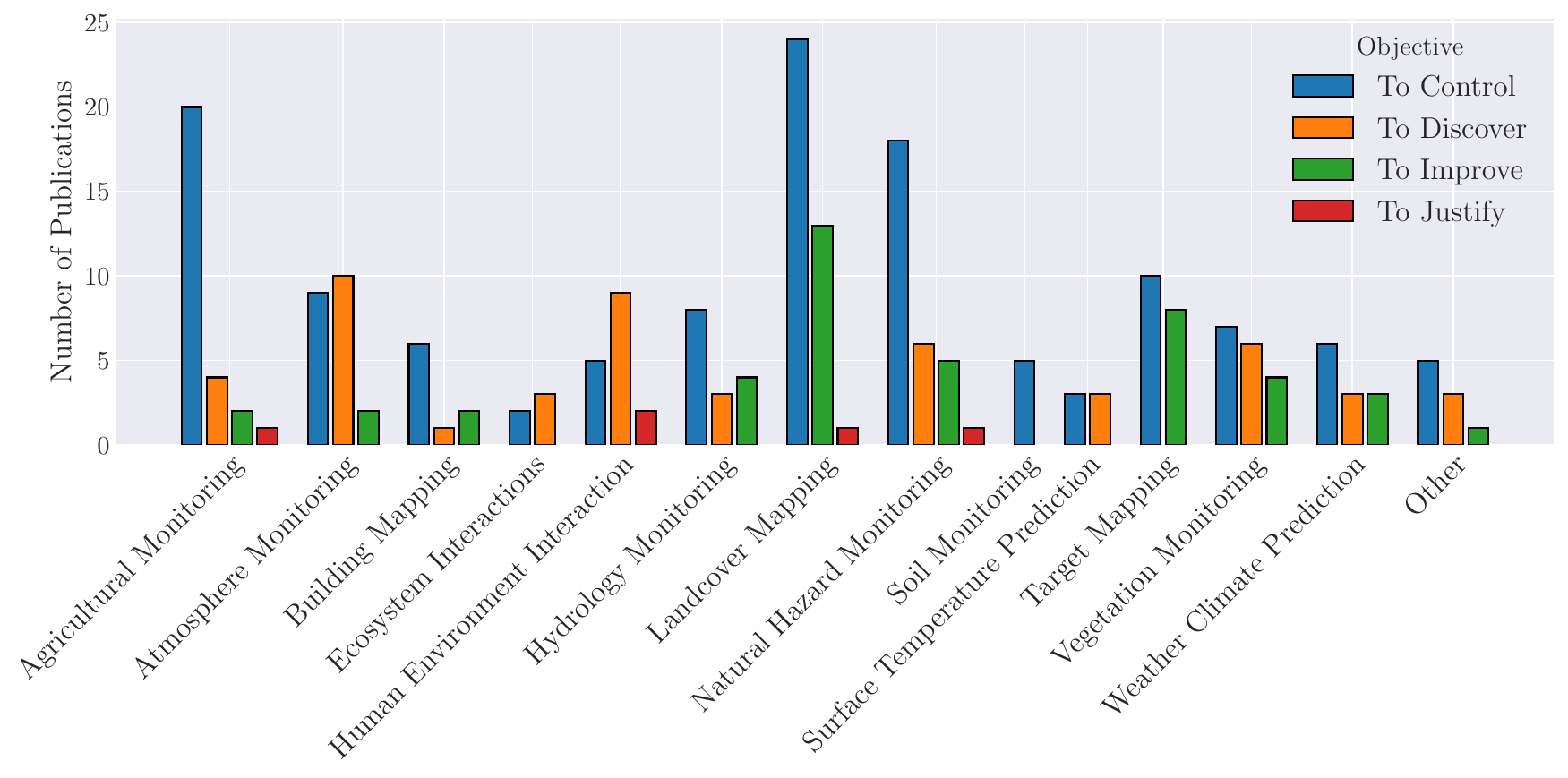}
  \caption{The frequency of the objectives for using \gls{xai} in \gls{rs} identified in the analyzed works, categorized according to the scheme introduced in Section \ref{chap:xai_obj}. Papers can appear in multiple categories due to ambiguous meanings or multiple motivations and objectives. The objectives to control and improve are widely used among the traditional \gls{eo} tasks such as landcover or target mapping. The objective to discover is frequently found among the more recent \gls{eo} tasks like monitoring the atmosphere and human environment interaction. Finally, the \gls{xai} approaches in \gls{rs} rarely have the objective to justify.}
  \label{fig:objectives}
\end{figure}

In this section, we analyze the motivation for applying \gls{xai} in \gls{rs} according to the common objectives specified in Section \ref{chap:xai_obj}.
The frequency of the objectives across the \gls{eo} tasks is visualized in Figure \ref{fig:objectives} and reflects a similar trend as in \cite{gevaertExplainableAI2022}. Namely, the objective to control is the most common and is followed by a large gap with the objectives to discover insights and to improve. Finally, the objective to justify is rare and found only in a few works. Notably, this figure also shows that in contrast to the objective to control which is mostly identified in the three most common \gls{eo} tasks (landcover mapping, agricultural monitoring, and natural hazard monitoring), the objective to discover has a unique distribution across the \gls{eo} tasks as it frequently occurs in studies monitoring the atmosphere, vegetation, and human environment interaction.

The studies with the objective \emph{to control} are mainly comparing the inference mechanisms of various established \gls{ml} models used for \gls{eo}. Consequently, they are mostly conducted in landcover mapping. Next, they are often found in agricultural monitoring, where they mainly aim to assess the reliability of the proposed models by quantifying the relevance of the multitemporal information.
For instance, \citeauthor{russwurmSelfattentionRaw2020} use the gradient method to measure the temporal importance assigned by various \gls{dl} models for crop classification \cite{russwurmSelfattentionRaw2020}. Their analysis reveals that the transformer and the \gls{lstm} approaches ignore the observations obscured by clouds and focus on a relatively small number of observations when compared to the \gls{cnn} models. Following a similar approach, \citeauthor{xuInterpretingMultitemporal2021} evaluate the generalization capabilities of these models when inference is performed for different years than the ones used for model training \cite{xuInterpretingMultitemporal2021}. Their experiments indicate that the \gls{lstm} model adapts better to changes in crop phenology induced by late plantation compared to the transformer model. The objective to control also supports studies that anticipate the model decisions in scenarios that can occur in practical applications. For instance, \citeauthor{gawlikowskiExplainingEffects2022} investigate the impact of exposing a \gls{dnn}, initially trained on cloud-free images, to cloudy images \cite{gawlikowskiExplainingEffects2022}. They apply \gls{gradcam} to identify crucial regions for the classifier in both image types. Their findings reveal different factors why the network misclassifies the \gls{ood} examples, including the coverage of structures through cloud cover and shadows, as well as the homogeneity or heterogeneity induced by different cloud types.

The observed distribution of the \gls{eo} tasks with the objective \emph{to discover} in Figure \ref{fig:objectives}b indicates that using \gls{xai} for gaining new knowledge is usually applied for the recently explored \gls{eo} tasks. For instance, the problem of finding the key drivers for wildfire susceptibility is tackled in \cite{kondylatosWildfireDanger2022}. Applying \gls{shap} and \gls{pdp} on the trained \gls{ml} model reveals that soil moisture, humidity, temperature variables, wind speed, and \gls{ndvi} are among the most important factors associated with wildfires.
Regarding monitoring natural hazards, \citeauthor{biassInsightsVulnerability2022} uses \gls{shap} to identify that volcanic deposits, terrain properties, and vegetation types are strongly linked to vegetation vulnerability after volcanic eruptions \cite{biassInsightsVulnerability2022}. In detail, increased vegetation vulnerability is associated with higher lapilli accumulations, with crops and forests being the most and the least susceptible vegetation types, respectively.
In another study, \citeauthor{stombergExploringWilderness2022} aims to discover new knowledge about the ambiguously defined concept of wilderness \cite{stombergExploringWilderness2022}. By analyzing occlusion sensitivity maps from their proposed deep learning models for predicting wilderness characteristics, they reveal that wilderness is characterized by large areas containing natural undisrupted soils. This is in contrast to anthropogenic areas that have specific edge shapes and lie close to impervious structures.
Besides discovering new knowledge, scientific insights can also be used to identify the reasons for inaccuracies in Earth system models. \citeauthor{silvaUsingExplainable2022} provide insights into a lightning model's structural deficits by predicting its error with a gradient boosting algorithm and interpreting it with \gls{shap}  \cite{silvaUsingExplainable2022}. The error is computed as the difference between the output of the model and satellite observational data. Their analysis reveals potential deficits in high convective precipitation and landcover heterogeneities.

A large part of the works with the objective \emph{to improve} focus on adapting the existing \gls{xai} methods to \gls{rs} tasks and are already described in Section \ref{sec:results_adapted_approaches}.
Concerning the other techniques for model improvement based on \gls{xai} insights, we found that it is often used for data augmentation. For instance, \citeauthor{bekerDeepLearning2023} simulate synthetic data for training a \gls{cnn} model for the detection of volcanic deformations \cite{bekerDeepLearning2023}. By conducting an explainability analysis with \gls{gradcam} on real data, the authors identify that the model wrongly predicts volcanic deformations on some patterns not considered in the simulated data. These insights are used to improve the prediction performance by fine-tuning the model on a hybrid synthetic-real dataset that accounts for these patterns.
Further, \citeauthor{kimFederatedOnboardground2022} propose an iterative classification model improvement for satellite onboard processing through a weakly supervised human-in-the-loop process \cite{kimFederatedOnboardground2022}. An inconsistency metric is introduced to measure the similarity of the attribution maps emphasizing commonly highlighted regions to identify uncertain explanations across the attention blocks. Experts refine badly performing samples by labeling the incorrect pixels in the attention map. In the last step, the onboard model weights get updated by retraining with the refined data.

Lastly, the objective \emph{to justify} usually discusses the relevance of the explainability insights from the stakeholder's perspective. 
For instance, \citeauthor{campos-tabernerUnderstandingDeep2020} argue that the interpretability of the \gls{dl} models in agricultural applications is critical to ensure fair payouts to the farmers according to the EU \gls{cap} \cite{campos-tabernerUnderstandingDeep2020}. By applying a perturbation approach, they find that the summer acquisitions and the red and near-infrared Sentinel-2 spectral bands carry essential information for land use classification. 
Further, the human footprint index, which represents the human pressure on the landscape and can be a valuable metric for environmental assessments, is predicted from Landsat imagery in \cite{keysMachinelearningApproach2021}. In order to build the policymakers' trust, \gls{lrp} is leveraged to visually highlight the relevant features in the images.
Lastly, the approach presented in \cite{ishikawaExamplebasedExplainable2023} can assist individual users during the production phase of a \gls{ml} model, as it justifies the model's validity for inference by providing example-based explanations. If the provided example does not fit the input instance, the model can be considered unreliable for the \gls{rs} image classification task.

\begin{scriptsize}
\onecolumn 
\begin{longtable}[h]{|p{2cm}|p{4.4cm}|p{4.4cm}|p{2cm}|p{.06cm}p{.06cm}p{.06cm}|p{.06cm}p{.06cm}p{.06cm}p{.06cm}|}
    \caption{A complete list of all relevant papers in this review, aggregated by EO Task and xAI Category. (\textsuperscript{\dag xy} indicates a new method which was derived from xy; a full list of acronyms can be found in Appendix \ref{sec:glossary}.)}
    \label{tab:summary} \\
    \hline
    Earth Observation Task & Explainable AI Category & Paper, Explainable AI Methods & Model & Evaluation & & & Objective & & & \\
    & & & & \multicolumn{3}{c|}{\begin{tikzpicture}
    \begin{axis}[
        ybar,
        bar width=.4cm,
        ticks=none,
        clip=false,
        axis lines = none,
        width=3*.9cm,
        height=3cm,
    ]
    \addplot coordinates {(0,3) (1,196) (2,16)};
    \node at (axis cs:0,3) [rotate=90,anchor=west,xshift=-0.1cm] {Toy Task};
    \node at (axis cs:1,3) [rotate=90,anchor=west,xshift=-0.1cm] {Anecdotal};
    \node at (axis cs:2,3) [rotate=90,anchor=west,xshift=-0.1cm] {Quantitative};
    \end{axis}
\end{tikzpicture}} & \multicolumn{4}{c|}{\begin{tikzpicture}
    \begin{axis}[
        ybar,
        bar width=.4cm,
        ticks=none,
        clip=false,
        axis lines = none,
        width=4*.84cm,
        height=3cm,
    ]
    \addplot coordinates {(0,127) (1,48) (2,51) (3,5)};
    \node at (axis cs:0,5) [rotate=90,anchor=west,xshift=-0.1cm] {Control};
    \node at (axis cs:1,5) [rotate=90,anchor=west,xshift=-0.1cm] {Improve};
    \node at (axis cs:2,5) [rotate=90,anchor=west,xshift=-0.1cm] {Discover};
    \node at (axis cs:3,5) [rotate=90,anchor=west,xshift=-0.1cm] {Justify};
    \end{axis}
\end{tikzpicture}} \\\hline \hline
    \endhead
Agricultural Monitoring & Backpropagation & \cite{geInterpretableDeep2023} PWGrad-CAM\textsuperscript{\dag Grad-CAM} & CNN &  & \ding{51} &  & \ding{51} &  &  &  \\
 &  & \cite{wolaninEstimatingUnderstanding2020} RAM & CNN &  & \ding{51} &  &  &  & \ding{51} &  \\ \cline{2-11}
 & Backpropagation, Embedding Space & \cite{russwurmSelfattentionRaw2020} Gradient, Activation Assessment, Attention & CNN, LSTM, Transformer, ConvLSTM &  & \ding{51} &  & \ding{51} &  &  &  \\ \cline{2-11}
 & Backpropagation, Feature Selection, Embedding Space & \cite{xuInterpretingMultitemporal2021} Gradient, Activation Assessment, MDI, Attention & aLSTM, Transformer, RF &  & \ding{51} &  & \ding{51} &  &  &  \\ \cline{2-11}
 & Backpropagation, Joint Training & \cite{arunLearningPhysically2022} Explanation Association, LRP & GAN &  & \ding{51} &  & \ding{51} & \ding{51} &  &  \\ \cline{2-11}
 & Backpropagation, Local Approximation & \cite{mateo-sanchisInterpretableLongShort2023} SHAP, IG & LSTM &  & \ding{51} &  & \ding{51} &  &  &  \\
 &  & \cite{paudelInterpretabilityDeep2023} SHAP, IG & LSTM &  & \ding{51} &  & \ding{51} &  &  &  \\ \cline{2-11}
 & Embedding Space & \cite{saintefaregarnotSatelliteImage2020} Attention & Transformer &  & \ding{51} &  & \ding{51} &  &  &  \\ \cline{2-11}
 & Feature Selection & \cite{orynbaikyzyCropType2020} MDI, GFFS & RF &  & \ding{51} &  & \ding{51} &  &  &  \\ \cline{2-11}
 & Feature Selection, Model Translation & \cite{newmanExplainableMachine2021} MDI, Rule Extraction & DT, RF &  & \ding{51} &  & \ding{51} &  &  &  \\ \cline{2-11}
 & Interpretable by Design & \cite{celikExplainableArtificial2023} EBM & EBM &  & \ding{51} &  & \ding{51} &  &  &  \\ \cline{2-11}
 & Joint Training & \cite{rosaLearningCrop2023} Prototype & CNN &  & \ding{51} &  & \ding{51} &  &  &  \\ \cline{2-11}
 & Local Approximation & \cite{bromsCombinedAnalysis2023} SHAP & GB &  & \ding{51} &  & \ding{51} &  &  &  \\
 &  & \cite{hanExplainableXGBoost2022} SHAP & GB &  & \ding{51} &  & \ding{51} &  &  &  \\
 &  & \cite{huberExtremeGradient2022} SHAP & GB &  & \ding{51} &  & \ding{51} &  &  &  \\
 &  & \cite{iatrouRepresentationLearning2022} SHAP & RF &  & \ding{51} &  &  &  & \ding{51} &  \\
 &  & \cite{jonesIdentifyingCauses2022} SHAP & GB &  & \ding{51} &  & \ding{51} &  & \ding{51} &  \\
 &  & \cite{singhSimulationMultispectral2023} SHAP & GLM, kNN, GB, SVM, RF, FNN &  & \ding{51} &  & \ding{51} &  &  &  \\ \cline{2-11}
 & Perturbation & \cite{brenningSpatialMachinelearning2023} Spatial Variable Importance Profiles\textsuperscript{\dag PFI} & kNN LDA, LDA, RF &  & \ding{51} &  &  & \ding{51} &  &  \\
 &  & \cite{filippiIdentifyingCrop2022} PFI & RF &  & \ding{51} &  & \ding{51} &  &  &  \\
 &  & \cite{mateo-sanchisLearningMain2021}  & GP &  & \ding{51} &  & \ding{51} &  &  &  \\ \cline{2-11}
 & Perturbation, Embedding Space & \cite{campos-tabernerUnderstandingDeep2020} Activation Assessment, Occlusion & LSTM &  & \ding{51} &  &  &  &  & \ding{51} \\
 &  & \cite{obadicExploringSelfAttention2022} Attention, Occlusion & Transformer &  &  & \ding{51} & \ding{51} &  &  &  \\ \cline{2-11}
 & Perturbation, Interpretable by Design & \cite{martinez-ferrerCropYield2021} Occlusion, GP & GP &  & \ding{51} &  & \ding{51} &  &  &  \\
 &  & \cite{nayakInterpretableMachine2022} ALE, PDP, PFI, GLM & GLM, RF &  & \ding{51} &  &  &  & \ding{51} &  \\ \hline 
Atmosphere Monitoring & Backpropagation & \cite{luoExplainableSpatial2022} IG & NN Ensemble &  & \ding{51} &  &  &  & \ding{51} &  \\ \cline{2-11}
 & Backpropagation, Perturbation, Local Approximation & \cite{valdesExplainableAI2022} XRAI, SHAP, Occlusion & CNN &  & \ding{51} &  &  &  & \ding{51} &  \\ \cline{2-11}
 & Embedding Space & \cite{yanNewInterpretable2020} Activation Assessment & FNN &  & \ding{51} &  &  &  & \ding{51} &  \\
 &  & \cite{yanUnderstandingGlobal2021} Attention & FNN &  & \ding{51} &  &  &  & \ding{51} &  \\ \cline{2-11}
 & Feature Selection & \cite{chenEstimationAtmospheric2022} MDI & Tree Ensemble &  & \ding{51} &  & \ding{51} &  &  &  \\
 &  & \cite{chenEstimationNearsurface2023} MDI & Tree Ensemble &  & \ding{51} &  & \ding{51} &  &  &  \\
 &  & \cite{chenExploringHighresolution2023} MDI & Tree Ensemble &  & \ding{51} &  & \ding{51} &  &  &  \\ \cline{2-11}
 & Feature Selection, Joint Training, Embedding Space & \cite{yanSpatialTemporalInterpretable2021} Model Association, Activation Assessment, MDI & FNN, NN+ML Ensemble, RF &  & \ding{51} &  & \ding{51} &  &  &  \\ \cline{2-11}
 & Feature Selection, Local Approximation & \cite{milaEstimatingDaily2023} MDI, SHAP & RF &  & \ding{51} &  & \ding{51} &  &  &  \\ \cline{2-11}
 & Interpretable by Design & \cite{maranzanoAdaptiveLASSO2023} GLM\textsuperscript{\dag GLM} & GLM &  & \ding{51} &  &  & \ding{51} &  &  \\ \cline{2-11}
 & Local Approximation & \cite{ebrahimi-khusfiDeterminingContribution2021} SHAP & FNN &  & \ding{51} &  &  &  & \ding{51} &  \\
 &  & \cite{shaoEstimationDaily2023} SHAP & GB, GLM &  & \ding{51} &  &  &  & \ding{51} &  \\
 &  & \cite{wangEstimatingParticulate2023} SHAP & GB &  & \ding{51} &  & \ding{51} &  & \ding{51} &  \\
 &  & \cite{zipfelMachinelearningBased2022} SHAP & GB &  & \ding{51} &  &  &  & \ding{51} &  \\ \cline{2-11}
 & Perturbation & \cite{liangIntegratingLowcost2023} PFI & aCNN &  & \ding{51} &  & \ding{51} &  &  &  \\
 &  & \cite{zhenEvaluationMACC2023} ALE, PFI & RF &  & \ding{51} &  & \ding{51} & \ding{51} &  &  \\ \cline{2-11}
 & Perturbation, Feature Selection & \cite{wuEstimateNearsurface2023} MDI, PDP & RF &  & \ding{51} &  &  &  & \ding{51} &  \\ \cline{2-11}
 & Perturbation, Local Approximation & \cite{valdesMachineLearning2021} Ceteris Paribus Profiles, PFI, LIME, SHAP & RF &  & \ding{51} &  &  &  & \ding{51} &  \\
 &  & \cite{zhangDataaugmentationApproach2022} PDP, SHAP & GB &  & \ding{51} &  & \ding{51} &  &  &  \\ \hline 
Building Mapping & Backpropagation & \cite{chengUncertaintyawareConvolutional2022} Grad-CAM & CNN &  & \ding{51} &  & \ding{51} &  &  &  \\
 &  & \cite{hasanpourzaryabiUnboxingBlack2022} DeepLift, A*G & CNN &  & \ding{51} &  & \ding{51} &  &  &  \\
 &  & \cite{seydiBDDNetBuilding2023} Grad-CAM & aCNN &  & \ding{51} &  & \ding{51} &  &  &  \\
 &  & \cite{suWhichCAM2022} Grad-CAM++, CAM, Grad-CAM, Score-CAM, SmoothGrad-CAM++ & CNN &  &  & \ding{51} & \ding{51} &  &  &  \\ \cline{2-11}
 & Backpropagation, Embedding Space & \cite{seydiLeveragingInvolution2023} Attention, Grad-CAM & CNN &  & \ding{51} &  & \ding{51} &  &  &  \\ \cline{2-11}
 & Backpropagation, Perturbation, Embedding Space & \cite{xiongDisentangledLatent2022} Activation Assessment, Occlusion, IG & CNN, Transformer &  & \ding{51} &  & \ding{51} & \ding{51} &  &  \\ \cline{2-11}
 & Feature Selection, Local Approximation & \cite{matinEarthquakeinducedBuildingdamage2021} MDI, SHAP & FNN, RF &  & \ding{51} &  &  & \ding{51} &  &  \\ \cline{2-11}
 & Local Approximation & \cite{liExplainableDimensionality2023} SHAP & GB &  & \ding{51} &  &  &  & \ding{51} &  \\ \hline 
Ecosystem Interactions & Interpretable by Design & \cite{zhouRelativeImportance2021} GAM & GAM &  & \ding{51} &  &  &  & \ding{51} &  \\ \cline{2-11}
 & Interpretable by Design, Local Approximation & \cite{zhuInvestigatingImpacts2022} GLM, SHAP & GB, GLM &  & \ding{51} &  &  &  & \ding{51} &  \\ \cline{2-11}
 & Local Approximation & \cite{descalsLocalInterpretation2023} SHAP & GB &  & \ding{51} &  & \ding{51} &  &  &  \\
 &  & \cite{liExploringIndividualized2022} SHAP & GB &  & \ding{51} &  &  &  & \ding{51} &  \\ \cline{2-11}
 & Perturbation, Local Approximation & \cite{zhuExplainableMachine2023} PFI, SHAP & GB &  & \ding{51} &  & \ding{51} &  &  &  \\ \hline 
Human Environment Interaction & Backpropagation & \cite{abitbolInterpretableSocioeconomic2020} GuidedGrad-CAM & CNN &  & \ding{51} &  &  &  & \ding{51} &  \\
 &  & \cite{keysMachinelearningApproach2021} LRP & CNN &  & \ding{51} &  & \ding{51} &  &  & \ding{51} \\ \cline{2-11}
 & Backpropagation, Perturbation, Embedding Space & \cite{kucklickTacklingAccuracyinterpretability2023} Grad-CAM, Activation Assessment, Occlusion & NN Ensemble, GLM &  &  & \ding{51} &  &  &  & \ding{51} \\ \cline{2-11}
 & Interpretable by Design & \cite{scepanovicJaneJacobs2021} GLM & GLM &  & \ding{51} &  & \ding{51} &  &  &  \\ \cline{2-11}
 & Joint Training & \cite{leveringPredictingLiveability2023} Explanation Association & CNN &  & \ding{51} &  & \ding{51} &  &  &  \\
 &  & \cite{leveringRelationLandscape2021} Explanation Association & CNN &  & \ding{51} &  &  &  & \ding{51} &  \\ \cline{2-11}
 & Local Approximation & \cite{chenMeasuringImpacts2020} SHAP & RF &  & \ding{51} &  &  &  & \ding{51} &  \\
 &  & \cite{dingResidentialGreenness2023} SHAP & GB &  & \ding{51} &  &  &  & \ding{51} &  \\
 &  & \cite{kimModelingPredicting2022} SHAP & GB &  & \ding{51} &  & \ding{51} &  &  &  \\
 &  & \cite{scepanovicMedSatPublic2023} SHAP & GB &  & \ding{51} &  &  &  & \ding{51} &  \\
 &  & \cite{temenosNovelInsights2022} LIME, SHAP & RF &  & \ding{51} &  &  &  & \ding{51} &  \\ \cline{2-11}
 & Perturbation & \cite{wangUnpackingInter2023} ALE & RF &  & \ding{51} &  &  &  & \ding{51} &  \\
 &  & \cite{wojcikPredictingIntraurban2022} PFI & CNN &  & \ding{51} &  & \ding{51} &  &  &  \\ \cline{2-11}
 & Perturbation, Embedding Space & \cite{stombergExploringWilderness2022} Activation Assessment, Occlusion & CNN &  & \ding{51} &  &  &  & \ding{51} &  \\ \cline{2-11}
 & Perturbation, Local Approximation & \cite{prasadEnhancedJoint2023} Occlusion, LIME, SHAP & FNN &  & \ding{51} &  &  &  & \ding{51} &  \\ \cline{2-11}
Hydrology Monitoring & Backpropagation & \cite{blixGaussianProcess2017} Gradient & GP & \ding{51} & \ding{51} &  &  & \ding{51} &  &  \\
 &  & \cite{blixLearningRelevant2022} Gradient & GP &  & \ding{51} &  & \ding{51} &  &  &  \\
 &  & \cite{chenTransparentDeep2023} Grad-CAM & CNN &  & \ding{51} & \ding{51} & \ding{51} & \ding{51} &  &  \\
 &  & \cite{hongMonitoringVertical2021} Grad-CAM & CNN &  & \ding{51} &  & \ding{51} &  &  &  \\
 &  & \cite{liuExplainableDeep2023} Gradient & CNN &  & \ding{51} &  &  &  & \ding{51} &  \\
 &  & \cite{rampalInterpretableDeep2022} Grad-CAM & CNN &  & \ding{51} &  &  & \ding{51} &  &  \\ \cline{2-11}
 & Backpropagation, Local Approximation & \cite{yilmazMarineMucilage2023} SHAP, IG & CNN &  & \ding{51} &  & \ding{51} &  &  &  \\ \cline{2-11}
 & Backpropagation, Perturbation & \cite{jingInterpretingRunoff2023} DeepLift, PDP, Expected Gradients, IG & LSTM &  & \ding{51} &  & \ding{51} &  & \ding{51} &  \\ \cline{2-11}
 & Embedding Space & \cite{pyoCyanobacteriaCell2021} Attention & CNN &  & \ding{51} &  & \ding{51} &  &  &  \\
 &  & \cite{zhangFeatureImportance2023} Attention & FNN & \ding{51} & \ding{51} &  &  & \ding{51} &  &  \\ \cline{2-11}
 & Local Approximation & \cite{huWaterStorage2023} SHAP & FNN &  & \ding{51} &  & \ding{51} &  &  &  \\
 &  & \cite{saeidiWaterDepth2023} SHAP & RF, GB, GB, DT &  & \ding{51} &  & \ding{51} &  &  &  \\ \cline{2-11}
 & Perturbation, Interpretable by Design, Local Approximation & \cite{chengImprovedUnderstanding2022} ALE, PFI, Cubist, LIME & GB, Cubist &  & \ding{51} &  &  &  & \ding{51} &  \\ \hline 
Landcover Mapping & Backpropagation & \cite{deluciaExplainableAI2022} 3DGrad-CAM\textsuperscript{\dag Grad-CAM} & CNN &  & \ding{51} & \ding{51} & \ding{51} & \ding{51} &  &  \\
 &  & \cite{gawlikowskiExplainingEffects2022} Grad-CAM & CNN &  & \ding{51} &  & \ding{51} &  &  &  \\
 &  & \cite{guoVisualExplanations2023} CSG-CAM\textsuperscript{\dag Grad-CAM}, Grad-CAM++, Grad-CAM, Score-CAM & CNN &  & \ding{51} & \ding{51} &  & \ding{51} &  &  \\
 &  & \cite{huangBetterVisual2022} ERC-CAM\textsuperscript{\dag CAM} & CNN &  & \ding{51} &  &  & \ding{51} &  &  \\
 &  & \cite{hungIntegratingImage2021} XRAI & CNN &  & \ding{51} &  & \ding{51} &  &  &  \\
 &  & \cite{jeonRecursiveVisual2023} Grad-CAM & CNN &  & \ding{51} & \ding{51} &  & \ding{51} &  &  \\
 &  & \cite{Sen2020} Grad-CAM & CNN &  & \ding{51} &  & \ding{51} &  &  &  \\
 &  & \cite{songMedianpoolingGradCAM2021} Grad-CAM++, SmoothGrad-CAM++, Grad-CAM, MPGrad-CAM\textsuperscript{\dag Grad-CAM} & CNN &  & \ding{51} & \ding{51} &  & \ding{51} &  &  \\
 &  & \cite{vasuResiliencePlasticity2020} CAM & CNN &  & \ding{51} & \ding{51} & \ding{51} &  &  &  \\ \cline{2-11}
 & Backpropagation, Embedding Space & \cite{kimFederatedOnboardground2022} Attention, Layer-CAM, Attention, CAM, Attention, Grad-CAM & CNN &  &  & \ding{51} &  & \ding{51} &  &  \\ \cline{2-11}
 & Backpropagation, Perturbation, Local Approximation & \cite{kakogeorgiouEvaluatingExplainable2021} DeepLift, I*G, Gradient, Grad-CAM, Occlusion, IG, GuidedBackprop, LIME, GuidedGrad-CAM & CNN &  & \ding{51} & \ding{51} & \ding{51} &  &  &  \\ \cline{2-11}
 & Counterfactuals & \cite{dantasCounterfactualExplanations2023} GAN & CNN &  &  & \ding{51} & \ding{51} &  &  &  \\ \cline{2-11}
 & Embedding Space & \cite{chengPolSARImage2021} Activation Assessment & CapsuleNet, CNN &  & \ding{51} &  & \ding{51} &  &  &  \\
 &  & \cite{guidiciOneDimensionalConvolutional2017} Activation Assessment & CNN &  & \ding{51} &  & \ding{51} &  &  &  \\
 &  & \cite{iencoWeaklySupervised2020} Attention & CNN &  & \ding{51} &  & \ding{51} &  &  &  \\
 &  & \cite{megerExplainingDeep2022} Activation Assessment, Attention & CNN &  & \ding{51} &  &  & \ding{51} &  &  \\ \cline{2-11}
 & Example-based & \cite{ishikawaExamplebasedExplainable2023} WIK & CNN &  & \ding{51} &  & \ding{51} & \ding{51} &  & \ding{51} \\ \cline{2-11}
 & Feature Selection, Local Approximation & \cite{fisherUncertaintyawareInterpretable2022} MDI, SHAP & CNN, RF &  & \ding{51} &  & \ding{51} &  &  &  \\
 &  & \cite{jayasingheCausesTea2023} MDI, LIME & RF &  & \ding{51} &  & \ding{51} &  &  &  \\ \cline{2-11}
 & Interpretable by Design & \cite{fengBidirectionalFlow2022} Rules & NN+Tree Ensemble &  & \ding{51} &  & \ding{51} &  &  &  \\
 &  & \cite{karmakarFrameworkInteractive2022} LDA & LDA &  & \ding{51} &  & \ding{51} &  &  &  \\ \cline{2-11}
 & Interpretable by Design, Embedding Space & \cite{huangPhysicallyExplainable2022} Activation Assessment, LDA & NN Ensemble &  & \ding{51} &  &  & \ding{51} &  &  \\ \cline{2-11}
 & Joint Training & \cite{maMulticropFusion2022} Prototype & CNN &  & \ding{51} &  &  & \ding{51} &  &  \\ \cline{2-11}
 & Local Approximation & \cite{burguenoScalableApproach2023} SHAP & RF &  & \ding{51} &  & \ding{51} &  &  &  \\
 &  & \cite{elliottIdentifyingCritical2022} LIME, SHAP & CNN &  & \ding{51} &  & \ding{51} &  &  &  \\
 &  & \cite{hosseinyUrbanLand2022} SHAP & CNN, GB, SVM, RF &  & \ding{51} &  & \ding{51} &  &  &  \\
 &  & \cite{temenosInterpretableDeep2023} SHAP & CNN &  & \ding{51} &  & \ding{51} &  &  &  \\
 &  & \cite{vermaExplainableCustom2021} LIME & CNN &  & \ding{51} &  & \ding{51} &  &  &  \\ \cline{2-11}
 & Local Approximation, Embedding Space & \cite{andersonFuzzyChoquet2018} Activation Assessment, SHAP & NN Ensemble &  & \ding{51} &  & \ding{51} &  &  &  \\
 &  & \cite{hungRemoteSensing2020} Activation Assessment, LIME & CNN &  & \ding{51} &  &  & \ding{51} &  &  \\
 &  & \cite{islamEnablingExplainable2020} Activation Assessment, SHAP & NN Ensemble &  & \ding{51} &  & \ding{51} & \ding{51} &  &  \\
 &  & \cite{murrayInformationFusion2text2020} Activation Assessment, SHAP & NN Ensemble &  & \ding{51} &  & \ding{51} &  &  &  \\ \cline{2-11}
 & Perturbation, Feature Selection, Local Approximation & \cite{chenEnhancingLand2023} MDI, PFI, SHAP & RF &  & \ding{51} &  & \ding{51} &  &  &  \\ \cline{2-11}
 & Perturbation, Local Approximation & \cite{brenningInterpretingMachinelearning2023} ALE, PDP, PFI, SHAP & RF &  & \ding{51} &  &  & \ding{51} &  &  \\ \hline 
Natural Hazard Monitoring & Backpropagation, Embedding Space & \cite{chenAttributionDeep2023} DeepLift, Attention & NN Ensemble &  & \ding{51} &  &  & \ding{51} & \ding{51} &  \\ \cline{2-11}
 & Backpropagation, Perturbation, Embedding Space & \cite{bekerDeepLearning2023} Activation Assessment, Occlusion, Grad-CAM & CNN &  & \ding{51} &  & \ding{51} & \ding{51} &  &  \\ \cline{2-11}
 & Backpropagation, Perturbation, Local Approximation & \cite{kondylatosWildfireDanger2022} PDP, SHAP, IG & LSTM &  & \ding{51} &  &  &  & \ding{51} &  \\ \cline{2-11}
 & Interpretable by Design & \cite{maxwellExplainableBoosting2021} EBM & EBM &  & \ding{51} &  & \ding{51} &  &  &  \\
 &  & \cite{stroppianaFullyAutomatic2021} Fuzzy Rules & Rules &  & \ding{51} &  & \ding{51} &  &  & \ding{51} \\
 &  & \cite{youssefLandslideSusceptibility2022} SNN\textsuperscript{\dag FNN} & GAM &  & \ding{51} &  &  & \ding{51} &  &  \\ \cline{2-11}
 & Joint Training & \cite{zhangInterpretableDeep2022} Prototype & CNN &  & \ding{51} &  &  & \ding{51} &  &  \\ \cline{2-11}
 & Local Approximation & \cite{abdollahiExplainableArtificial2023} SHAP & FNN &  & \ding{51} &  &  &  & \ding{51} &  \\
 &  & \cite{al-najjarNovelMethod2022} SHAP & SVM, RF &  & \ding{51} &  & \ding{51} &  &  &  \\
 &  & \cite{aydinPredictingAnalyzing2022} SHAP & GB, RF &  & \ding{51} &  & \ding{51} &  &  &  \\
 &  & \cite{dahalExplainableArtificial2022} SHAP & FNN &  & \ding{51} &  & \ding{51} &  &  &  \\
 &  & \cite{ibanSnowAvalanche2023} SHAP & GB &  & \ding{51} &  & \ding{51} &  &  &  \\
 &  & \cite{inanExplainableAI2023} SHAP & GB &  & \ding{51} &  &  & \ding{51} &  &  \\
 &  & \cite{jenaExplainableArtificial2023} SHAP & NN+Tree Ensemble &  & \ding{51} &  & \ding{51} &  &  &  \\
 &  & \cite{liuResidualNeural2023} SHAP & CNN &  & \ding{51} &  & \ding{51} &  &  &  \\
 &  & \cite{sainiE2AlertNetExplainable2023} LIME, SHAP & CNN &  & \ding{51} &  & \ding{51} &  &  &  \\
 &  & \cite{vegaLandslideModeling2023} SHAP & RF &  & \ding{51} &  & \ding{51} &  &  &  \\
 &  & \cite{zhangInsightsGeospatial2023} SHAP & GB &  & \ding{51} &  &  &  & \ding{51} &  \\ \cline{2-11}
 & Local Approximation, Embedding Space & \cite{fangNewApproach2023} Activation Assessment, SHAP & GAMI-Net &  & \ding{51} &  & \ding{51} &  & \ding{51} &  \\ \cline{2-11}
 & Perturbation & \cite{Taylor2020} PDP, PFI & RF &  & \ding{51} &  &  &  & \ding{51} &  \\ \cline{2-11}
 & Perturbation, Feature Selection & \cite{rubiPerformanceComparison2023} MDI, PFI & NB, GLM, kNN, SVM, RF, Tree Ensemble, FNN, GLM &  & \ding{51} &  & \ding{51} &  &  &  \\ \cline{2-11}
 & Perturbation, Local Approximation & \cite{alqadhiIntegratedDeep2023} PDP, PFI, SHAP & FNN, CNN, NN Ensemble &  & \ding{51} &  & \ding{51} &  &  &  \\
 &  & \cite{chenTunnelGeothermal2023} PDP, Occlusion, LIME & Tree Ensemble &  & \ding{51} &  & \ding{51} &  &  &  \\
 &  & \cite{cilliExplainableArtificial2022} PFI, SHAP & RF &  & \ding{51} &  & \ding{51} &  &  &  \\
 &  & \cite{sunAssessmentLandslide2023} PDP, LIME & RF &  & \ding{51} &  & \ding{51} &  &  &  \\
 &  & \cite{wangXGBoostSHAPApproach2023} PDP, SHAP & GB &  & \ding{51} &  & \ding{51} &  &  &  \\ \hline 
Soil Monitoring & Backpropagation, Perturbation & \cite{huangInterpretingConvLSTM2023} SquareGrad, IG, I*G, Gradient, PFI, VarGrad, SmoothGrad & ConvLSTM &  & \ding{51} &  & \ding{51} &  &  &  \\ \cline{2-11}
 & Local Approximation & \cite{mukhamedievSoilSalinity2023} SHAP & GB &  & \ding{51} &  & \ding{51} &  &  &  \\
 &  & \cite{pradhanNewMethod2022} SHAP & CNN &  & \ding{51} &  & \ding{51} &  &  &  \\
 &  & \cite{zhouIdentificationSoil2022} SHAP & SVM &  & \ding{51} &  & \ding{51} &  &  &  \\ \cline{2-11}
 & Perturbation & \cite{Smorkalov2022} PFI & RF &  & \ding{51} &  & \ding{51} &  &  &  \\ \hline 
Surface Temperature Prediction & Backpropagation & \cite{labePredictingSlowdowns2022} LRP & FNN &  & \ding{51} &  & \ding{51} &  &  &  \\ \cline{2-11}
 & Feature Selection, Local Approximation & \cite{milaEstimatingDaily2023} MDI, SHAP & RF &  & \ding{51} &  & \ding{51} &  &  &  \\ \cline{2-11}
 & Local Approximation & \cite{huPixelLevel2023} SHAP & GB &  & \ding{51} &  &  &  & \ding{51} &  \\
 &  & \cite{kimExaminingRelationship2022} SHAP & GB &  & \ding{51} &  & \ding{51} &  &  &  \\
 &  & \cite{luoUnderstandingRelationship2023} SHAP & GB &  & \ding{51} &  &  &  & \ding{51} &  \\
 &  & \cite{shenUsingGeoAI2023} SHAP & GB &  & \ding{51} &  &  &  & \ding{51} &  \\ \hline 
Target Mapping & Backpropagation & \cite{fengSelfMatchingCAM2021} Self-Matching-CAM\textsuperscript{\dag Grad-CAM} & CNN &  &  & \ding{51} &  & \ding{51} &  &  \\
 &  & \cite{liAutomaticBridge2023} CAM & CNN &  & \ding{51} &  & \ding{51} &  &  &  \\
 &  & \cite{liDeepNetworks2018} CAM & CNN &  &  &  &  & \ding{51} &  &  \\
 &  & \cite{luoGlassboxingDeep2021} Score-CAM, IG & NN Ensemble &  & \ding{51} &  & \ding{51} &  &  &  \\
 &  & \cite{yinG2GradCAMRLObject2023} Grad-CAM & CNN &  & \ding{51} &  & \ding{51} &  &  &  \\ \cline{2-11}
 & Backpropagation, Embedding Space & \cite{liSARBagNetAntehoc2022} Activation Assessment, CAM & BagNet &  & \ding{51} &  & \ding{51} &  &  &  \\ \cline{2-11}
 & Backpropagation, Local Approximation & \cite{pengClutterinvariantRegularization2023} SHAP, GuidedBackprop & CNN &  & \ding{51} &  &  & \ding{51} &  &  \\
 &  & \cite{zhuLIMEBasedData2022} CAM, LIME & CNN &  & \ding{51} &  &  & \ding{51} &  &  \\ \cline{2-11}
 & Backpropagation, Perturbation, Embedding Space & \cite{xiongInterpretableFusion2022} Activation Assessment, Attention, Grad-CAM, Empirical Receptive Field & CNN &  & \ding{51} &  & \ding{51} & \ding{51} &  &  \\ \cline{2-11}
 & Embedding Space & \cite{fengPANPart2023} Attention & CNN &  & \ding{51} &  &  & \ding{51} &  &  \\
 &  & \cite{guoSARAutomatic2021} Activation Assessment & VAE &  & \ding{51} &  & \ding{51} &  &  &  \\
 &  & \cite{liSARADBagNetInterpretable2022} Activation Assessment & BagNet &  & \ding{51} &  & \ding{51} &  &  &  \\
 &  & \cite{zhouLocalAttention2020} Attention & aCNN &  & \ding{51} &  & \ding{51} &  &  &  \\ \cline{2-11}
 & Local Approximation & \cite{kawauchiSHAPBasedInterpretable2022} SHAP & CNN &  & \ding{51} &  &  & \ding{51} &  &  \\
 &  & \cite{mandeepDeepLearningbased2020} LIME & CNN &  & \ding{51} &  & \ding{51} &  &  &  \\
 &  & \cite{oveisLIMEAssistedAutomatic2023} LIME & CNN &  & \ding{51} &  &  & \ding{51} &  &  \\ \cline{2-11}
 & Perturbation & \cite{fengSARTarget2021} Occlusion & NN Ensemble &  & \ding{51} &  & \ding{51} &  &  &  \\ \hline 
Vegetation Monitoring & Backpropagation & \cite{marvasti-zadehCrownCAMInterpretable2023} Eigen-CAM, Crown-CAM\textsuperscript{\dag Grad-CAM}, AugScore-CAM, Score-CAM & CNN &  &  & \ding{51} &  & \ding{51} &  &  \\
 &  & \cite{onishiExplainableIdentification2021} GuidedGrad-CAM & CNN &  & \ding{51} &  & \ding{51} &  &  &  \\ \cline{2-11}
 & Backpropagation, Local Approximation & \cite{jiaStudyingExploiting2021} Gradient, SHAP, GuidedGrad-CAM, Deconvolution & CNN &  &  & \ding{51} &  & \ding{51} &  &  \\ \cline{2-11}
 & Feature Selection, Local Approximation & \cite{sotomayorSupervisedMachine2023} MDI, LIME & GB &  & \ding{51} &  & \ding{51} &  &  &  \\ \cline{2-11}
 & Interpretable by Design & \cite{lambertVegetationChange2020} GLM & GLM &  & \ding{51} &  &  &  & \ding{51} &  \\
 &  & \cite{leon-garzaBigBangbig2020} Fuzzy Rules & FLS &  & \ding{51} &  & \ding{51} &  &  &  \\ \cline{2-11}
 & Joint Training & \cite{nguyenMappingForest2022} Explanation Association & CNN &  & \ding{51} &  & \ding{51} &  &  &  \\ \cline{2-11}
 & Local Approximation & \cite{abdollahiUrbanVegetation2021} SHAP & FNN &  & \ding{51} &  & \ding{51} & \ding{51} &  &  \\
 &  & \cite{andresiniSILVIAEXplainable2023} SHAP & GB, SVM, RF &  & \ding{51} &  & \ding{51} &  &  &  \\
 &  & \cite{arrogante-funesAssessmentRegeneration2023} SHAP & Extra Tree &  & \ding{51} &  & \ding{51} &  &  &  \\
 &  & \cite{liWidespreadIncreasing2022} SHAP & RF &  & \ding{51} &  &  &  & \ding{51} &  \\ \cline{2-11}
 & Model Translation & \cite{zhaoBetterUnderstanding2023} Rule Extraction & RF &  & \ding{51} &  &  &  & \ding{51} &  \\
 &  & \cite{zhaoIdentifyingMangroves2023} Rule Extraction & RF &  & \ding{51} &  &  & \ding{51} &  &  \\ \cline{2-11}
 & Perturbation & \cite{schillerHigherCrop2023} PDP, PFI & RF &  & \ding{51} &  &  &  & \ding{51} &  \\ \cline{2-11}
 & Perturbation, Local Approximation & \cite{biassInsightsVulnerability2022} PFI, SHAP & GB &  & \ding{51} &  &  &  & \ding{51} &  \\
 &  & \cite{mullerFeaturesPredisposing2022} PFI, SHAP & RF &  & \ding{51} &  &  &  & \ding{51} &  \\ \hline 
Weather Climate Prediction & Backpropagation & \cite{hilburnDevelopmentInterpretation2020} LRP & CNN &  & \ding{51} &  & \ding{51} &  &  &  \\
 &  & \cite{martinUsingSimple2022} LRP & FNN &  & \ding{51} &  & \ding{51} &  &  &  \\
 &  & \cite{mayerSubseasonalForecasts2021} LRP & FNN &  & \ding{51} &  &  &  & \ding{51} &  \\
 &  & \cite{rampalHighresolutionDownscaling2022} Grad-CAM & CNN &  & \ding{51} &  & \ding{51} &  &  &  \\
 &  & \cite{yangPredictorSelection2023} GuidedBackprop & CNN &  & \ding{51} &  &  & \ding{51} &  &  \\ \cline{2-11}
 & Backpropagation, Feature Selection & \cite{liAdvancingSatellite2021} MDI, Grad-CAM & CNN, RF &  & \ding{51} &  & \ding{51} &  &  &  \\ \cline{2-11}
 & Feature Selection & \cite{upadhyayaClassifyingPrecipitation2021} TreeInterpreter & RF &  & \ding{51} &  &  &  & \ding{51} &  \\ \cline{2-11}
 & Joint Training & \cite{barnesThisLooks2022} Prototype\textsuperscript{\dag ProtoPNet} & CNN & \ding{51} & \ding{51} &  &  & \ding{51} &  &  \\ \cline{2-11}
 & Local Approximation & \cite{mardianUnderstandingDrivers2023} SHAP & GB &  & \ding{51} &  & \ding{51} &  &  &  \\
 &  & \cite{silvaUsingExplainable2022} SHAP & GB &  & \ding{51} &  &  & \ding{51} & \ding{51} &  \\ \cline{2-11}
 & Perturbation, Local Approximation & \cite{antoniadouComparisonDatadriven2023} PFI, SHAP & FNN, SVM, GLM, RF &  & \ding{51} &  & \ding{51} &  &  &  \\ \hline 
Other & Backpropagation & \cite{maddyMIIDAPSAIExplainable2021} Gradient & FNN &  & \ding{51} &  & \ding{51} &  &  &  \\ \cline{2-11}
 & Feature Selection & \cite{mudeleModelingTemporal2020} MDI & RF &  & \ding{51} &  & \ding{51} &  &  &  \\ \cline{2-11}
 & Interpretable by Design & \cite{karmakarFeaturefreeExplainable2021} LDA & LDA &  & \ding{51} &  &  &  & \ding{51} &  \\ \cline{2-11}
 & Interpretable by Design, Embedding Space & \cite{huangPhysicallyExplainable2022} Activation Assessment, LDA & NN Ensemble &  & \ding{51} &  &  & \ding{51} &  &  \\ \cline{2-11}
 & Joint Training & \cite{bergamascoEXPLAINABLECONVOLUTIONAL2020} Explanation Association & CNN &  & \ding{51} &  & \ding{51} &  &  &  \\ \cline{2-11}
 & Local Approximation & \cite{kimTrueGlobal2023} SHAP & RF &  & \ding{51} &  & \ding{51} &  &  &  \\
 &  & \cite{leeInterpretableMachine2022} SHAP & RF &  & \ding{51} &  & \ding{51} &  &  &  \\
 &  & \cite{wagnerUsingExplainable2022} SHAP & GB &  & \ding{51} &  &  &  & \ding{51} &  \\ \cline{2-11}
 & Perturbation & \cite{taconetDatadrivenInterpretable2021} PDP, PFI & RF &  & \ding{51} &  &  &  & \ding{51} &  

\\
\hline
\end{longtable}
\twocolumn 
\end{scriptsize}
\clearpage
\section{Discussion}
\label{sec:discussion}
\noindent In this section, we address research questions RQ4 and RQ5 and discuss the usage of \gls{xai} methods in \gls{rs}, the adaptation of \gls{xai} methods to \gls{rs} problems, as well as the evaluation of such methods (RQ4). Furthermore, we highlight the challenges, limitations, and emerging research directions in the field (RQ5). 

\subsection{RQ4: Recommended Practices in Explainable AI for Remote Sensing}
\label{sec:rq4_recommended_partices}

\subsubsection{Choice of Explainable AI Method}
The high usage of specific model-agnostic and post-hoc methods, such as \gls{shap} and \gls{cam}, observed in Section \ref{sec:xai_patterns_in_rs}, suggests that they are often used regardless of the specific properties of the \gls{eo} task. We want to draw attention to this trend since there is \emph{no single method that fits every problem} \cite{molnarPitfallsAvoid2020}. Each model-agnostic method represents one approach or paradigm to generate explanations, regardless of the trained model and input dataset. Moreover, the post-hoc approaches only provide an approximation of the workings of the underlying model, which limits their faithfulness. 
These limitations illustrate a trade-off when selecting the \gls{xai} method for a specific application; although the intrinsic interpretable \gls{xai} approaches can overcome the faithfulness issues of the post-hoc methods, they require additional effort during the model design phase, particularly in the form of injecting domain knowledge for deriving interpretable and useful features for the task at hand or for designing a more complex neural network that can provide ante-hoc explanations \cite{rudinStopExplaining2019}. Therefore, the choice of the \gls{xai} method should depend on the specific goal, taking into account properties such as the need for global or local explanations, their model-agnostic or model-specific nature, the computational time required for the generation of explanations, the quality or detail of these explanations, or the suitability for the end-user, among others.
Another essential factor to consider is that each \gls{xai} method makes certain assumptions (e.g., \gls{shap} assumes feature independence that simplifies the problem's definition). These assumptions can often be violated when applying the \gls{xai} methods in \gls{rs}, particularly due to the specific properties of \gls{rs} data related to scale, temporal dependencies, and geographical confounders (described below in Section \ref{sec:rs_properties_challenge}). For instance, the works listed in Section \ref{sec:results_adapted_approaches} show that \gls{gradcam} struggles to accurately localize the objects in the remote sensing imagery due to their scale, and several methods are proposed to overcome this limitation. Thus, the adapted methods can be a valuable starting point for the choice of the \gls{xai} method in \gls{rs}.

In summary, the choice of the appropriate \gls{xai} method for a specific \gls{eo} application depends on many factors and is not straightforward. This calls for a thorough discussion and should be as much a part of the scientific dialogue as the model evaluation. Nevertheless, numerous papers in this review did not mention their arguments for selecting their \gls{xai} methods. We encourage the authors to discuss this issue to lay the basis for the understanding and scientific evaluation of their approaches. Good perspectives in this respect are provided by \citeauthor{liExploringIndividualized2022} in \cite{liExploringIndividualized2022}.

\subsubsection{Towards Reliable Explanations}
In order to get meaningful and reliable explanations, \gls{xai} methods should only be applied to \gls{ml} models that have achieved a good generalization \cite{molnarPitfallsAvoid2020}. Furthermore, the \emph{robustness of \gls{ml} models and \gls{xai} methods} should be quantified, as it is common for them to disagree \cite{krishnaDisagreementProblem2022}. 
To maintain consistent outcomes, it is advisable to consider a set of initializations of \gls{xai} methods and different \gls{ml} models, as it is done when benchmarking the performance of a \gls{ml} model. For instance, in the works identified in our review, different model types \cite{aydinPredictingAnalyzing2022, andresiniSILVIAEXplainable2023, alqadhiIntegratedDeep2023, chenTransparentDeep2023,  al-najjarNovelMethod2022, xuInterpretingMultitemporal2021, jingInterpretingRunoff2023}, model seeds or configurations \cite{paudelInterpretabilityDeep2023, jingInterpretingRunoff2023}, and \gls{xai} method seeds \cite{liangIntegratingLowcost2023} were used. Furthermore, the outcomes of different \gls{xai} methods can be compared, as it is done in \cite{al-najjarNovelMethod2022,chengImprovedUnderstanding2022,guoVisualExplanations2023,hasanpourzaryabiUnboxingBlack2022,jingInterpretingRunoff2023,kimFederatedOnboardground2022,luoUnderstandingRelationship2023,marvasti-zadehCrownCAMInterpretable2023,mateo-sanchisInterpretableLongShort2023,nayakInterpretableMachine2022,newmanExplainableMachine2021,songMedianpoolingGradCAM2021,suWhichCAM2022,zhuInvestigatingImpacts2022}.
For example, \citeauthor{jingInterpretingRunoff2023} apply \gls{ig}, \gls{eg}, and DeepLift to exhaustively interpret different \gls{rnn}-based architectures (\gls{rnn}, \gls{lstm}, \acrshort{gru}), also varying their numbers of neurons \cite{jingInterpretingRunoff2023}. Other related works are \cite{zhuInvestigatingImpacts2022}, where \citeauthor{zhuInvestigatingImpacts2022} compare the results of gradient boosting and \gls{shap} applied to a \gls{lr} model, or \cite{hasanpourzaryabiUnboxingBlack2022}, where \citeauthor{hasanpourzaryabiUnboxingBlack2022} compare the use of \gls{I*G} and \gls{deeplift} in several layers.
Additionally, when using methods where the input is altered, like \gls{ig} or \gls{shap}, the baseline is the neutral input, which should capture the absence of any meaningful feature, and all outcomes are compared to the output for the baseline input. Hence, the choice of the baseline is crucial, and some methods have a built-in baseline by construction. This is especially important when different methods need to be compared \cite{haugBaselinesLocal2021, mamalakisCarefullyChoose2023}.
Finally, for verifying the correctness of the \gls{xai} methods, \emph{evaluating} their outputs with some of the functional metrics described in Section \ref{sec:func_eval} is recommended, as discussed below in Section \ref{sec:ev_xai}.

Often, practitioners are interested in causal relationships; however, the explanation results of black-box models might lead to \emph{unjustified causal claims} \cite{molnarPitfallsAvoid2020}. \gls{xai} methods do not provide the cause. Even when the explanation supports the hypothesis from the research objective, it does not mean that this hypothesis has been proven. Standard \gls{ml} approaches approximate correlations from data; therefore, the model cannot assess the causal structure of the data. However, it can help experts find unknown correlations by giving them a tool to investigate the learned correlations. The user must be very careful when making such assumptions to discover underlying structures and be aware that observational data often lacks common confounders, has strong feature correlations, and has a causal structure that is usually unknown. Causal inference is a complementary research field to \gls{xai}, not covered in this review. Nevertheless, we identified several works discussing the causality problem in \gls{xai} \cite{biassInsightsVulnerability2022, hanExplainableXGBoost2022, descalsLocalInterpretation2023, labePredictingSlowdowns2022}.

\subsubsection{Evaluation of Explainable AI}
\label{sec:ev_xai}
As mentioned in Section \ref{sec:results_eval_xai}, anecdotal evidence is frequently reported for the evaluation of \gls{xai} approaches. Cherry-picking and qualitative evaluation of explanations represent a challenge to humans. Because human perception is mainly visual, humans are biased toward certain types of \gls{xai} explanations. Besides others, humans introduce cognitive distortion by drawing more attention to negative examples and looking for simple but complete explanations \cite{bertrandHowCognitive2022}. That is accompanied by other human biases, like the confirmation bias, which is well-known in psychology and favors explanations fitting the expectations, while contrary explanations are ignored \cite{nickersonConfirmationBias1998}. These self-introduced biases can lead to wrong reasoning about the explanations and promoting a particular type of visualization. Hence, it is hard to quantify the results objectively through anecdotal evaluation and in case it is used as the only assessment criteria, it does not lead to a trustworthy evaluation of \gls{xai}. 

Beyond evaluation based on anecdotal evidence, there also exist quantitative evaluation metrics (see Section \ref{sec:background_xai_methods_evaluation}) to assess the quality of the explanation methods. These studies can reveal the shortcomings of the existing \gls{xai} methods and the scenarios in which the explanations are unfaithful to the underlying model. For instance, \citeauthor{adebayoSanityChecks2018} \cite{adebayoSanityChecks2018} demonstrate that perturbing the weights of the neural network does not significantly change the resulting explanations for several \gls{xai} methods while \citeauthor{Sixt2019} \cite{Sixt2019} builds a theoretical framework, showing that many methods fail to provide class-sensitive explanations and highlight only low-level features \cite{Sixt2019}. However, Section \ref{sec:results_eval_xai} demonstrates that only a few works in \gls{rs} evaluate the estimated feature importance of the used \gls{xai} method with the functional evaluation metrics. 
In contrast to these studies, it has been observed that a more detailed quantitative evaluation of \gls{xai} is conducted in climate science. For instance, \citeauthor{bommerFindingRight2023} \cite{bommerFindingRight2023} evaluate seven backpropagation methods, among them three \gls{cam}-based methods. They use a climate model to get reliable ground truth data for their explanations and compare the employed \gls{xai} methods with functional evaluation by computing the following metrics: robustness, faithfulness, randomization, complexity, and localization. Some of these backpropagation methods are also compared for their fidelity in \cite{mamalakisInvestigatingFidelity2022}. First, the methods are evaluated on a synthetically created dataset. Then, the results are related to those on a climate simulation classification dataset.

Until now, there is no standard way to evaluate \gls{xai}, and discussions on how to do it properly are still ongoing.
Nonetheless, different toolboxes have been developed for comparing \gls{xai} methods using quantitative metrics \cite{hedstromQuantusExplainable2023, belaidWeNeed2022, agarwalOpenXAITransparent2023}. Last but not least, only one survey \cite{paudelInterpretabilityDeep2023} evaluates the usefulness of the explanations to experts. However, with the emphasis on human-centered \gls{ai} in the current research landscape \cite{xuTransitioningHuman2023}, user studies are becoming significant to quantify the benefit of the explanations and the understanding of the end-users. Because they have been largely unexplored in the context of \gls{xai} in \gls{rs}, they pose a promising research direction.

\subsubsection{Explainable AI Benchmark Datasets}
Real-world datasets lack a controlled ground truth for evaluating \gls{xai} methods and a regulated environment is needed to lay the foundation for fully transparent datasets. Similar to \cite{arrasCLEVRXAIBenchmark2022}, there are efforts to create synthetic datasets for \gls{eo} tasks. 
\citeauthor{mamalakisNeuralNetwork2022} generated a fully synthetic dataset where they leverage local piece-wise linear functions to create a non-linear response to the input drawn from a Gaussian distribution \cite{mamalakisNeuralNetwork2022}. This method allows for creating a regional climate prediction task from \gls{sst}. They show that a simple \gls{fnn} can approximate this function and evaluate different post-hoc \gls{xai} methods. 
To get reliable ground truth data, it is also possible to use a simulation. For example, climate models \cite{labeDetectingClimate2021} can be leveraged.
Other approaches to generate fully-synthetic \gls{rs} images include \cite{berksonSyntheticData2019, shermeyerRarePlanesSynthetic2021, hanEfficientGeneration2017}. While \citeauthor{berksonSyntheticData2019} simulate different \gls{rs} sensors, atmospheres, and scenes, including different terrains, materials, and weather conditions \cite{berksonSyntheticData2019}, \citeauthor{shermeyerRarePlanesSynthetic2021} add different compositions of the vehicles \cite{shermeyerRarePlanesSynthetic2021} and \citeauthor{hanEfficientGeneration2017} show that the motions of vehicles can be included in the simulation \cite{hanEfficientGeneration2017}.
A framework for generating synthetic \gls{eo} datasets is presented in \cite{hoeserSyntEOSynthetic2022}, demonstrating the efficiency of these datasets for \gls{ml} on off-shore wind farm detection. The framework aims to extract expert knowledge about the objects to be modeled in a machine-readable format (e.g., structure, relationships, etc.), which can be combined to create new datasets.
However, there is no work comparing \gls{xai} methods on these synthetic \gls{rs} datasets.

\subsection{RQ5: Challenges, Limitations, and Future Directions}
\begin{figure}[!t]
 \centering
    \includegraphics[width=.5\textwidth]{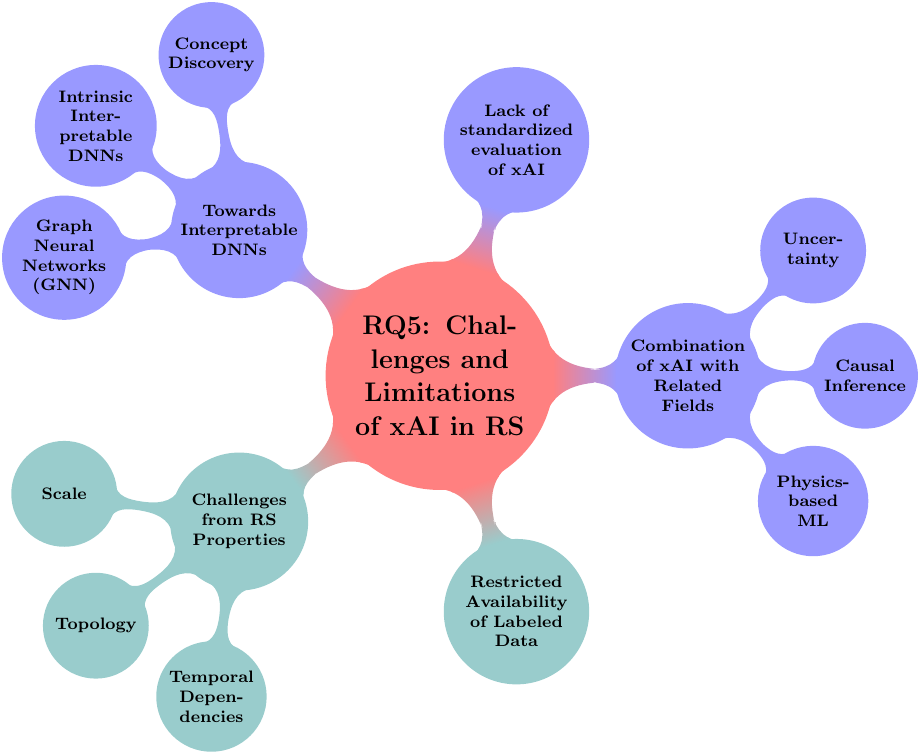}
  \caption{Challenges of \gls{xai} in \gls{rs}. Those emerging from \gls{rs} and \gls{xai} are in green and blue, respectively.}
  \label{fig:chall_mindmap}
\end{figure}

\noindent We identified several challenges that emerge from \gls{xai}, \gls{rs}, or the combination of them. Figure \ref{fig:chall_mindmap} gives an overview of these challenges, which are categorized by color according to the area from which they emerge, i.e., those emerging from \gls{rs} and \gls{xai} are in green and blue, respectively.

\subsubsection{Combination of Explainable AI with Related Fields}
\paragraph{Physics-aware Machine Learning} Physics-aware \gls{ml} and \gls{xai} share the objective of enhancing model reliability, trustworthiness, and transparency. Both rely on domain expertise: \gls{xai} enables experts to discover new insights into physical processes. This is illustrated by studies such as \cite{chengImprovedUnderstanding2022}, which explores runoff and evapotranspiration parameterization, and \cite{silvaUsingExplainable2022}, where the errors of an Earth system model are predicted to hypothesize about wrong model assumptions using \gls{xai}.
Conversely, integrating expert knowledge from physics helps to understand the models because the incorporated prior knowledge is human-understandable. The combination of both fields can yield physically sound and interpretable models. We have identified several approaches relying on physics-aware features applied for knowledge extraction from \gls{sar} data. For instance, \citeauthor{zhaoContrastiveregulatedCNN2019} introduce a \gls{cnn} operating in the complex domain which aims to predict physical scattering properties from \gls{polsar} images \cite{zhaoContrastiveregulatedCNN2019}. These properties are derived with the $H-A-\alpha$ target decomposition method and describe various urban, vegetation, and ocean surfaces.
\citeauthor{huangPhysicallyExplainable2022} inject similar physics-aware concepts into a \gls{cnn} model for sea-ice classification and conducts a \gls{xai} analysis to show that the fusion provides a better separation of the different classes based on their physical properties \cite{huangPhysicallyExplainable2022}. 
Given the vehicle classification task, \citeauthor{fengSARTarget2021} extract vehicle parts with the physics-based \gls{asc} model \cite{fengSARTarget2021}. The vehicle parts are used as filters for the convolution of the original image. With occlusion analysis, the authors demonstrate that incorporating the vehicle parts in the model inputs helps to improve model robustness, enabling the practitioners to obtain intuitive explanations and validate the model workings with domain knowledge. In a follow-up work, \citeauthor{fengPANPart2023} propose an attention-based \gls{cnn} model that derives the queries from the original image, while the keys and the values are extracted from the individual vehicle parts \cite{fengPANPart2023}. Finally, a convolution operation is applied to factor the contribution of the vehicle parts into the class logits, thus explaining the model prediction based on the different parts.
All these works support that physics-aware \gls{ml} and \gls{xai} could complement each other. Further design of these approaches and their extension to other types of \gls{rs} data, such as optical or hyperspectral images, is a promising research direction for model improvement and may also constitute an important step towards consolidating the usage of explainable architectures.

\paragraph{Uncertainty} The uncertainty intervals give insight into the generalization and training process of the model, which is especially useful when dealing with extrapolation and anomaly detection. While explanations can give more detailed information about the models' internal representations, they lack an important metric: model confidence or explanation reliability. Explanations which have a high variability or are inaccurate may lead to misinterpretations.
The combination of uncertainty and \gls{xai} can enhance the information we can get from only one approach. For instance, \citeauthor{marx_but_nodate} \cite{marx_but_nodate} and \citeauthor{slack_reliable_2021} \cite{slack_reliable_2021} propose methods to provide uncertainty intervals around the explanations.
\gls{rs} has not been excluded from the convergence of these two disciplines \cite{huangUncertaintyExploration2023, blomerusFeedbackassistedAutomatic2022}. Through the adaptation of a perturbation method for object detection (D-RISE) and its combination with Deep Gaussian models, \citeauthor{huangUncertaintyExploration2023} get attribution maps for model uncertainty \cite{huangUncertaintyExploration2023}. They show the efficiency of their approach on \gls{sar} object detection, where trustworthy predictions are especially needed since \gls{sar} images are hard to interpret for humans.
An uncertainty aware, interpretable-by-design model frequently applied is \glspl{gp} \cite{blixGaussianProcess2017, blixLearningRelevant2022, martinez-ferrerCropYield2021, mateo-sanchisLearningMain2021}. Further, \citeauthor{blixGaussianProcess2017} analytically derive the input features' sensitivity to the variance estimate of a \gls{gp} model that can be later used for uncertainty evaluation and selection of relevant features \cite{blixGaussianProcess2017}.\\
Hence, complementing \gls{xai} methods with uncertainty constitutes an interesting research direction to ensure more reliable explanations and better interpretation results.

\paragraph{Causal Inference} 
Understanding causality in \gls{xai} is fundamental for uncovering the cause-and-effect relationships within the model's decision-making process. Techniques like causal inference, counterfactual reasoning, and causal graphical models are employed to trace the causal relationships between input features and model predictions \cite{pearl2009causality}. These methods aim to not just highlight correlations but elucidate the direct causal links, enabling better comprehension of how the AI system arrives at its decisions. Moreover, the integration of causal reasoning into interpretable \gls{ml} models, like causal Bayesian networks or causal decision trees, has shown promise in elucidating causal links between input features and predictions \cite{Scholkopf2012OnCA}.
In \gls{rs}, approaches such as structural equation modeling, directed acyclic graphs, and Granger causality have been utilized to untangle causal relationships within \gls{rs} datasets \cite{perez-suayCausalInference2019}. These methods aim to identify causal pathways between environmental variables, allowing scientists to comprehend how changes in one variable may cause alterations in others. However, causality for explaining trained models in \gls{rs} is yet to be extensively explored, with many opportunities ahead in environmental studies, land-use planning, disaster management, and climate change research.

\subsubsection{Remote Sensing Properties}
\label{sec:rs_properties_challenge}
\begin{figure}[!t]
    \centering
    \includegraphics[width=.5\textwidth]{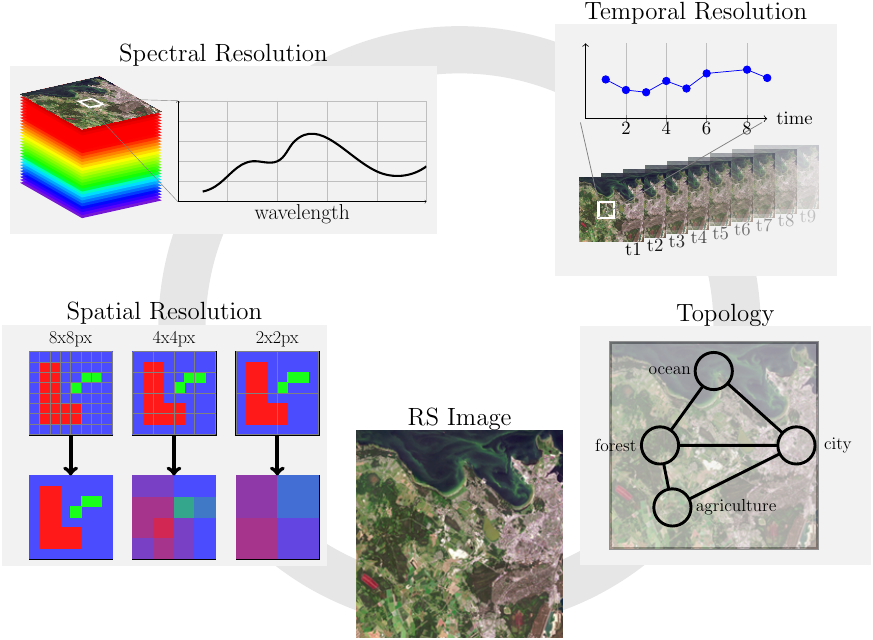}
    \caption{Visualization of challenges emerging from the distinctive \gls{rs} image properties compared to \gls{cv}. The scale has varying spatial and spectral resolution depending on the sensor. As well as the dependencies from the temporal dimension and topology, which are often not considered by traditional \gls{xai} methods. (Image contains modified Copernicus Sentinel 2 data (2020), processed by ESA.)}
    \label{fig:rs_props}
\end{figure}
\gls{xai} in \gls{cv} applications do not take into account \gls{rs} image properties, such as the presence of different sources, scales, geographic relationships, and temporal dependencies.
Figure \ref{fig:rs_props} visualizes the challenges emerging from the \gls{rs} properties.
For instance, the scale, including the spatial and spectral resolution, has semantic implications. While the spatial resolution determines which level of detail can be captured, the spectral bands are sensitive to the ground conditions. Therefore, different sensors allow various interpretations of the same scene.
Further, a large number of \gls{eo} tasks involve analyzing time series data with \glspl{dnn} \cite{millerDeepLearning2024}. However, most \gls{xai} methods were originally designed for images and do not explicitly consider temporal dependencies, making some of them unreliable or inaccurate \cite{ismailBenchmarkingDeep2020}.
Lastly, the geographic dependencies are not captured by the images. For example, the Earth's system components interact with each other, and systems can be influenced by distant processes through teleconnections, often observed in the atmosphere and oceans. Subsequently, the interpretation of a single scene is usually influenced by unobserved factors that can play a crucial role in the processes.
An in-depth discussion about the works identified in this review that tackle these specific \gls{rs} properties can be found in \cite{hoehl_obadic_cvpr_2024}.

\subsubsection{Towards Interpretable Deep Neural Networks}
\label{sec:towards_interpretable_dnns}
\begin{figure*}[htbp]
    \centering
    \includegraphics[trim=0cm 0.9cm 0cm 1.2cm,clip,width=.9\textwidth]{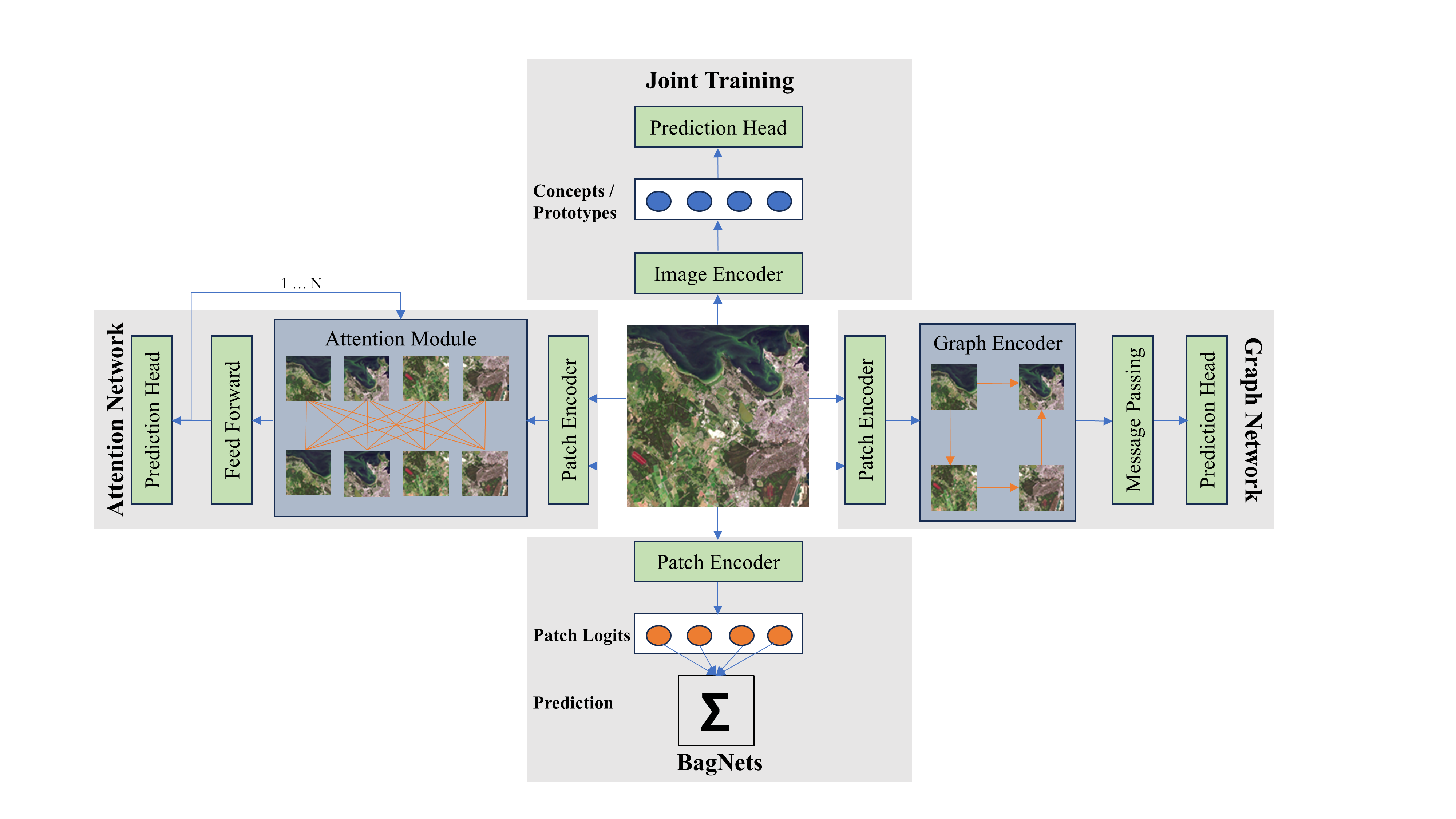}
    \caption{Overview of the interpretable neural network architectures used in \gls{rs}. The joint training approaches typically process the image with a \gls{cnn} encoder that outputs concept or prototype activations to which the model predictions are conditioned. In contrast, the three other approaches first split the image into smaller patches. On the one hand, the BagNet model outputs class logits per patch and then linearly aggregates these logits to form the final prediction. On the other hand, attention-based models and graph neural networks perform the inference by learning relationships between patches. Attention models include a series of layers, each consisting of an attention-weight matrix that captures the pairwise relationships between the patches, while graph neural networks model the relationships in a graph structure. (Image contains modified Copernicus Sentinel 2 data (2020), processed by ESA.)}
    \label{fig:iDNNs}
\end{figure*}
Figure \ref{fig:xai_models_categories_methods} demonstrates that backpropagation methods such as \gls{gradcam} are among the most utilized \gls{xai} approaches for explaining \gls{dl} models. Yet, one of the main drawbacks of these methods is that they attribute importance to individual pixels and regions in the image that might not correspond to high-level concepts that humans can easily interpret \cite{kim2018interpretability}. Moreover, these methods struggle to provide further insights into the semantics that the model has extracted from the attributed regions \cite{achtibat2022towards}.
These limitations have been recently addressed with alternative explanation paradigms such as concept-based explanations and intrinsically interpretable \glspl{dnn} \cite{holzinger2020xxai}.

The concept-based explanations enable global interpretation in terms of intuitive high-level visual concepts that are understandable to humans. Amid the growing number of these approaches in \gls{cv} \cite{kim2018interpretability, ghorbani2019towards, yeh2020completeness, achtibat2022towards, ji2023spatial}, in \gls{rs} they are only utilized in the work of \citeauthor{Obadic_2024_CVPR} \cite{Obadic_2024_CVPR}. In particular, the authors apply the \gls{tcav} method to reveal how various urban and natural concepts relate to continuous intervals of socioeconomic outcomes. Further, they discuss specific challenges in defining visual concepts in \gls{rs}, such as the scale of geographical concepts and their vagueness. Hence, defining visual concepts for the different \gls{rs} sensors is a promising research direction toward more intuitive explanations of the black-box models in \gls{rs}.

Yet, similar to the saliency maps, the concept-based explanation methods are also post-hoc approaches. As such, they approximate the underlying model behavior and do not necessarily make it more transparent.
While mainly simple and easily understandable \gls{ml} models are meant with interpretable models, we want to emphasize the efforts of trying to make \glspl{dnn} inherently interpretable. Current \gls{xai} literature is going in that direction, either by introducing layers in the neural network, like the attention mechanism that can offer insights into the relevant features, or by designing intrinsically interpretable \glspl{dnn}. This review identified the four different approaches of intrinsically interpretable \glspl{dnn} used in \gls{rs}, namely attention networks, joint training models, BagNets \cite{brendelApproximatingCNNs2019}, and \glspl{gnn}, which are visualized in Figure \ref{fig:iDNNs}.

Attention-based models such as the Transformer were initially developed for sequential inputs. Their \gls{cv} counterpart, \glspl{vt} \cite{DosovitskiyB0WZ21}, adapts their structure for the two dimensions of an image. The transformer networks developed for \gls{rs} applications such as the Earthformer \cite{gaoEarthformerExploring2023} further extend the attention mechanism to tackle spatiotemporal inputs. When it comes to the joint training approaches, in addition to the concept bottleneck models described in Section \ref{sec:results_application_existing_methods_in_rs}, mostly prototype networks are used that enforce a reasoning process that classifies input examples based on their similarity to prototypical parts of images of a given class. Further, a recent work introduces a more interpretable \gls{cnn} model, the BagNet \cite{brendelApproximatingCNNs2019} which splits the image into a set of small patches and predicts class logits for each patch. Next, these logits are linearly combined to form the prediction at the image level. For details about the specific works in \gls{rs} using the above three approaches, we refer the reader to \cite{hoehl_obadic_cvpr_2024}.

In contrast to the above \gls{nn} approaches, graphs provide an intuitive representation of many real-world problems regarding objects and their relations. \glspl{gnn} emerged as a popular paradigm for learning high-level object representation by message-passing from the object neighbors \cite{velivckovic2023everything}. Opposed to \glspl{cnn}, which operate on homogeneous grids \cite{zhou2020graph}, \glspl{gnn} enable a reasoning process to represent complex patterns of relationships between objects and generalize their applicability to arbitrary geometric structures. Moreover, \glspl{gcn} \cite{kipf2016semi}, which perform message-passing with the convolution operation, enables efficient semi-supervised learning in scenarios where not all objects in the graph are labeled.
These properties have fostered the usage of \glspl{gnn} in \gls{rs}, especially in hyperspectral image classification tasks. Concretely, to overcome the limitation of the popular \gls{cnn} approaches to capture the topological relations and the irregular object shapes inherent to hyperspectral data, various \gls{gcn} approaches have been proposed \cite{hong2020graph, e0964942fc3b4a8aa75207d1c77c036f, yu2023two}, which demonstrate the potential of using \glspl{gnn} for improving hyperspectral image classification results. Additionally, recent works such as \cite{mou2020nonlocal, yu2023hyperspectral} have also tackled the scarcity of labels challenge with \gls{gcn} approaches. 
Although \glspl{gnn} produce an intuitive representation of the problem, they are considered black-box models \cite{GNNBook-ch7-liu}. Therefore, one promising research direction is adopting the novel \gls{xai} methods for interpreting \glspl{gnn} \cite{yuan2022explainability} to the approaches used in \gls{rs}. 

\subsubsection{Restricted Availability of Labeled data}
\label{sec:discussion_challenge_restricted_labels}
The high number of active airborne platforms like Sentinel-2 generate vast amounts of data. However, only a small portion of this data can be labeled.
The lack of labels represents a challenge in the typical supervised knowledge extraction pipeline in \gls{rs}. Hence, a popular approach is to apply \emph{\gls{ssl}}. \gls{ssl} allows learning latent representations from the unlabeled data. These are useful in efficiently solving the downstream learning task for which only limited labels are available. 
\gls{ssl} is achieved by optimizing a \gls{nn} to solve a pretext task with pseudo-labels \cite{wang2022self}. Combining the traditional \gls{ml} modeling approaches with the recent \gls{dl} approaches represents a promising research direction for improving model interpretability.
A pretext task for the problem of \gls{sar} sea ice classification is proposed in \cite{huangPhysicallyExplainable2022}, where the pseudo-labels are derived from a mixture of \gls{lda} topics. Topics are based on physical scattering properties extracted from unlabeled \gls{sar} images. Thus, the model can provide an interpretable representation of the sea-ice types in terms of their physical properties (e.g., water bodies and floating sea ice are represented with similar properties that match the actual semantic definition of these classes). Their \gls{ssl} pretraining also preserves the physics consistency of the features throughout the \gls{nn} layers. As a result, the different-looking images of the same class are positioned close in the latent space, leading to improved classification results compared to a supervised CNN. 
Further, \citeauthor{maMulticropFusion2022} leverage contrastive learning to create prototypes for landcover classification \cite{maMulticropFusion2022}. Their online learning approach fuses the images with their enhanced counterparts at different resolutions. The latent space is mapped into a unit sphere where the prototypes are clustered. Another unsupervised prototype approach is provided in \cite{rosaLearningCrop2023}.

\subsubsection{Lack of Standardized and Objective Evaluation of xAI}
As mentioned in Section \ref{sec:rq4_recommended_partices}, the qualification of the interpretability is essential for \gls{xai} approaches \cite{lipton2018mythos}. Contrary to this, there is no definition of a sufficient explanation or interpretation. Furthermore, no standardized and objective evaluation for \gls{xai} methods has been established, posing a challenge for potential partitioners outside of the \gls{xai} domain. Although metrics and tools for the evaluation exist, incorporating \gls{xai} into a \gls{ml} pipeline would mean acquiring the domain knowledge from \gls{xai} to provide a good evaluation for the methods used by a \gls{rs} partitioner.
An alternative approach to the evaluation metrics is the conduction of user studies. These studies evaluate how useful the interpretation appears to the participants. 
However, conducting a well-designed user study involves a lot of effort, costs, and domain knowledge, which again would need to be acquired by the partitioner.
In summary, the evaluation of \gls{xai} lacks a standardized methodology, potentially limiting non-experts applying the methods. This circumstance might have contributed to the large number of anecdotal evidence we encountered in this review.

\section{Conclusion}
\label{sec:conclusion}
\noindent This paper provides a detailed overview of the state-of-the-art of \gls{xai} in \gls{rs} by conducting a systematic review of the existing work in the field. First, we collected a large set of publications covering the last 6 years by executing extensive search queries in the established literature databases. Subsequently, we introduced a categorization for the existing \gls{xai} methods to structure these works. Our analysis reveals that a substantial amount of work is motivated by the assessment of the trustworthiness of the \gls{ml} models for traditional \gls{eo} tasks like landcover mapping and agricultural monitoring. More recently, \gls{xai} has been increasingly utilized for the discovery of scientific insights for critical \gls{eo} problems related to climate change, extreme events, or urbanization. Although dominantly established \gls{xai} methods like \gls{shap} or \gls{gradcam} are frequently used, we observe the increased development of adapted \gls{xai} methods to capture the specifics of \gls{rs} data. Further, we discuss the works combining \gls{xai} with other fields, like physics and uncertainty, ultimately enhancing the quality of the extracted explanations.

These highlights clearly illustrate that \gls{xai} in \gls{rs} is a young field with high potential to augment the \gls{ml} knowledge extraction process. At the same time, comparing how the identified practices in this review relate to the latest developments in \gls{xai} and the current limitations in \gls{eo} gives a hint to the fundamental challenges that need to be addressed in future studies. One research direction of paramount importance is the development of interpretable \glspl{dnn} to address the shortcomings of the widely-used post-hoc in \gls{rs}. Further, as over 90\% of the studies conduct only anecdotal evaluation, verifying the reliability of the \gls{xai} outputs supports future work on quantitative evaluation and user studies. Moreover, the adapted methods indicate that the traditional \gls{xai} approaches do not conform to the properties of \gls{rs} data. Therefore, we encourage the development of methods to address those drawbacks related to scale, topology, and temporal dependencies in \gls{rs} data. Another challenge is the lack of labeled data, which is currently tackled by combining \gls{ssl} with \gls{xai} to design approaches that outperform supervised \gls{ml} models, additionally offering intuitive model interpretability. Summarizing our contributions, we hope that the insights provided in this review enable the researchers to better understand the state-of-the-art in the field and promote the development of novel methods by tackling the research directions proposed.

\section*{Acknowledgments}
The work of A. Höhl was funded by the project ML4Earth by the German Federal Ministry for Economic Affairs and Climate Action under grant number 50EE2201C. The work of I. Obadic is funded by the Munich Center for Machine Learning. M.Á. Fernández-Torres acknowledges the support from the European Research Council (ERC) under the ERC Synergy Grant USMILE (grant agreement 855187) and the European Union’s Horizon 2020 research and innovation program within the project `XAIDA: Extreme Events - Artificial Intelligence for Detection and Attribution,' under grant agreement No 101003469. The work of M.Á. Fernández-Torres and X. Zhu is supported by the German Federal Ministry of Education and Research (BMBF) in the framework of the international future AI lab "AI4EO -- Artificial Intelligence for Earth Observation: Reasoning, Uncertainties, Ethics and Beyond" (grant number: 01DD20001). H. Najjar acknowledges support through a scholarship from the University of Kaiserslautern-Landau. The authors are responsible for the content of this publication.

\section*{Competing interests}
The authors declare that they have no known competing financial interests or personal relationships that could have appeared to influence the work reported in this paper.

\printbibliography

\newpage

\section*{Biographies}
\begin{IEEEbiographynophoto}{Adrian Höhl}
(\href{mailto:adrian.hoehl@tum.de}{adrian.hoehl@tum.de}) received his M.Sc. degree in Robotics, Cognition, Intelligence from the Technical University of Munich, Munich, Germany, in 2021. He is pursuing his Ph.D. degree at the Technical University of Munich, Munich, Germany. He spent a three-month research stay at the Alfred-Wegener Institute, a Helmholtz Centre for Polar and Marine Research, in Potsdam, Germany, in 2023. His research interests include explainable artificial intelligence, remote sensing, and extreme climate events.
\end{IEEEbiographynophoto}

\begin{IEEEbiographynophoto}{Ivica Obadic}
(\href{mailto:ivica.obadic@tum.de}{ivica.obadic@tum.de}) received his BSc in Computer Science and Engineering from the Faculty of Computer Science and Engineering, North Macedonia in 2017 and his M.Sc. degree in Data Science from the Ludwig Maximilian University of Munich, Germany in 2019. He is pursuing his Ph.D. degree at the Technical University of Munich, Munich, Germany. His research interests include remote sensing, deep learning, explainable artificial intelligence, and graph mining. He has spent a two-month research stay at the Lancaster Intelligent, Robotic and Autonomous Systems Centre, Lancaster Univeristy in the UK in 2023.
\end{IEEEbiographynophoto}

\begin{IEEEbiographynophoto}{Miguel-Ángel Fernández-Torres}
(\href{mailto:miguel.a.fernandez@uv.es}{miguel.a.fernandez@uv.es}) received the Degree in Audiovisual Systems Engineering and the Master’s and Ph.D. Degrees in Multimedia and Communications from the Universidad Carlos III de Madrid, Spain, in 2013, 2014, and 2019, respectively. He works at present as a postdoctoral researcher in the Image and Signal Processing Group at the Universitat de València, Spain. His research interests include the design of spatio-temporal deep generative models and machine attention mechanisms to be deployed for extreme event detection, forecasting, and understanding. He currently takes part in several European projects which tackle these topics, including the \href{https://www.usmile-erc.eu/}{USMILE} European Research Council (ERC) grant, as well as the \href{https://xaida.eu/}{XAIDA} Horizon 2020 (H2020) and the \href{https://rsc4earth.de/project/deepextremes/}{ESA DeepExtremes} project. In addition, he has participated in other projects within the Computer Vision field, which include visual attention modelling and understanding, image and video analysis, and medical image analysis and classification. In all these fields, he has coauthored several papers in prestigious international journals and revised conferences. He has also done two research stays at the Visual Perception Laboratory of Purdue University, West Lafayette, Indiana, USA, and the International Future Lab AI4EO at Technische Universität München, Munich, Germany, in 2016 and 2022, respectively.  
\end{IEEEbiographynophoto}

\begin{IEEEbiographynophoto}
{Hiba Najjar} (\href{mailto:hiba.najjar@dfki.de}{hiba.najjar@dfki.de}) received her M.Sc. degree in Applied Mathematics from the École Nationale Supérieure des Mines de Nancy, Nancy, France, in 2021. She is pursuing her Ph.D. degree at the University of Kaiserslautern-Landau, Kaiserslautern, Germany. Her research interests include explainable AI, remote sensing, and agricultural applications.
\end{IEEEbiographynophoto}

\begin{IEEEbiographynophoto}{Dario Augusto Borges Oliveira}
(\href{mailto:dario.oliveira@fgv.br}{dario.oliveira@fgv.br}) received his Ph.D. from Puc-Rio (Brazil) in 2013, doing part of his Ph.D. research at the Leibniz University of Hannover, Germany, and the Instituto Superior Técnico, Lisbon, Portugal. From 2014 to 2015, he worked as a research fellow at the Institute of Mathematics and Statistics from the University of Sao Paulo, and in 2015 he transitioned to the industry working at General Electric Global Research Center in Rio de Janeiro and at the IBM Research lab in São Paulo, Brazil. In 2021, he returned to the academy as a Guest Professor at the Technical University of Munich, Germany. Since 20222, he has worked as a Professor at the School of Applied Mathematics - Getulio Vargas Foundation, Rio de Janeiro, Brazil.
\end{IEEEbiographynophoto}

\begin{IEEEbiographynophoto}{Zeynep Akata}(\href{mailto:zeynep.akata@helmholtz-munich}{zeynep.akata@helmholtz-munich.de})
is the director of the Institute for Explainable Machine Learning at Helmholtz Munich and Helmholtz AI. She also holds the chair for Interpretable and Reliable Machine Learning at the Technical University of Munich. After completing her PhD at the INRIA Rhone Alpes in 2014, she worked as a post-doctoral researcher at the Max Planck Institute for Informatics and at University of California Berkeley until 2017 and as an assistant professor at the University of Amsterdam  between 2017 and 2019. Before moving to Munich in 2024, she was a professor of Computer Science (W3) within the Cluster of Excellence Machine Learning at the University of Tübingen between 2019 and 2023. She received a Lise-Meitner Award for Excellent Women in Computer Science from Max Planck Society in 2014, a young scientist honour from the Werner-von-Siemens-Ring foundation in 2019, an ERC-2019 Starting Grant from the European Commission, The DAGM German Pattern Recognition Award in 2021, The European Computer Vision Association Young Researcher Award from the European Computer Vision Association in 2022 and the Alfried Krupp Young Researcher Award from the Alfried Krupp-von-Bohlen-und-Halbach Foundation in 2023. She is a fellow of ELLIS Society and an Associate Editor in Chief of IEEE Transactions on Pattern Analysis and Machine Intelligence. Her research interests include multimodal learning and explainable AI.
\end{IEEEbiographynophoto}

\begin{IEEEbiographynophoto}
{Andreas Dengel}
(\href{mailto:andreas.dengel@dfki.de}{andreas.dengel@dfki.de})
is the Executive Director of DFKI in Kaiserslautern. He received Diploma from University of Kaiserslautern in 1986 and a PhD from University of Stuttgart in 1989. He worked for IBM, Siemens and Xerox Parc and became a Professor at the Department of Computer Science at TU Kaiserslautern in 1993. Since 2009, he has held another professorship (kyakuin) in the Department of Computer Science and Intelligent Systems at Osaka Metropolitan University with teaching and examination rights. At this university, he was also named "Distinguished Honorary Professor" (tokubetu eiyo kyoju) in March 2018, a distinction received by only five researchers within 135 years. Andreas has chaired numerous international conferences and serves on the editorial boards of international journals and book series. He has written or edited 14 books and is the (co-)author of more than 600 peer-reviewed scientific publications, many of which have received Best Paper Awards. To date, he has supervised more than 500 doctoral, master's, and bachelor's theses. In addition to a number of keynote lectures at international meetings and conferences, Andreas has given invited technical presentations at numerous prestigious universities and research institutions. These include MIT, Stanford University, PARC, UC Berkeley, CMU, London School of Economics, Cambridge University, ATR, NII, Tokyo University, Chinese Academy of Science, and Google Research or MS Research. Also activities as a lecturer, e.g., in the Joint Executive MBA Program at Johannes Gutenberg University in Mainz, the University of Texas in Austin and Dongbei University of Finance and Economics in Dalian, China, are part of Andreas' previous activities. He is an IAPR fellow and a member of the member of the National Academy of Science and Technology (acatech). In addition to the honors already mentioned, Andreas has received other important recognitions for his work, some of which are mentioned below. Back in 1997, Andreas already received one of the most prestigious personal science awards in Germany, the Alcatel/SEL Award for Technical Communication, for his scientific achievements. In 2019, he also received the Outstanding Achievement Award from the Int'l Conference on Document Analysis and Recognition (ICDAR) in Sydney, Australia. Finally, in 2021, Andreas was awarded the oldest Japanese order, the "The Order of the Rising Sun, Gold Rays with Neck Ribbon" in the name of His Majesty Emperor Naruhito and in 2022 he was awarded the Order of Merit of the State Rhineland-Palatinate.
\end{IEEEbiographynophoto}

\begin{IEEEbiographynophoto}{Xiao Xiang Zhu}
(\href{mailto:xiaoxiang.zhu@tum.de}{xiaoxiang.zhu@tum.de}) (S'10-M'12-SM'14-F'21) received the Master (M.Sc.) degree, her doctor of engineering (Dr.-Ing.) degree and her “Habilitation” in the field of signal processing from Technical University of Munich (TUM), Munich, Germany, in 2008, 2011 and 2013, respectively. She is currently the Professor for Data Science in Earth Observation (former: Signal Processing in Earth Observation) at Technical University of Munich (TUM) and the Head of the Department ``EO Data Science'' at the Remote Sensing Technology Institute, German Aerospace Center (DLR). Since 2019, Zhu is a co-coordinator of the Munich Data Science Research School (www.mu-ds.de). Since 2019 She also heads the Helmholtz Artificial Intelligence -- Research Field ``Aeronautics, Space and Transport". Since May 2020, she is the director of the international future AI lab "AI4EO -- Artificial Intelligence for Earth Observation: Reasoning, Uncertainties, Ethics and Beyond", Munich, Germany. Since October 2020, she also serves as a co-director of the Munich Data Science Institute (MDSI), TUM. Prof. Zhu was a guest scientist or visiting professor at the Italian National Research Council (CNR-IREA), Naples, Italy, Fudan University, Shanghai, China, the University of Tokyo, Tokyo, Japan and University of California, Los Angeles, United States in 2009, 2014, 2015 and 2016, respectively. She is currently a visiting AI professor at ESA's Phi-lab. Her main research interests are remote sensing and Earth observation, signal processing, machine learning and data science, with a special application focus on global urban mapping. Dr. Zhu is a member of young academy (Junge Akademie/Junges Kolleg) at the Berlin-Brandenburg Academy of Sciences and Humanities and the German National Academy of Sciences Leopoldina and the Bavarian Academy of Sciences and Humanities. She serves in the scientific advisory board in several research organizations, among others the German Research Center for Geosciences (GFZ) and Potsdam Institute for Climate Impact Research (PIK). She is an associate Editor of IEEE Transactions on Geoscience and Remote Sensing and serves as the area editor responsible for special issues of IEEE Signal Processing Magazine. She is a Fellow of IEEE. 
\end{IEEEbiographynophoto}

\clearpage
\newpage
\begin{refsection}
\newpage

\begin{appendix}

\subsection{Research Method}
\label{sec:method}
\noindent In this review, we aim for transparency and reproducibility of our work by following the PRISMA scheme \cite{page_prisma_2021}. We apply the appropriate PRISMA requirements for our field to provide a transparent, complete, and trustworthy review.
Since literature search engines, such as Google Scholar, do not provide the same search results for all users (the results depend on the geographic location and time) \cite{gusenbauerWhichAcademicSearch2020}, we avoided the use of such search engines and relied on databases where the results can be reproduced.

\subsubsection{Search Procedure}
\label{chap:search_query}
\noindent The search query consists of two major parts: keywords related to \gls{xai} and keywords related to \gls{eo}. All the nested keywords are connected via an $\operatorname{OR}$ operator, while the two parts are connected via the $\operatorname{AND}$ operator.
Due to the interchangeably used taxonomy in both fields, we added additional keywords to the generally known terms to receive as many relevant papers as possible, attempting not to excessively increase the false positive rate. For example, we included specific types of remote sensing sensors and used wildcards to cover different ways authors might refer to \gls{xai}. 

\lstdefinelanguage{LanQuery}{
    morekeywords={or, and},
    sensitive=false,
}


\lstdefinestyle{mystyle}{
    basicstyle=\ttfamily\footnotesize,
    breakatwhitespace=false,         
    breaklines=true,                 
    captionpos=b,                    
    numbers=left,                    
    numbersep=5pt,                  
    showspaces=false,                
    showstringspaces=false,
    showtabs=false,                  
    tabsize=2,
    numbers=none,
     xleftmargin=\parindent,
     xrightmargin=\parindent,
}

\lstset{style=mystyle}
\begin{lstlisting}[language = LanQuery,mathescape=true]
$\Bigl[$Earth observation OR remote sensing OR earth science OR $\bigl($(satellite OR aerial OR airborne OR spaceborne OR radar) AND (image OR data)$\bigl)$ OR LiDAR OR SAR OR UAV OR Sentinel OR Landsat OR MODIS OR gaofen OR ceres$\Bigl]$
                        AND
$\Bigl[$xai OR $\bigl($(interpret* OR explain*) AND (deep learning OR machine learning OR artificial intelligence OR dl OR ml OR ai OR model)$\bigl)$$\Bigl]$

\end{lstlisting}

This general search query was adapted to the different search filters in three databases: Scopus\footnote{\href{https://www.scopus.com/}{https://www.scopus.com/}}, Springer\footnote{\href{https://link.springer.com/}{https://link.springer.com/}}, and IEEE\footnote{\href{https://ieeexplore.ieee.org/}{https://ieeexplore.ieee.org/}}.

\begin{figure*}[htbp]
  \centering
    \includegraphics[trim={1.2cm 3.5cm 1.2cm 1.9cm},clip,width=\textwidth]{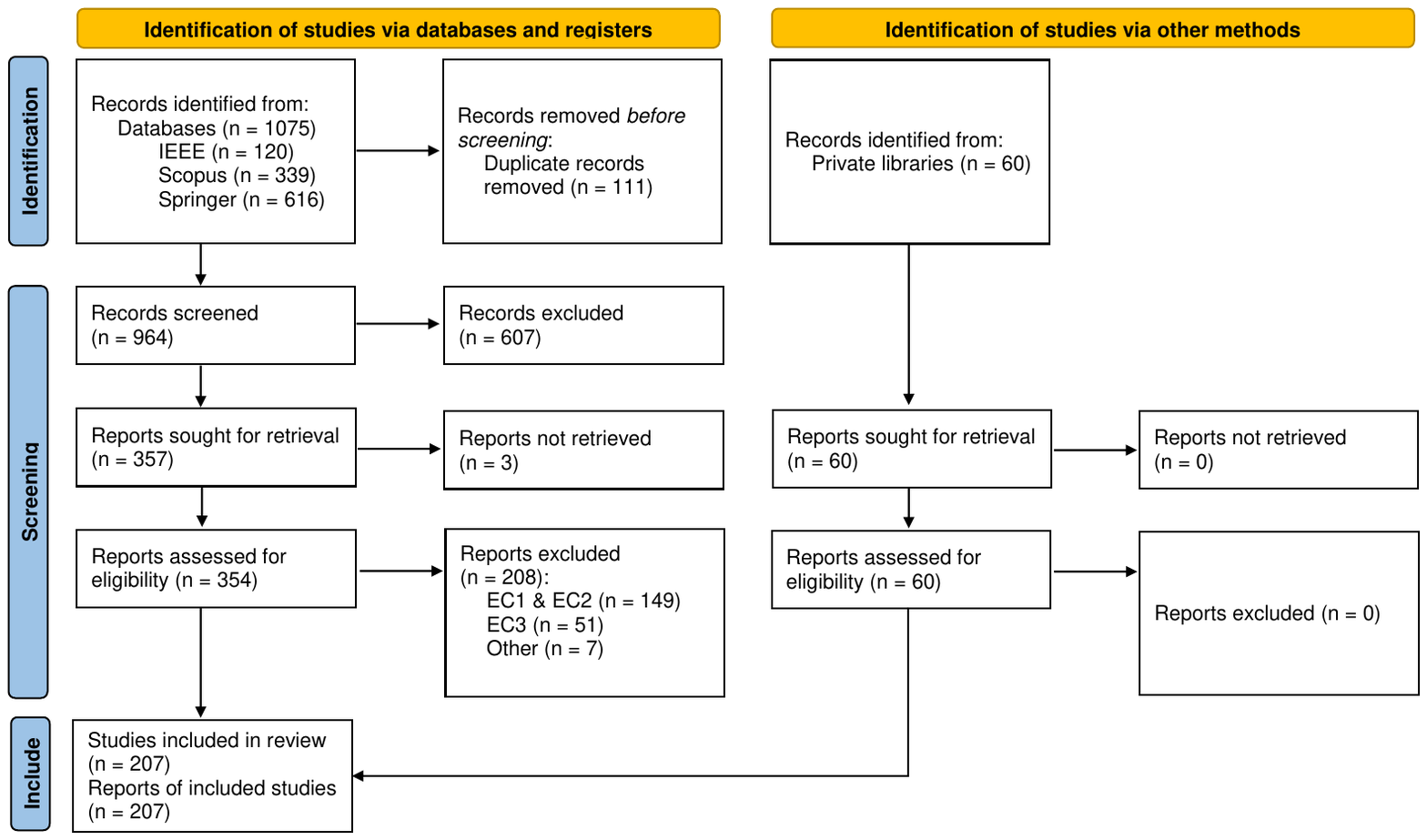}
  \caption{Flow Diagram of PRISMA. Figure adapted from \cite{page_prisma_2021}.}
  \label{fig:flowchart_prisma}
\end{figure*}

The field of \gls{xai} started to gain momentum from 2014 \cite{viloneExplainableArtificial2020}. However, we found that search results of \gls{xai} papers in \gls{rs} before 2017 are rarely relevant, supporting that the methods have only been used in the last few years in this area. Therefore, we searched journal and conference papers between 01.01.2017 and 31.10.2023, which is the last time the search was executed in the databases. Hence, our review comprises more than six years. We considered papers discussing \gls{rs} in \gls{eo}, namely \gls{rs} sensors mounted to aerial vehicles and satellites. We included all the available data from these sensors, as well as high-level and compound products, such as \glspl{dem} and the ERA5 dataset \cite{hersbachERA5GlobalReanalysis2020}. By contrast, the filtering or exclusion criteria we followed can be summarized as follows:
    \begin{itemize}
        \item EC1 - Publication unrelated to \acrlong{xai}
        \item EC2 - Publication unrelated to remote sensing
        \item EC3 - Review/survey/short conference paper
        \item EC4 - Publication published before 2017
        \item EC5 - Publication not written in English
    \end{itemize}
The covered work must discuss the interpretability of the method or the explanation results. Hence, it is not enough to only use a model which is interpretable by itself, like a decision tree or linear regression (EC1). We explicitly excluded in-situ measurements and pictures actively taken in the surrounding environment, which are non-\gls{rs} products, as well as papers out of the scope of \gls{rs} and \gls{xai} (EC2). Furthermore, review and survey papers were excluded (EC3) since they would be included in the related work section. Also, short conference papers that do not exceed a minimum of 5 pages are excluded (EC3) because they usually contain preliminary or incomplete results. The results of the search were filtered in three steps: (i) removing duplicates, conference abstracts, and reports, (ii) screening through the abstract, and (iii) screening through the full text. At least one author read each abstract and paper; whenever there was doubt, the other authors read the corresponding paper as well.
In total, our search results in 1075 papers, merging the different sources and removing duplicates, left us with 964 papers. In the first shallow abstract screening, we removed 607 papers because they were unrelated to our review topic. Three papers had to be excluded due to a lack of access rights. After the full-text screening, we were left with 147 papers. These papers were accompanied by 60 papers we had in our libraries. The procedure is summarized in Figure \ref{fig:flowchart_prisma}.
\subsection{Common Explanation Methods in Remote Sensing}
\label{ssec:prom_xai_methods}
\noindent In this section, we provide an in-depth explanation of several popular \gls{xai} methods in the field of \gls{rs}.

\vspace{0.25cm}
\subsubsection{Class Activation Mapping (CAM) and Grad-CAM}
\citeauthor{zhou2016learning} introduce \gls{cam} \cite{zhou2016learning} to obtain visual heatmaps for class discriminative localization in \glspl{cnn}. This technique aims to identify important regions within an image that influence the model's decision toward each of the $C$ classes. The approach requires the presence of a global average pooling layer between the last convolutional layer and the top fully connected layer of the \gls{cnn} architecture. To calculate the class activation maps, the weights $w_{c,k}$ at the fully connected layer associated with a particular class $c$ are used to estimate the importance of each feature or activation map $A_{k}$ at the input of the global average pooling layer. Finally, the saliency map highlighting the discriminative regions for the class $c$ is computed as the weighted sum of these $K$ activation maps with the following equation:
\begin{equation}
    \label{eq:cam}
    \operatorname{S}^{CAM}_c = \sum_{k=1}^{K} w_{c,k}A_k
\end{equation}
\citeauthor{selvaraju2017grad} generalize \gls{cam} to a broader range of \gls{cnn} architectures by introducing \gls{gradcam} \cite{selvaraju2017grad}. In contrast to \gls{cam}, which makes use of the weights corresponding to class $c$, \gls{gradcam} utilizes the average gradient of the logit $y_c$ for each of the $K$ feature maps in the last convolutional layer to assess their importance for class $c$. Additionally, a ReLU function is applied, propagating only the positive values. Therefore Eq. \ref{eq:cam} becomes:
\begin{equation}
    \label{eq:gradcam}
    \operatorname{S}^{Grad-CAM}_c = \operatorname{ReLU}\bigg(\sum_{k=1}^{K}\Big(\frac{1}{N} \sum_i \sum_j \frac{\partial y_c}{\partial A_{i,j}} \Big)A_k\bigg),
\end{equation}
\noindent where $(i,j)$ are the spatial coordinates of each of the $N$ locations in every activation map $A_k$. 
Because the convolutional layers of the networks usually reduce the input size, the activation maps need to be upsampled to the original input size. However, this upsampling process can lead to imprecisions and other backpropagation methods produce more fine-grained heatmaps, which directly attribute to each pixel, such as \gls{lrp} or \gls{ig}. Nevertheless, recent method revisions, such as LayerCAM \cite{jiangLayerCAMExploring2021}, have aimed to produce more detailed heatmaps.

\vspace{0.25cm}
\subsubsection{Occlusion Senstivity}
\citeauthor{occ_zeiler2014visualizing} propose a perturbation-based method \cite{occ_zeiler2014visualizing} to visualize the importance of different image regions. Their approach slides a patch over the image and observes the sensitivity of the model's prediction. Different values for the patch, its size, and sampling techniques can be considered. The assigned importance is directly proportional to the drop in performance of the model after occluding the patch. Consequently, this method is model-agnostic and can be applied to any kind of architecture.

\vspace{0.25cm}
\subsubsection{Local Interpretable Model-agnostic Explanation
(LIME)} \citeauthor{ribeiroWhyShouldTrust2016} proposed \gls{lime} \cite{ribeiroWhyShouldTrust2016}, a model-agnostic method that approximates the behavior of a complex model locally, in the neighborhood of a target instance. Concretely, to explain the prediction of a complex model $f$ for a target instance $x$, \gls{lime} performs the following main steps: 1) a dataset is created around the neighborhood of $x$ by randomly performing perturbations on it (e.g., adding noise, hiding or blurring parts of the input, etc.), 2) an interpretable by-design surrogate model $g$ is trained on this dataset and 3) the internals of $g$ are inspected to provide an explanation. It is important to note that \gls{lime} generates explanations on a simplified representation space that is interpretable to humans. For example, when the input is an image, the simplified representation can correspond to a binary vector, indicating the presence of superpixels decomposing the image. Therefore, in the first step, \gls{lime} creates a dataset by perturbing the simplified representation of $x$. This dataset is labeled according to the predictions of the complex model $f$ on the perturbed instances (the perturbed instances are reverse-transformed into the original input representation before feeding it into the complex model). 
In the second step, the surrogate model $g$ is trained on this dataset by weighting the perturbed samples based on their similarity with $x$. Finally, the internals of $g$ are inspected to explain the prediction of $f$ for $x$. For instance, in case $g$ is a linear regression model, its coefficients can be used to assess the feature importance. In contrast, if $g$ is a decision tree, inspecting its rules can serve to explain the predictions of the complex model $f$. 

\vspace{0.25cm}
\subsubsection{SHapley Additive exPlanations (SHAP)} 
The Shapley values \cite{shapley:book1952} constitute an approach from cooperative game theory used to estimate the importance of the input features for the prediction of a \gls{ml} model $f$ on a local instance, as well as their average marginal contribution across all possible coalitions of features. Concretely, given an instance $x$, the Shapley value $\phi_i$ for feature $i$ is computed as follows:
\begin{equation}
    \label{eq:shap}
    \phi_i(f,x)=\sum_{S \subseteq F \backslash \{i\} } \frac{\left|S\right| !\left(F-\left|S\right|-1\right) !}{F !}\left[f_x\left(S \cup \{i\} \right)-f_x\left(S \right)\right],
\end{equation} 
where $F$ is the set of all input features, $S$ is a coalition of features, and $f_x\left(S\right)$ is the marginalized prediction over the features not included in $S$ while the features in the coalition $S$ take the values of the instance $x$. 
The Shapley values are considered to fairly assign the contribution of the input features to the model prediction as they satisfy the following properties: 1) the sum of the feature contributions adds up to the difference between the model prediction for the instance $x$ and the average model prediction on the dataset (efficiency), 2) the contributions of two features are the same if they equally contribute to all coalitions (symmetry), 3) a zero contribution is assigned to the features that do not change the model prediction (dummy), and 4) the Shapley value of a feature for an ensemble of \gls{ml} models can be computed by aggregating the individual Shapley values across the models in the ensemble (additivity) \cite{Molnar2019}.

The analytical solution for the Shapley values can be a computationally expensive operation for models trained on more than a few features, as Eq. \ref{eq:shap} requires iterating over all possible feature coalitions and computing the marginal contribution in each coalition. Therefore, in practice, the Shapley values are estimated with approximation techniques. \citeauthor{lundbergUnifiedApproach2017} introduce the \gls{shap} framework \cite{lundbergUnifiedApproach2017} as a unified approach for model interpretability based on the family of additive feature attribution methods. This family represents the explanation through the coefficients of a linear model and has a similar set of desired explanation properties to the ones of the Shapley values. Although other \gls{xai} methods like \gls{lime}, \gls{lrp}, and \gls{deeplift} \cite{shrikumar2017learning} can be represented in the form of additive feature attribution methods, the only additive method that satisfies the desired explanation properties is the one having the Shapley values as its linear coefficients (referred as \gls{shap} values). This formulation enables the approximation of the Shapley values with a model-agnostic approach based on \gls{lime}. Concretely, the authors introduce Kernel SHAP, an approach that estimates the Shapley values by constraining the \gls{lime} method to rely on a linear model as a surrogate model and to use a specific similarity function for the weighting of the perturbed instances. To leverage the internal structure of the \gls{ml} models for a fast approximation of the Shapley values, the authors also propose model-specific approaches, namely Deep SHAP for \glspl{dnn} and Tree SHAP \cite{lundberg2018consistent} for tree-based models.

\vspace{0.25cm}
\subsubsection{Attention Mechanism} Attention mechanisms represent an integral component of neural networks that mimics cognitive attention \cite{brauwers2021general}. Although initially proposed for natural language processing tasks like machine translation and sentiment analysis \cite{bahdanau2014neural, vaswani2017attention}, they have also been leveraged recently in \gls{cv} applications \cite{DosovitskiyB0WZ21}, as well as for graph-structured data \cite{velikovi2017graph}. These modules induce attention weights on the input features, determining how a neural network combines them to produce a high-level feature representation. Thus, visualizing the attention weights is a well-known approach to understanding the relevant features for the model predictions and assessing the interaction of the input features in the context of the learning task \cite{clark-etal-2019-bert, caron2021emerging}. 
\\
An attention mechanism is usually defined for a query $\bf q$ and matrices of key and value pairs $\mathbf{K} = [\mathbf{k}_{1},...,\mathbf{k}_{L}]$ and $\mathbf{V} = [ \mathbf{v}_{1},...,\mathbf{v}_{L}]$, respectively. The output is a high-level representation $\bf c$ depending on the values and based on the alignment of the keys $\bf K$ with the query $\bf q$. The alignment function is specified according to the attention weights and can be computed using various functions proposed in the literature \cite{chaudhari2021attentive}. One of the most widely used mechanisms is the \emph{scaled dot-product attention} introduced by \citeauthor{vaswani2017attention} in \cite{vaswani2017attention}, where attention weights $\alpha$ are computed as follows:
\begin{equation}
    \label{eq:att_vaswani}
    \alpha = \operatorname{softmax}(\frac{\mathbf{q} \mathbf{K}^{T}}{\sqrt{d_k}}),
\end{equation}
being $d_k$ the embedding dimension of the keys. 
Then, the high-level value representation $\bf v$ is computed as a linear combination of the attention weights and the value vectors $\bf c = \alpha \bf V$.

\subsection{Usage of Explainable AI for the Most Common Earth Observation Tasks}
\label{sec:add_eo_task_plots}
\label{sec:3_eo_tasks}
\begin{figure}
  \centering
  \subfloat[Landcover mapping.]{\includegraphics[trim={2cm 1.75cm 2cm 1cm},clip, width=0.5\textwidth]{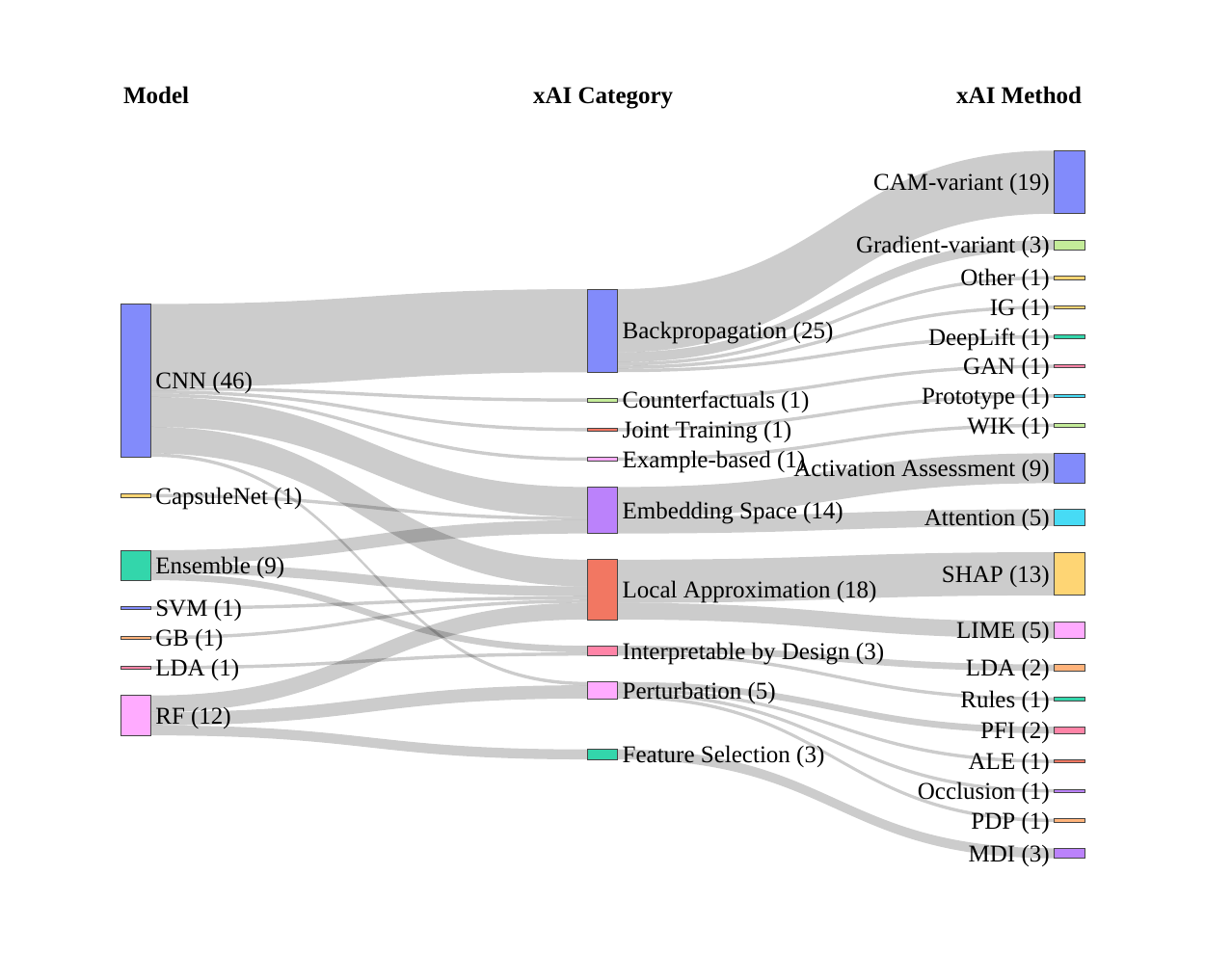}}\\
  \subfloat[Agricultural monitoring.]{\includegraphics[trim={2cm 1.75cm 2cm 2.5cm},clip,width=0.5\textwidth]{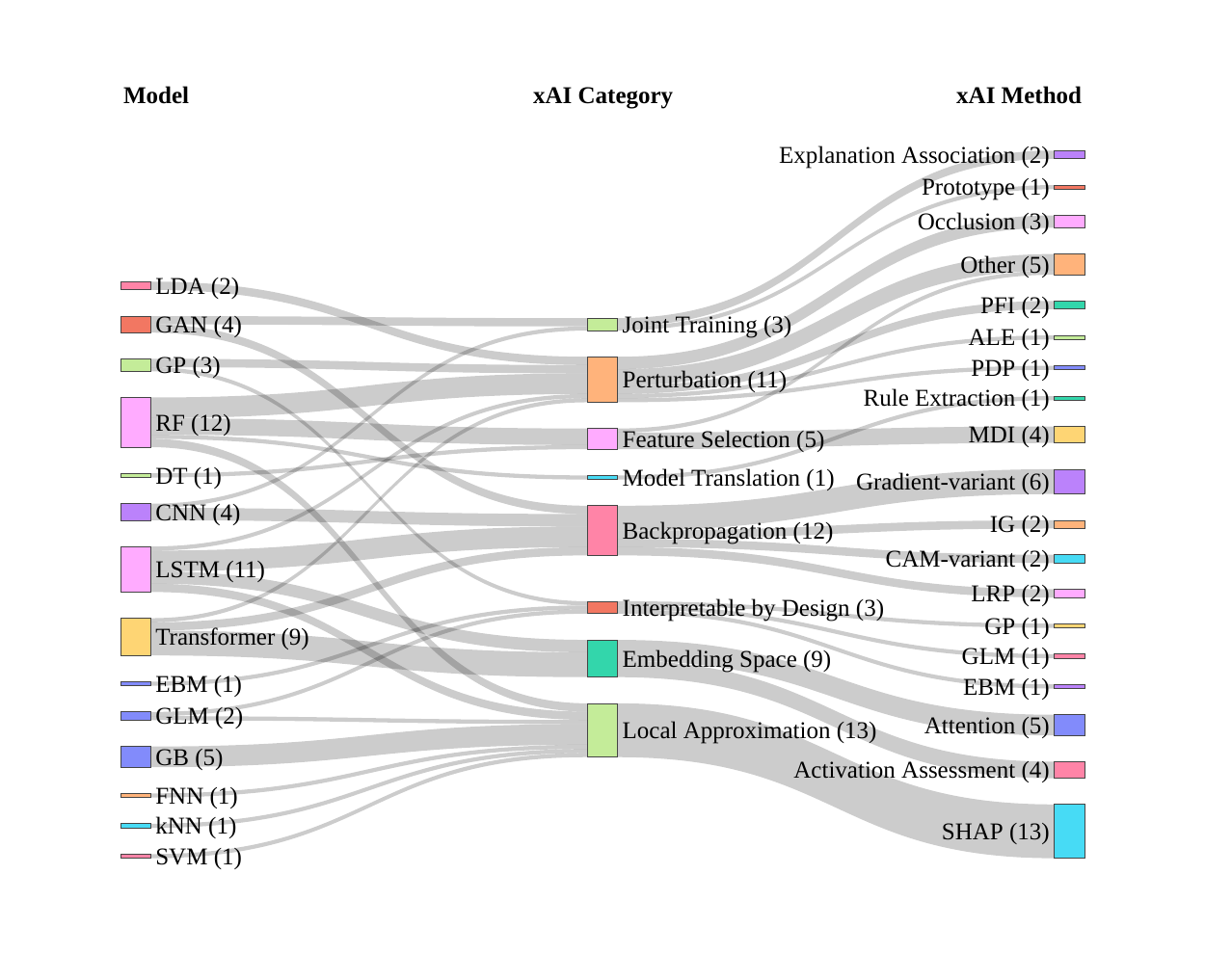}}\\
  \subfloat[Natural hazard monitoring.]{\includegraphics[trim={2cm 1.75cm 2cm 2cm},clip,width=0.5\textwidth]{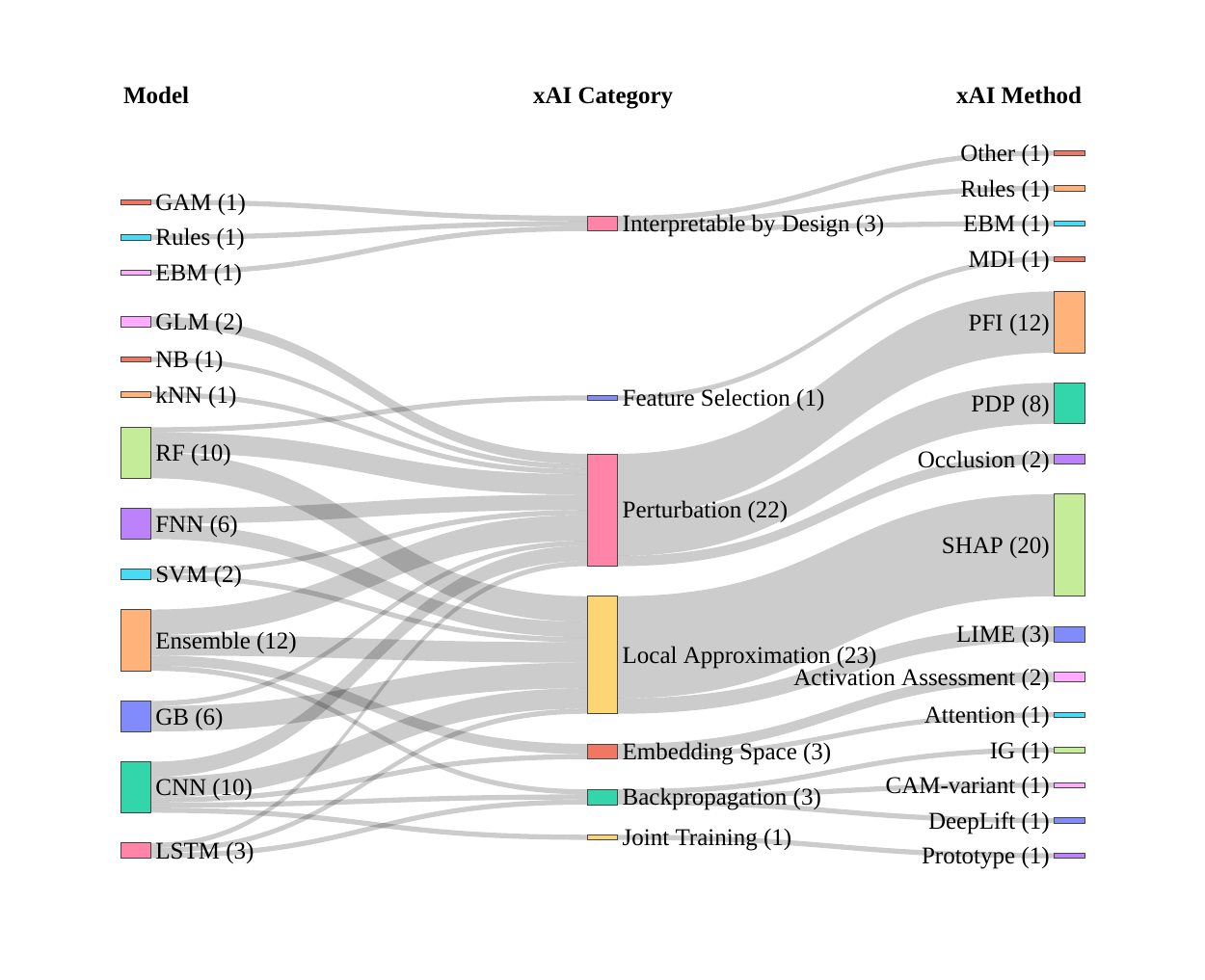}}\\

  \caption{The number of times models, \gls{xai} categories, and \gls{xai} methods are used for the most common \gls{eo} tasks.}
  \label{fig:xai_xaif_in_eo}
\end{figure}

To provide a deeper insight, we will now discuss in detail the three most representative \gls{eo} tasks according to the number of papers, which are landcover mapping, agriculture monitoring, and natural hazard monitoring. The graphs in Figure \ref{fig:xai_xaif_in_eo} show these tasks and the frequency of the combination in the usage of \gls{ml} models and \gls{xai} methods.\\

\noindent \emph{Landcover Monitoring}\\
Due to the well-established datasets, the most prominent \gls{eo} task is often employed for evaluating or developing new \gls{xai} methods. Further, a low amount of expert knowledge is needed to elucidate the model outputs for this task compared to others, such as atmosphere monitoring or ecosystem interactions. The majority of the 36 landcover mapping papers present \gls{cnn} architectures, leading to the favored use of backpropagation methods, especially from the \gls{cam} family. 
For example, \citeauthor{Sen2020} leverage \gls{gradcam} to identify the reasons for misclassified images \cite{Sen2020}, while \citeauthor{vasuResiliencePlasticity2020} assess transfer learning and the interpretation resilience with \gls{cam} \cite{vasuResiliencePlasticity2020}. The numerous publications proposing novel or evaluating methods can be found in the corresponding Sections \ref{sec:results_adapted_approaches} and \ref{sec:results_eval_xai}, respectively. Albeit with lower frequency than the \gls{cam} approaches, the workings of the \gls{cnn} models are also interpreted based on the local approximation approaches. For example, \gls{shap} is utilized to reveal the salient pixels and the global importance of the spectral bands in \cite{temenosInterpretableDeep2023}, \gls{lime} is used to improve model performance on the misclassified examples in \cite{hungRemoteSensing2020}, and both approaches are utilized in \cite{elliottIdentifyingCritical2022} to validate the model predictions.

The second outstanding \gls{ml} model for landcover monitoring is \gls{rf} and exclusively model-agnostic \gls{xai} methods are used for its interpretation. \citeauthor{jayasingheCausesTea2023} explain the correlations regarding land use for tea cultivation through \gls{lime} \cite{jayasingheCausesTea2023}. The most frequently used approach, \gls{shap}, is leveraged in \cite{burguenoScalableApproach2023}, while \gls{pfi} is explored in \cite{chenEnhancingLand2023} and \cite{brenningInterpretingMachinelearning2023} to rank reflectance data from various satellites and \gls{dem} information. The \gls{dem} information seems to be valuable since both \cite{burguenoScalableApproach2023} and \cite{brenningInterpretingMachinelearning2023} identify slope as the most important feature, while elevation is ranked the highest in \cite{chenEnhancingLand2023}.

Unique to this task, the less common counterfactual and example-based methods are applied to \glspl{cnn}. \citeauthor{dantasCounterfactualExplanations2023} utilize a \gls{gan} to generate counterfactual time series for the \gls{ndvi} of different landcover classes. The generator introduces perturbations to instances and attempts to change the prediction of the pre-trained classifier. Meanwhile, a discriminator ensures the quality of the newly generated instance \cite{dantasCounterfactualExplanations2023}.
Moreover, \citeauthor{ishikawaExamplebasedExplainable2023} show examples of the dataset belonging to the same class by computing the similarity in the latent space of the last layer. These examples can be used to identify \gls{ood} instances or cases where the classifier lacks generalization \cite{ishikawaExamplebasedExplainable2023}.\\

\noindent \emph{Agricultural Monitoring}\\
The primary tasks in this group are crop yield prediction and crop type classification. Two additional tasks are tackled: irrigation scheme classification \cite{arunLearningPhysically2022} and lodging detection \cite{hanExplainableXGBoost2022}.
The methodologies employed revolve around model-agnostic and backpropagation approaches, namely perturbation, local approximation techniques, and \gls{cam} methods.
These methods are frequently integrated with gradient boosting and \gls{rf} tree-based models. Gradient boosting models are consistently interpreted with \gls{shap} \cite{bromsCombinedAnalysis2023, hanExplainableXGBoost2022, huberExtremeGradient2022, jonesIdentifyingCauses2022, singhSimulationMultispectral2023}.
For example, \citeauthor{hanExplainableXGBoost2022} combine gradient boosting with \gls{shap} to identify maize lodging from \gls{uav} images \cite{hanExplainableXGBoost2022}.
The key features identified are plant height computed from a \gls{dem} and \gls{dsm} and textural features from the Gray-Level Co-Occurrence Matrix.

Many different \gls{xai} strategies have been considered for crop yield prediction. \citeauthor{huberExtremeGradient2022} assess the feature importance of weather and spectral bands across three different crop phenology times on administrative boundary levels by summarizing the \gls{shap} values \cite{huberExtremeGradient2022}, while \citeauthor{bromsCombinedAnalysis2023} examine the relevance of soil features and \gls{dem} \cite{bromsCombinedAnalysis2023}. Both studies emphasize the importance of specific time steps and that the spectral bands provide valuable information.
\citeauthor{jonesIdentifyingCauses2022} investigate soil features, \gls{dem}, and spectral reflectance with a finer spatial resolution at subfield yields. Interestingly, no spatially consistent limiting factors are identified for an entire field, indicating the potential of countermeasures on a subfield level \cite{jonesIdentifyingCauses2022}.

Other studies utilize \gls{rf} and perturbation methods, specifically \gls{pfi} \cite{filippiIdentifyingCrop2022} and \gls{ale} or \gls{pdp} \cite{nayakInterpretableMachine2022}. The feature importance of soil, weather, and spectral data on yield potential is analyzed in \cite{filippiIdentifyingCrop2022}, while \gls{pdp} and \gls{ale} are used to assess the interactions between management, weather, and soil data in \cite{nayakInterpretableMachine2022}. Their findings suggest that residue management or rate application decisions can significantly influence crop yield.

Time series data is frequently employed for crop mapping, leading to the utilization of \glspl{rnn} \cite{campos-tabernerUnderstandingDeep2020,mateo-sanchisInterpretableLongShort2023, paudelInterpretabilityDeep2023, xuInterpretingMultitemporal2021} and transformers \cite{obadicExploringSelfAttention2022, russwurmSelfattentionRaw2020, saintefaregarnotSatelliteImage2020, xuInterpretingMultitemporal2021} architectures. Given \gls{lstm} models for crop yield prediction, \gls{ig} and \gls{shap} are used to attribute soil moisture, weather, and reflectance data in \cite{mateo-sanchisInterpretableLongShort2023} and \cite{paudelInterpretabilityDeep2023}. Both studies reach the same conclusion: high temperatures during the growth season have a negative impact on crop yield. \gls{lstm} with attention and transformers allow to differentiate between corn and soybeans in \cite{xuInterpretingMultitemporal2021}. Both models are interpreted not only through their corresponding attention mechanisms but also considering gradients and activation projection with \gls{tsne}. Overall, both models agree in their attribution, emphasizing the middle of the year as an important period when corn starts to silk and soybeans begin to bloom.
\citeauthor{wolaninEstimatingUnderstanding2020} apply \glspl{cnn} to time series for predicting crop yield from vegetation indices and weather data. Attributions and scenario analysis for different weather conditions are provided with a method derived from \gls{cam} and modified for regression, named \gls{ram} \cite{wolaninEstimatingUnderstanding2020}.

Crop classification with prototypes through a \gls{cnn} encoder is proposed by \citeauthor{rosaLearningCrop2023} \cite{rosaLearningCrop2023}. The LLP-Co method uses a priori proportions of the classes to match the instance proportions assigned to the prototypes.
A distinctive approach based on a variational adversarial network for crop yield and irrigation scheme classification is introduced by \citeauthor{arunLearningPhysically2022} \cite{arunLearningPhysically2022}. To learn more meaningful latent representations, the discriminator of the \gls{gan} architecture also needs to classify the latent representations from the encoder into the correct classes. Additionally, \gls{lrp} is used to attribute the inputs. \citeauthor{newmanExplainableMachine2021} use a model translation method to identify \gls{rf} subtrees, common rules within the \gls{rf} model, and their associated error rates \cite{newmanExplainableMachine2021}.
Other works in agriculture monitoring encompass \glspl{gp} \cite{mateo-sanchisLearningMain2021, martinez-ferrerCropYield2021} and \gls{ebm} \cite{celikExplainableArtificial2023}.
Notably, the methods for agricultural monitoring are evaluated solely with anecdotal evidence. No study employs quantitative metrics, and only \cite{paudelInterpretabilityDeep2023} carry out a small user study.\\

\noindent\emph{Natural Hazard Monitoring}\\
The third main task consists of 26 papers on landslide susceptibility, fire, and flood monitoring, as well as geological hazard monitoring.
The typical features used as input for landslide susceptibility assessment are \gls{dem} variables, landcover and vegetation (e.g., \gls{ndvi}) information, and weather variables such as temperature or precipitation. Often, they are supplemented by human factors like the distance to roads or hydrological properties (e.g., drainage density, soil type).
Usually, these models are interpreted with \gls{shap}: \glspl{rf} \cite{vegaLandslideModeling2023, al-najjarNovelMethod2022}, \glspl{svm} \cite{al-najjarNovelMethod2022}, \glspl{cnn} \cite{alqadhiIntegratedDeep2023}, \glspl{fnn} \cite{alqadhiIntegratedDeep2023, dahalExplainableArtificial2022, alqadhiIntegratedDeep2023}, and \gls{gb} \cite{inanExplainableAI2023, zhangInsightsGeospatial2023}. Additionally, \gls{lime}, \gls{pfi} or \gls{pdp} are applied in the same fashion in \cite{sunAssessmentLandslide2023,alqadhiIntegratedDeep2023}.
Other approaches include interpretable models \cite{maxwellExplainableBoosting2021,fangNewApproach2023,youssefLandslideSusceptibility2022}.

In summary, rainfall \cite{fangNewApproach2023, chenAttributionDeep2023, alqadhiIntegratedDeep2023}, slope \cite{maxwellExplainableBoosting2021, inanExplainableAI2023}, elevation, aspect \cite{chenAttributionDeep2023}, curvature \cite{vegaLandslideModeling2023}, distance to road \cite{inanExplainableAI2023, sunAssessmentLandslide2023}, and \gls{ndvi} \cite{vegaLandslideModeling2023,sunAssessmentLandslide2023} are among the most important features for this task. Moreover, higher \gls{ndvi} suggests a decrease in landslide probability \cite{sunAssessmentLandslide2023}. In contrast, mines can increase the probability \cite{fangNewApproach2023}.

The flood mapping approaches rely on similar features as landslide susceptibility (being rainfall a permanent component) and leverage mostly tree-based models or \glspl{cnn} with \gls{shap} \cite{aydinPredictingAnalyzing2022,liuResidualNeural2023,wangXGBoostSHAPApproach2023}, \gls{pdp} \cite{wangXGBoostSHAPApproach2023} or prototypes \cite{zhangInterpretableDeep2022}. The elevation and slope have the biggest influence on floods worldwide and in Turkey, according to \cite{aydinPredictingAnalyzing2022,liuResidualNeural2023}, while rainfall, road density, and building density have more influence on urban floods following \cite{wangXGBoostSHAPApproach2023}.
\citeauthor{zhangInterpretableDeep2022} detect floods by performing adaptive k-means clustering of the image pixels in the latent space of a U-Net encoder \cite{zhangInterpretableDeep2022}. This approach associates the prototypes with the cluster centers and enables the interpretation of the model decisions in terms of linguistic \emph{IF ... THEN} rules. 

Model-agnostic methods (\gls{shap}, \gls{pfi}, \gls{pdp}) are mostly applied to explain fire susceptibility prediction models (\glspl{rf} and \glspl{fnn}, see \cite{abdollahiExplainableArtificial2023, cilliExplainableArtificial2022}), as the former tasks, they typically rely on landcover, weather, and \gls{dem} data. \cite{cilliExplainableArtificial2022} shows that a climate fire index (created index in another work with a DNN), NDVI, and slope are critical indicators for Italy, while \cite{abdollahiExplainableArtificial2023} finds that humidity, wind speed, and rainfall are the most important factors for Australia. Further, \cite{Taylor2020} reveals that fires are more severe in areas with higher elevation and where the dominant vegetation types are shrubs and open forests. 

Various works explore geological hazards. For example, \cite{sainiE2AlertNetExplainable2023} uses \gls{cnn} to classify various disaster events (e.g., building damage, fire) from aerial imagery and reveals the salient regions with a weighted combination of \gls{lime} and \gls{shap} attributions. Further, \cite{bekerDeepLearning2023} utilizes \gls{gradcam} to identify the mistakes of the \gls{cnn} model for predicting volcano deformation patterns and uses \gls{tsne} to evaluate the differences between the latent space representations of real and simulated data.
Regarding other geological hazards, \cite{chenTunnelGeothermal2023} explores the tunnel geothermal disaster susceptibility based on land surface temperature, river density, and other geological factors. They predict the susceptibility based on an ensemble of machine learning models and evaluate the factor importance with \gls{pfi}, \gls{pdp}, and \gls{lime}. Their analysis reveals that land surface temperature, fault density, earthquake peak acceleration, and river density are among the critical indicators. Another study, \cite{jenaExplainableArtificial2023} explores earthquake probability prediction with \gls{xgboost} and explains the model predictions with \gls{shap}.
\subsection{Glossary}
\label{sec:glossary}
We have categorized related \gls{eo} tasks into groups to provide a better overview. Here, we provide a glossary of these groups and the tasks they include.
\begin{description}[]
	\item[Agricultural Monitoring:] All crop-related tasks, like crop yield prediction, crop type classification, irrigation scheme classification, and crop lodging detection.
	\item[Atmosphere Monitoring:] The prediction of atmospheric phenomena, like air quality, aerosol optical depth, and dust storm indices.
	\item[Building Mapping:] All tasks related to buildings and urban structures, like building footprint classification and building damage mapping.
	\item[Ecosystem Interactions:] All interactions of the ecosystem with other systems, e.g., the atmosphere and the hydrosphere. This includes the ecosystem CO2 exchange or the sun-induced fluorescence prediction.
	\item[Human Environment Interaction:] The monitoring of human structures and the environment, like human footprint estimation, socioeconomic status estimation, or well-being prediction.
	\item[Hydrology Monitoring:] Tasks related to hydrology, like runoff forecasting, water quality, streamflow prediction, and water segmentation, but excluding floods.
	\item[Landcover Mapping:] The most common EO task includes mainly landcover classification but also related tasks like slum mapping.
	\item[Natural Hazard Monitoring:] Monitoring of natural hazards, like landslides, wildfires, floods, earthquakes, and volcanos.
	\item[Soil Monitoring:] Monitoring soil properties, like soil texture, respiration, moisture, and salinity.
	\item[Surface Temperature Prediction:] The prediction of the Earth's surface temperature.
	\item[Target Mapping:] Tasks related to the mapping of specific targets, like vehicles and objects.
	\item[Vegetation Monitoring:] Monitoring of vegetation, excluding crops, like vegetation regeneration, tree monitoring, tree classification, and tree mapping.
	\item[Weather Climate Prediction:] Forecasting of weather and climate variables, like precipitation, temperature, and drought.
	\item[Other:] All the tasks which did not fit into the other groups. This includes change detection, urban mobility, sea ice classification, mosquito modeling, and satellite product quality tasks.
\end{description}

\glsadd{aCNN}
\glsadd{aLSTM}
\glsadd{GFFS}
\glsadd{knn}
\glsadd{SMLP}
\glsadd{gb}
\glsadd{vae}
\printglossary[type=\acronymtype,nonumberlist,title={List of Acronyms}]

\end{appendix}

\printbibliography
\end{refsection}

\end{document}